\documentclass[runningheads]{llncs}

\usepackage{eccv}

\usepackage{eccvabbrv}

\usepackage{graphicx}
\usepackage{booktabs}
\usepackage{makecell}
\usepackage{multirow}
\usepackage{stix}
\usepackage{overpic}
\usepackage{comicneue}
\usepackage{twemojis}
\usepackage{array}

\newcolumntype{C}[1]{>{\centering\arraybackslash}m{#1}}

\usepackage[accsupp]{axessibility}  

\usepackage{hyperref}

\usepackage{orcidlink}

\newcommand{\groupone}{Group 1 ($\Nearrow$)}
\newcommand{\grouptwo}{Group 2 ($\Searrow$)}
\newcommand{\groupthree}{Group 3 ($\Rightarrow$)}
\newcommand{\groupfour}{Group 4 ($\nearrow$)}

\newif\ifmaintextonly
\maintextonlyfalse
\newif\ifincludeannotdump
\includeannotdumpfalse

\begin{document}

\title{%
\texorpdfstring{{\color{black!85}A} {\color{black!84}M}%
{\color{black!83}u}%
{\color{black!82}l}%
{\color{black!81}t}%
{\color{black!80}i}%
{\color{black!79}-}%
{\color{black!78}A}%
{\color{black!77}n}%
{\color{black!76}n}%
{\color{black!75}o}%
{\color{black!74}t}%
{\color{black!73}a}%
{\color{black!72}t}%
{\color{black!71}o}%
{\color{black!70}r}  
{\color{black!69}S}%
{\color{black!68}t}%
{\color{black!67}u}%
{\color{black!66}d}%
{\color{black!65}y}  
{\color{black!64}o}%
{\color{black!63}f}  
{\color{black!62}S}%
{\color{black!61}e}%
{\color{black!60}g}%
{\color{black!59}m}%
{\color{black!58}e}%
{\color{black!57}n}%
{\color{black!56}t}%
{\color{black!55}a}%
{\color{black!54}t}%
{\color{black!53}i}%
{\color{black!52}o}%
{\color{black!51}n}  
{\color{black!50}N}%
{\color{black!49}o}%
{\color{black!48}i}%
{\color{black!47}s}%
{\color{black!46}e}  
{\color{black!45}a}%
{\color{black!44}n}%
{\color{black!43}d}  
{\color{black!42}U}%
{\color{black!41}n}%
{\color{black!40}c}%
{\color{black!39}e}%
{\color{black!38}r}%
{\color{black!37}t}%
{\color{black!36}a}%
{\color{black!35}i}%
{\color{black!34}n}%
{\color{black!33}t}%
{\color{black!32}y}  
{\color{black!31}i}%
{\color{black!30}n}  
{\color{black!29}T}%
{\color{black!28}u}%
{\color{black!27}r}%
{\color{black!26}b}%
{\color{black!25}i}%
{\color{black!24}d} {\color{black!23}U}%
{\color{black!22}n}%
{\color{black!21}d}%
{\color{black!20}e}%
{\color{black!19}r}%
{\color{black!18}w}%
{\color{black!17}a}%
{\color{black!16}t}%
{\color{black!15}e}%
{\color{black!14}r} {\color{black!12}I}%
{\color{black!10}m}%
{\color{black!8}a}%
{\color{black!7}g}%
{\color{black!5}e}%
{\color{black!3}s}%
}{A Multi-Annotator Study of Segmentation Noise and Uncertainty in Turbid Underwater Images}
}

\titlerunning{Annotation Uncertainty in Turbid Environments}

\author{Galadrielle Humblot-Renaux\inst{1,2}\orcidlink{0000-0001-7671-7583} \and
Vasiliki Ismiroglou\inst{1,2}\orcidlink{0009-0009-8428-1113} \and
Malte Pedersen\inst{1,2}\orcidlink{0000-0002-2941-9150}}

\authorrunning{G.~Humblot-Renaux et al.}

\institute{Visual Analysis and Perception Laboratory, Aalborg University, Denmark \and
Pioneer Centre for Artificial Intelligence, Denmark
\email{\{gegeh,vasilikii,mape\}@create.aau.dk}}

\maketitle

\begin{abstract}
Label uncertainty and annotator disagreement are common challenges in the field of computer vision, yet their study has largely been confined to the medical domain or to generic image-recognition datasets. Underwater datasets are particularly susceptible to these issues due to the need for domain expertise, degraded visibility conditions, and the inherent difficulty of establishing reliable ground truth in inaccessible environments. Despite these challenges, annotation uncertainty in underwater imagery remains largely unexplored. In this work, we present the first systematic multi-annotator study of segmentation in real underwater scenes, with over 100 participants, and across varying, controlled levels of turbidity. We show that underwater datasets face many of the same annotation challenges as other vision tasks, while turbidity introduces additional systematic errors. We further investigate the main factors driving label noise and explore ways to improve annotation quality in turbid underwater environments, including privileged information, individual effort and annotator ensembles. All (meta-) data collected in this study will be available on the project page: \href{https://vap.aau.dk/tubcertainty}{vap.aau.dk/tubcertainty}
  
\keywords{Underwater vision \and Image annotation \and Label noise \and Multi-annotator disagreement \and Low visibility \and Aleatoric uncertainty \and Segmentation}
\end{abstract}

\begin{figure}[h]
    \centering 
    \includegraphics[width=0.943\linewidth]{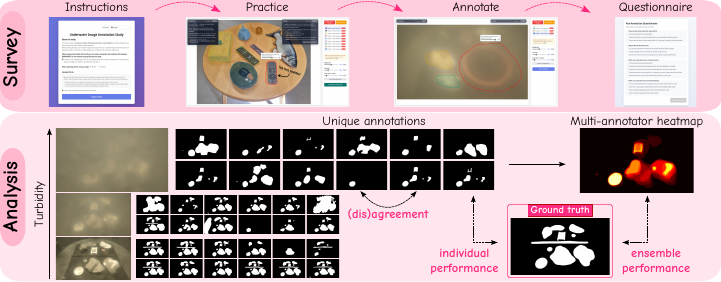}
    \caption{We design a multi-annotator study and collect hundreds of confidence-weighted instance segmentation masks of objects in real underwater scenes across 3 increasing levels of turbidity. We analyze annotation error, variability, and uncertainty in relation to varying visibility.}
    \label{fig:binary_masks_to_heatmap}
\end{figure}
\clearpage
\section{Introduction}

Image and object recognition tasks, such as detection and segmentation, play a crucial role in marine environments. Their applications span habitat coverage estimation, biodiversity and environmental monitoring, underwater construction, quality inspection, surveillance, navigation, and oceanographic surveys, among others~\cite{Raine2026,jaffe2015,Islam2024,suim-dataset_2020,review-cv-benthos_2025,survey-underwater-cv_2023}.
However, the challenging conditions of underwater environments make the validation of annotations particularly difficult, as the corresponding physical objects are often inaccessible for direct inspection. 
Furthermore, annotation quality can be affected by annotator variability, with disagreements arising even among experts due to factors such as environmental conditions, species camouflage, and individual biases~\cite{jambo_2024,multi-annotator-benthic_2015,durden_comparison_2016,schoening_semi-automated_2012}.

Turbidity especially complicates underwater image analysis. It is caused by suspended particles and organisms in the water column, which scatter and absorb light. As a result, underwater images often appear hazy or cloudy, with reduced contrast and blurred details.
As turbidity increases, both the scene and the objects within become increasingly difficult to perceive. This can create a compounding effect, where existing sources of disagreement and bias are likely to be exacerbated. 
High turbidity is very common and is typically observed in shallow, coastal waters, harbors, or during algal blooms. Despite the importance of reliable model performance in such environments, computer vision models are primarily trained and evaluated on data captured under favorable conditions with little turbidity \cite{sauder_coralscapes_2025, lian_watermask_2023, li_exploring_2026, islam_semantic_2020}.
Comparatively little attention has been given to data captured under high turbidity, possibly due to the inherent uncertainty involved in the annotation process \cite{pedersen_detection_2019, ismiroglou_beyond_2026, jahanbakht_semi-supervised_2023, jansi_rani_novel_2024}.
As a result, the quality of manual annotations and the extent of inter-annotator disagreement arising under degraded visual conditions in marine environments are not well understood.

In this work, we address this gap by studying the uncertainty associated with manually identifying and delineating objects in underwater imagery across varying levels of turbidity (\cref{fig:binary_masks_to_heatmap}). 
Our main contributions can be summarized as follows:

\begin{itemize}
    \item We conduct the first large-scale systematic study of human annotation in underwater environments with controlled levels of turbidity.
    \item We collect multi-user instance segmentation annotations in images together with per-user metadata capturing annotation behavior, confidence, device configuration, and participant characteristics.
    \item We compare crowdsourced annotations to each other and to gold-standard ground truth (GT) masks, investigating the effect of changes in visibility, privileged information, user experience, and user effort. We find that turbidity sets an upper bound on whether and how precisely objects can be identified and that its effects are non-linear, with some objects remaining discernible much longer than others.
    \item The collected annotations, metadata, and analysis scripts are made publicly available to support further research on underwater label noise and uncertainty.
\end{itemize}

\section{Related Work}

Humans are not always confident, consistent, or correct when labeling images. Noisy, missing, or biased labels are known to affect the performance of image recognition models as well as their evaluation~\cite{study-annotator-agreement-eval_2016,imagenet-labelling-issues-2020,test_set_errors-2021,labels-got-style_2023,imperfect-thesis,multi-annotator-dl-framework_2023}. However, the problem of incomplete and inaccurate annotations in computer vision is often studied by artificially degrading gold-standard labels~\cite{impact-annotation-error-seg_2022,effect-label-noise-remoteseg_2022,arctique-label-noise_2024,study-weak-noisy-annot_2025}. This allows the level and types of errors to be precisely controlled but does not provide insights into the complex mislabeling patterns occurring in real annotation tasks. 

Importantly, label noise is multidimensional. In \cite{sylolypavan_impact_2023}, authors identify four primary sources of labeling errors: insufficient information, lack of domain expertise, human slip-ups, and inherent subjectivity. Prior work also distinguishes between genuine annotation mistakes and acceptable human variation in labeling~\cite{weber-genzel_varierr_2024, uma_learning_2021}. Most multi-annotator studies capture noise as disagreement between annotators~\cite{tschirschwitz_kalos_2026, zhou_treasure_2023, schmarje_is_2022, rottmann_automated_2023} and few consider the uncertainty of individual annotators. The study presented in~\cite{gurari_investigating_2016} collected self-reported confidence from the annotators and found that annotators' perception of task ambiguity is influenced by their familiarity with the data and is misaligned with their actual performance. Overall, the way that mistakes and annotator disagreements present in practice, their prevalence, and their impact have received little attention in the field of computer vision and appear to depend strongly on the nature of the task and the dataset being annotated.

\subsubsection{Image Segmentation}
Systematic studies of manual image annotation are largely contained to the medical domain \cite{agreement-heatmaps-medseg_2023, yan_learning_2014, commowick_multiple_2021, gurari_how_2015}. For instance,~\cite{labeling-instructions-matter-biomed_2023} conducts a large-scale experiment investigating the effect of labeling instructions. In~\cite{what-can-we-learn-annotator-var-skin-seg_2026}, a large-scale study of intra- and inter-annotator variability is conducted in the context of skin lesion segmentation, showing a link between agreement and malignancy. However, sources and patterns of inter-annotator disagreement or uncertainty vary across tasks and modalities. 

Image segmentation introduces additional challenges beyond those encountered with label noise in classification tasks. In~\cite{agreement-heatmaps-medseg_2023}, a multi-annotator study on medical image segmentation shows that, even when annotators agree on the presence or absence of an object, substantial variability remains in the delineation of its boundaries. Furthermore, \cite{ribeiro_handling_2019} analyzes images annotated by at least two experts and reports a long tail of pronounced discrepancies, where differences between annotations can be so large that even determining which annotations to compare becomes non-trivial. Similar multi-annotator studies in the aerial imaging domain indicate that image complexity and object-boundary ambiguity significantly influence annotation variability \cite{kraff_uncertainties_2020, blushtein-livnon_performance_2025, joshi_vision_2025}.

In \cite{bauchwitz_task_2025}, the authors introduce the VACES dataset, composed of images sourced from widely used large-scale training datasets. Through a multi-annotator study with crowd-sourced labels collected under varying task configurations, they show that factors such as annotation tools, payment schemes, and image complexity significantly influence annotation outcomes.

\subsubsection{Low-Visibility Settings}
Reduced visibility can exacerbate annotator errors and disagreement; however, this issue has received relatively limited attention. Schumann et al.~\cite{schumann_consensus_2023} investigate the impact of challenging lighting conditions on skin color classification and show that humans can adapt to such variations. In \cite{hernandez-giron_low_2015}, annotators are asked to select, from a pair of low-contrast images, the one containing a predefined signal. Because the signal location is known in advance, this setup primarily evaluates limits of human perception rather than bias or ambiguity in annotation.

In computer vision domains where visibility is likely to degrade (e.g., adverse weather or low lighting in outdoor scenes), it is common practice to either circumvent manual annotation via synthetic data generation~\cite{synthia-dataset_2016,foggy-cityscapes_2018} or to leverage privileged information to aid manual annotation (e.g., normal-condition views of the same scene~\cite{acdc-dataset_2021,sakaridis_map-guided_2022} or adjacent video frames~\cite{erf-dataset-rain_2025}). However, existing work places little to no emphasis on understanding the impact of low visibility on the annotation process itself.

\subsubsection{Underwater Image Annotation}

Problems related to annotation quality have also been widely reported in the underwater domain \cite{joly_lifeclef_2026, lucas_underwater_2025, awad_ruod-r_2026, schoening_recomiarecommendations_2016}. Due to factors such as challenging environmental conditions, reduced visibility, annotation bias, and unclear labeling guidelines, existing datasets often contain incomplete or incorrect annotations. As a result, errors in evaluation data can lead to misleading assessments of model performance.
Some multi-annotator studies have demonstrated variability in human labeling \cite{wyatt_signal_2025, durden_comparison_2016, elsaser_seagrassfinder_2025}, but the full data, including both images and labels from all annotators, are not publicly available, limiting further investigation of the contributing factors. Efforts such as JAMBO~\cite{jambo_2024} and BenthicSurvey~\cite{multi-annotator-benthic_2015} provide publicly available multi-annotator datasets. However, they are largely restricted to image-level annotations and offer limited insight into the annotation process itself. \textit{To the best of our knowledge, there is currently no publicly available multi-annotator segmentation dataset in the underwater domain.}

\section{Annotation Study}
We are interested in the individual uncertainty and inter-annotator disagreement arising when manually annotating objects of interest in underwater scenes. We hypothesize that turbidity-induced image degradation amplifies annotation errors and uncertainty, and thus we systematically investigate the impact of turbidity on human annotators. For this study, we crowdsource annotations through a custom-made annotation tool deployed as a public-facing interactive web application.

In this section, we describe the dataset and annotation-collection protocol, as well as the design, functionality, and underlying rationale of the web application.

\begin{figure}[tb]
    \centering
    \includegraphics[width=0.93\linewidth]{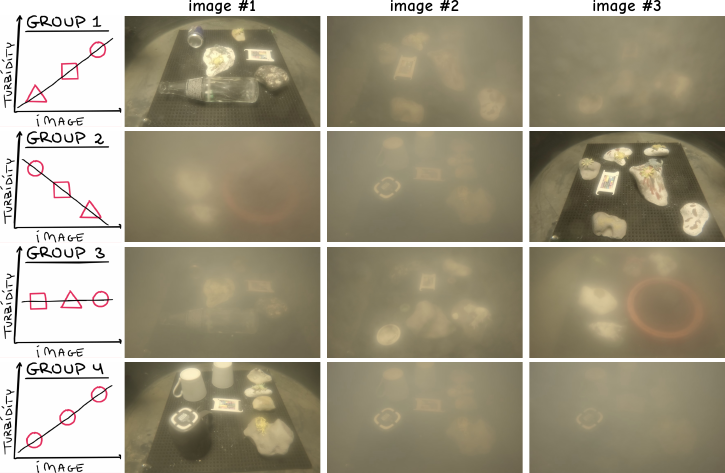}
    \caption{Scenes are selected randomly for every user, while ensuring that \groupone, \grouptwo, and \groupthree ~see three different scenes. \groupfour ~see the same (random) scene but with increasing turbidity levels. We show example scenes here for illustration.}
    \label{tab:participant-groups-imgs}
\end{figure}

\subsection{Dataset}\label{sec:annstudy:subsec:dataset}

As a basis for studying uncertainty in annotations of underwater images, we use the Turbid Underwater Baseline (TUB) dataset \cite{ismiroglou_beyond_2026}.
TUB is currently the only publicly available dataset that provides labeled images of identical static scenes captured under a controlled range of real turbidity conditions (i.e., without the use of synthetic data). 
It therefore enables the isolation of turbidity as a key source of uncertainty.
TUB includes instance segmentation masks for various object categories, such as rocks, trash, and mugs.
The objects were originally annotated in clear images and subsequently propagated to the corresponding turbid images.
Therefore, we treat these annotations  as reliable, high-confidence golden ground truth labels (GT) throughout the evaluation.

Since we aim to obtain a large number of crowd-sourced annotations per individual image, we restrict our study to 5 static scenes from the TUB dataset. For each scene, we select a low, medium, and high turbidity image based on visual inspection, resulting in a total of 15 unique images. We refer to Appendix~\ref{app:studydesign} for details about image selection.

\subsection{User groups}\label{sec:annstudy:subsec:groups}

We define four participant groups, each following a distinct image-sequence protocol. All groups are presented with three images from any of the five scenes described in \cref{sec:annstudy:subsec:dataset}. However, the levels of turbidity and their presentation order vary across groups, as summarized in \cref{tab:participant-groups-imgs}. This design enables two complementary analyses. First, by exposing participants to varying turbidity levels, we assess the impact of turbidity on annotation quality and perceived uncertainty. 
Second, we examine whether participants who are first exposed to a clear image, thereby forming an expectation of the scene layout, annotate subsequent turbid images differently compared to those who encounter the turbid version first. \groupfour ~represents an edge case with the highest level of privileged information: all presented images correspond to the exact same scene.

All participants receive identical task descriptions and guidelines regardless of their group. Furthermore, all participants are presented with the same practice page and example image before starting the main annotation task, which are described below.

\subsection{Study sequence and annotation task}\label{sec:annstudy:subsec:sequence}
\begin{figure}[tb]
    \centering
    \includegraphics[height=0.3\linewidth]{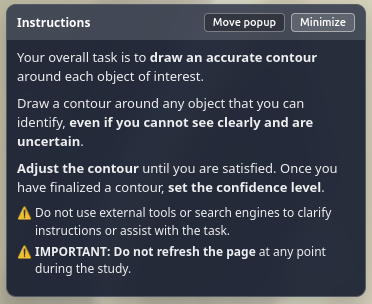}\hfill
    \includegraphics[height=0.3\linewidth]{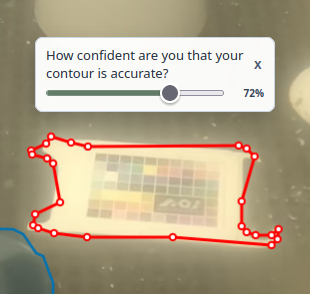}\hfill
    \includegraphics[height=0.3\linewidth,trim={0 2.5cm 0 6cm},clip]{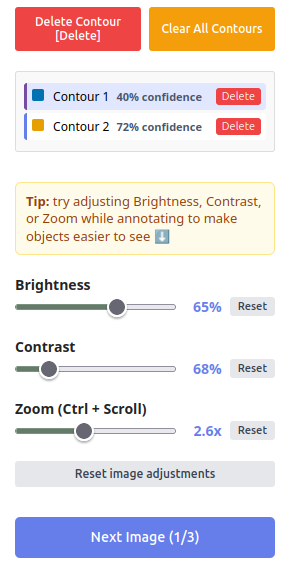}
    \caption{Screenshots of different parts of the annotation interface.}
    \label{fig:annotation-interface-screenshots}
\end{figure}

We designed a custom web-based annotation tool for the purpose of this study, which gives us control over the instructions and functionalities available to the user in the front-end, as well as the data collection in the back-end. The link to the study was shared widely through personal networks, professional channels, and on social media, with users participating on a voluntary basis. People were encouraged to participate regardless of their background or experience level. The study consists of four stages:

\begin{enumerate}
    \item \textbf{The front page} provides a brief description of the task and device requirements and includes a consent form outlining all stored data.
    \item \textbf{A practice page} is loaded after a participant agrees to start the study. This page introduces the user to the annotation-tool interface, where a partly annotated example image is shown (this image is not related to the TUB dataset cf. Appendix~\ref{app:practice-page}). The participants are then instructed to experiment with the tool by adding or editing annotations. The annotation tool enables users to create instance segmentation masks through freeform drawing. Freeform drawings are transformed into vector-based polygons, and the user can subsequently move, add, or delete individual points or entire masks.
    For every mask, users are prompted to indicate their confidence level using a slider ranging from 0 to 100\% (\cref{fig:annotation-interface-screenshots}).
    Image brightness, contrast, and zoom can be adjusted via sliders to potentially improve visibility of the objects.
    \item \textbf{The main task} begins once users have chosen to proceed from the practice image. The annotation-tool interface remains the same, but the images are now 3 images from the TUB dataset, selected and shown in a specific order, based on the automatically assigned group (\cref{sec:annstudy:subsec:groups}). 
    Annotation is performed one image at a time, and users cannot return to previous images once they have submitted their final annotations for an image, a fact that is highlighted through a popup when they choose to continue. There are no limits to the number of polygons, with 0 annotations being a valid submission. However, participants cannot proceed to the next image if any polygons have missing confidence scores. Instructions along with explanations of the controls remain visible throughout the study, and brightness/contrast/zoom slider values are reset for every image.
    \item \textbf{A short questionnaire} is presented after participants complete the annotation task. The questionnaire is designed to assess participants’ annotation expertise and their comfort with the tool throughout the process. No personally identifiable information is collected. At the end of the questionnaire, participants are asked to retain their unique ID in case they wish to request the deletion of their submission in the future. The full questionnaire is provided in Appendix~\ref{app:collecteddata}.
\end{enumerate}

In addition to saving the annotations submitted for each image, a logging system was implemented to enable detailed analysis of user interactions. Specifically, events were recorded whenever a participant created an annotation, created/modified/deleted a polygon or point, or adjusted any of the sliders.

\section{Analysis}

In total, 104 participants completed the study in its entirety, with 25-27 participants per group and 11-36 unique annotations per image. We obtained answers from users with a diverse pool of backgrounds, with over half having little to no experience with machine learning and 21\% who had annotated image segmentation datasets before.
Five participants were highly familiar with the dataset, 15 participants had previously seen images from the dataset, while 84 had never seen any TUB-related images prior to the study. Detailed participant statistics can be found in Appendix~\ref{app:userstats}, along with the participants' individual annotations (practice image and 3 TUB images) in Appendix~\ref{app:indiv_annot}.

\subsection{Evaluating annotations}

\begin{figure}[t]
    \centering
        \includegraphics[width=\linewidth]{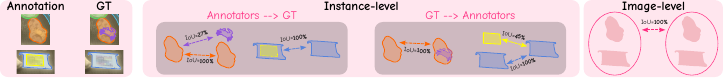}\caption{The matching protocols used when evaluating the collected annotations.}
        \label{fig:matching-diagram}

\end{figure}

We evaluate annotations from several angles, comparing them to:
\begin{enumerate}
    \item The \textbf{GT} at the \textbf{instance-level}, via IoU-based matching between annotated vs. GT object contours (cf.~\cref{fig:matching-diagram}). Given an annotated image, \textit{annotators}$\rightarrow$\textit{GT} is when every GT contour gets exactly 1 match. It evaluates whether GT objects were correctly identified, penalizing False Negatives (FN). \textit{GT}$\rightarrow$\textit{annotators} is when every annotated object gets exactly 1 match. It evaluates whether annotated objects should have been annotated, penalizing False Positives (FP). For both matching directions, we report the match IoU per object directly to avoid setting an arbitrary threshold.
    \item The \textbf{GT} at the \textbf{image-level}, treating the annotations and GT as binary segmentation masks. We report the segmentation IoU, precision, and recall per image.
    \item Other \textbf{users' annotations} at the \textbf{image-level} via \textbf{pairwise segmentation IoU} (similarly to~\cite{joshi_vision_2025,what-can-we-learn-annotator-var-skin-seg_2026}) to measure inter-annotator agreement.
\end{enumerate}
Additional details can be found in Appendix~\ref{app:evalmetrics}. Unless otherwise specified, we consider all annotated contours in our analysis, regardless of confidence (confidence threshold of $T=0$). In \cref{sec:analysis-confidence} we analyze the role of confidence.

\begin{figure}

        \centering
        \includegraphics[width=\linewidth]{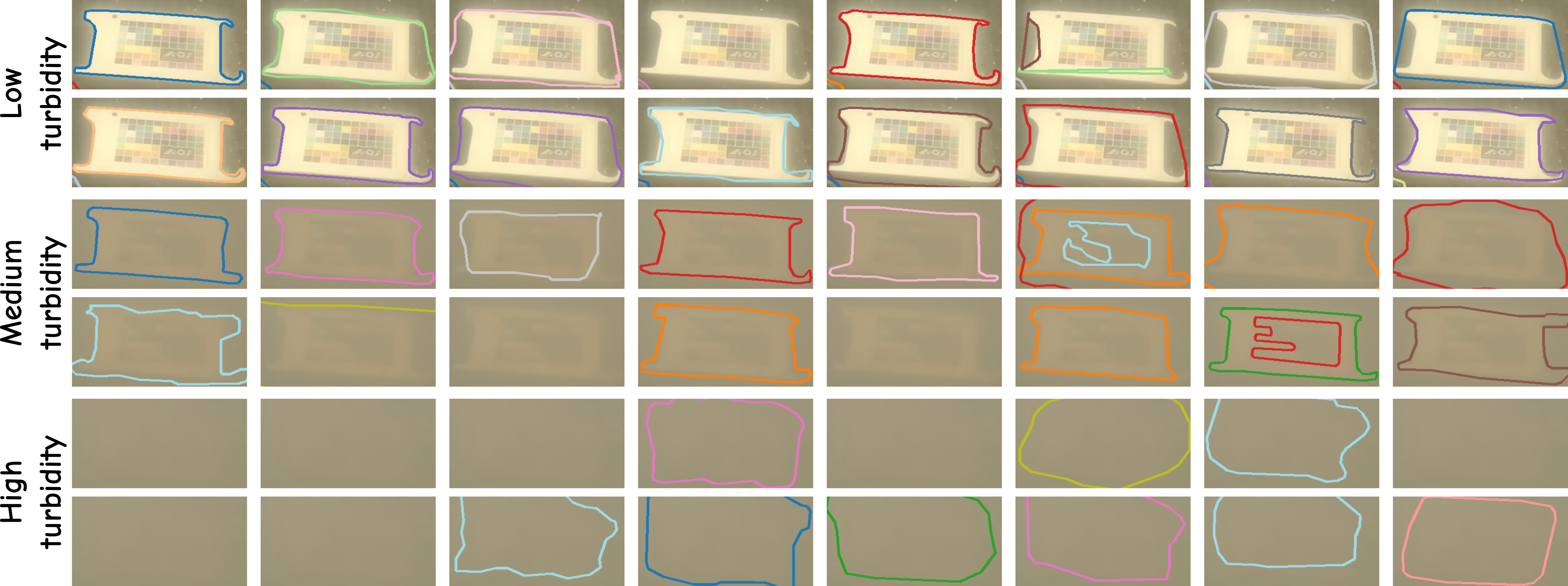}

        \caption{Randomly selected annotations for the same object in the same scene (\texttt{Exp30\_Cam1}), varying turbidity. Some users fail to annotate the object altogether, especially as turbidity increases.}
        \label{fig:checkerboard-annots}
    
\end{figure}

\subsection{Label noise and disagreement}

\subsubsection{Variation across users:} 
\begin{figure}[tb]
    \centering
    \includegraphics[width=0.2\linewidth]{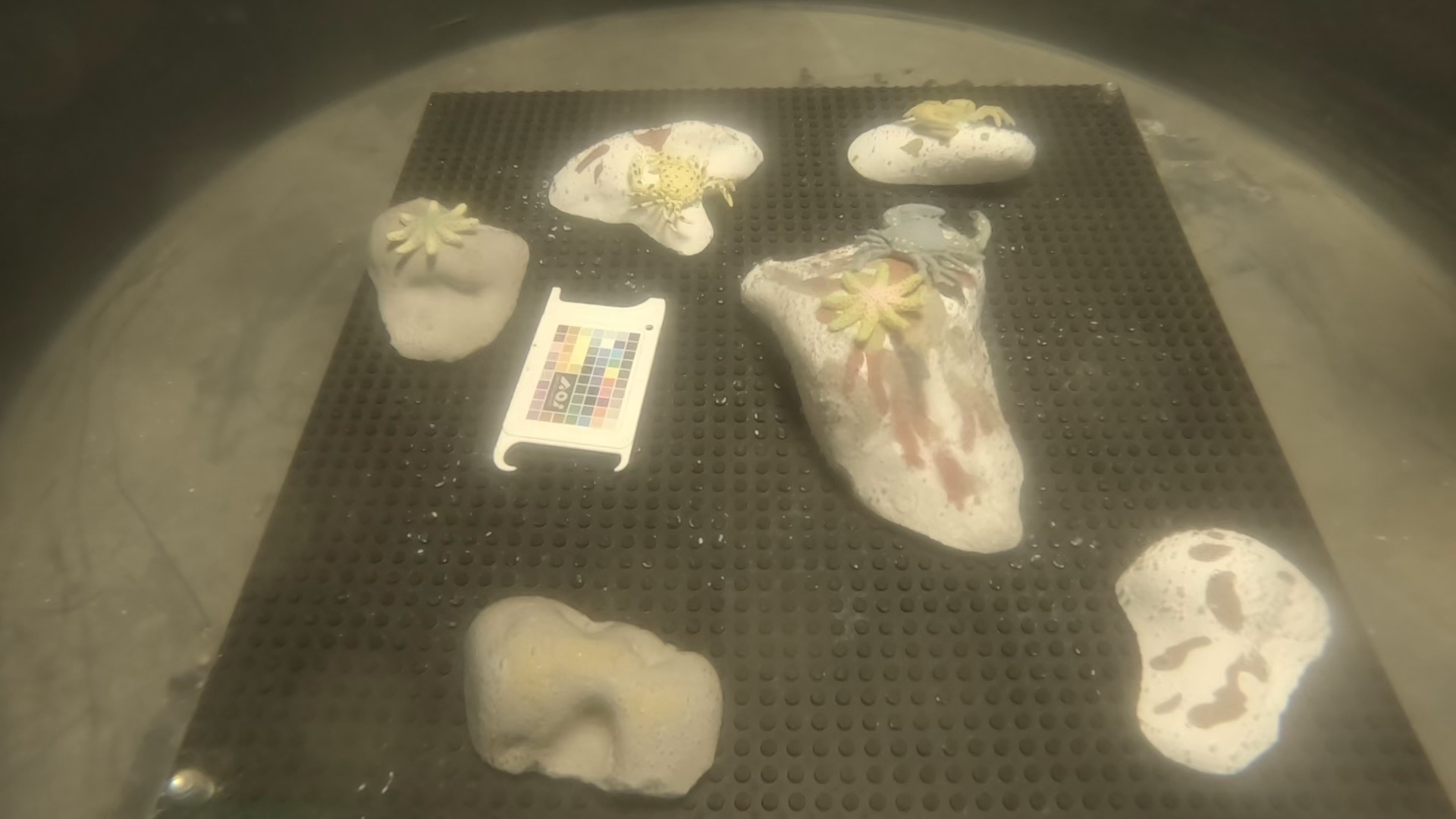}\includegraphics[width=0.2\linewidth]{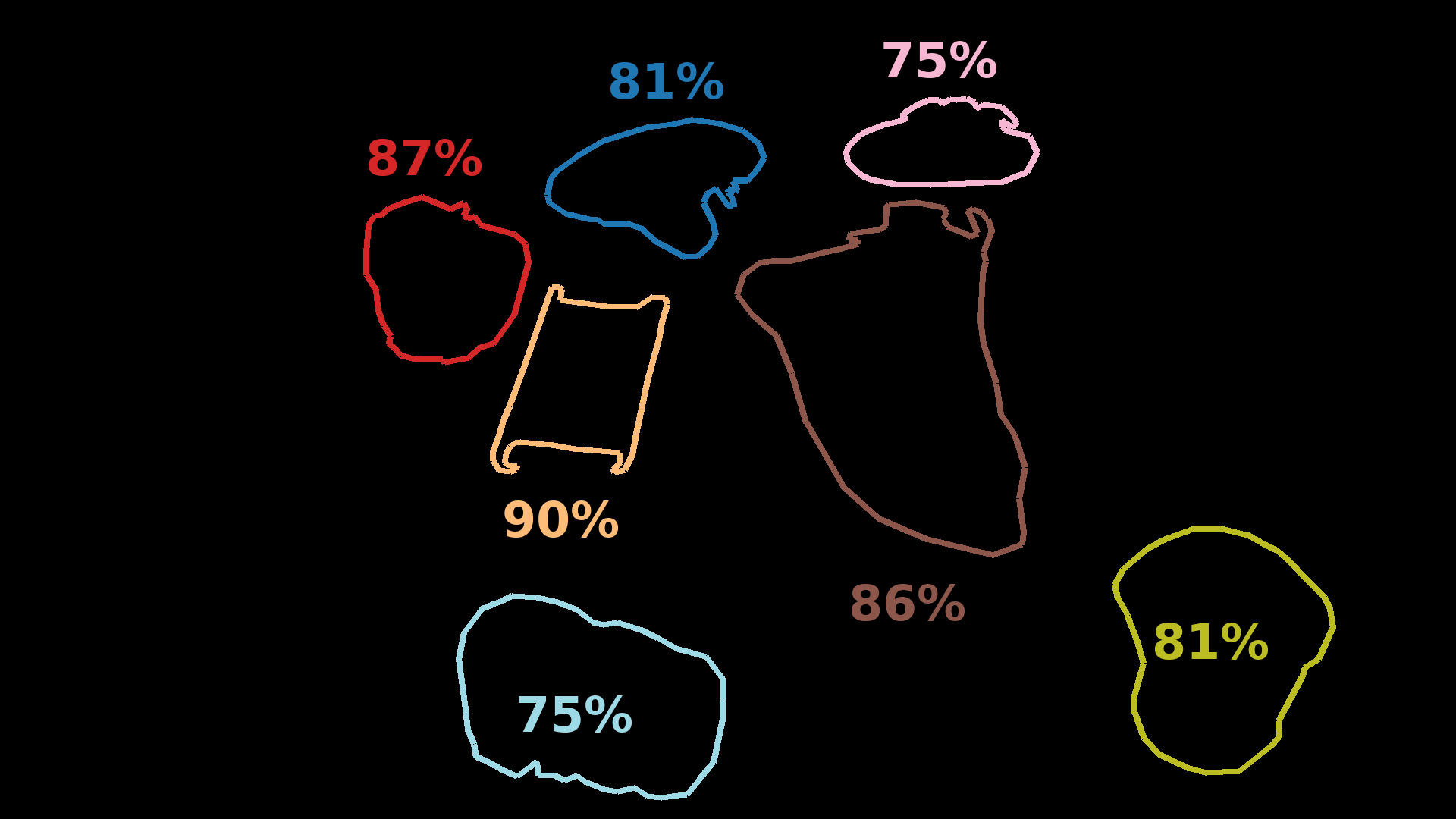}\includegraphics[width=0.2\linewidth]{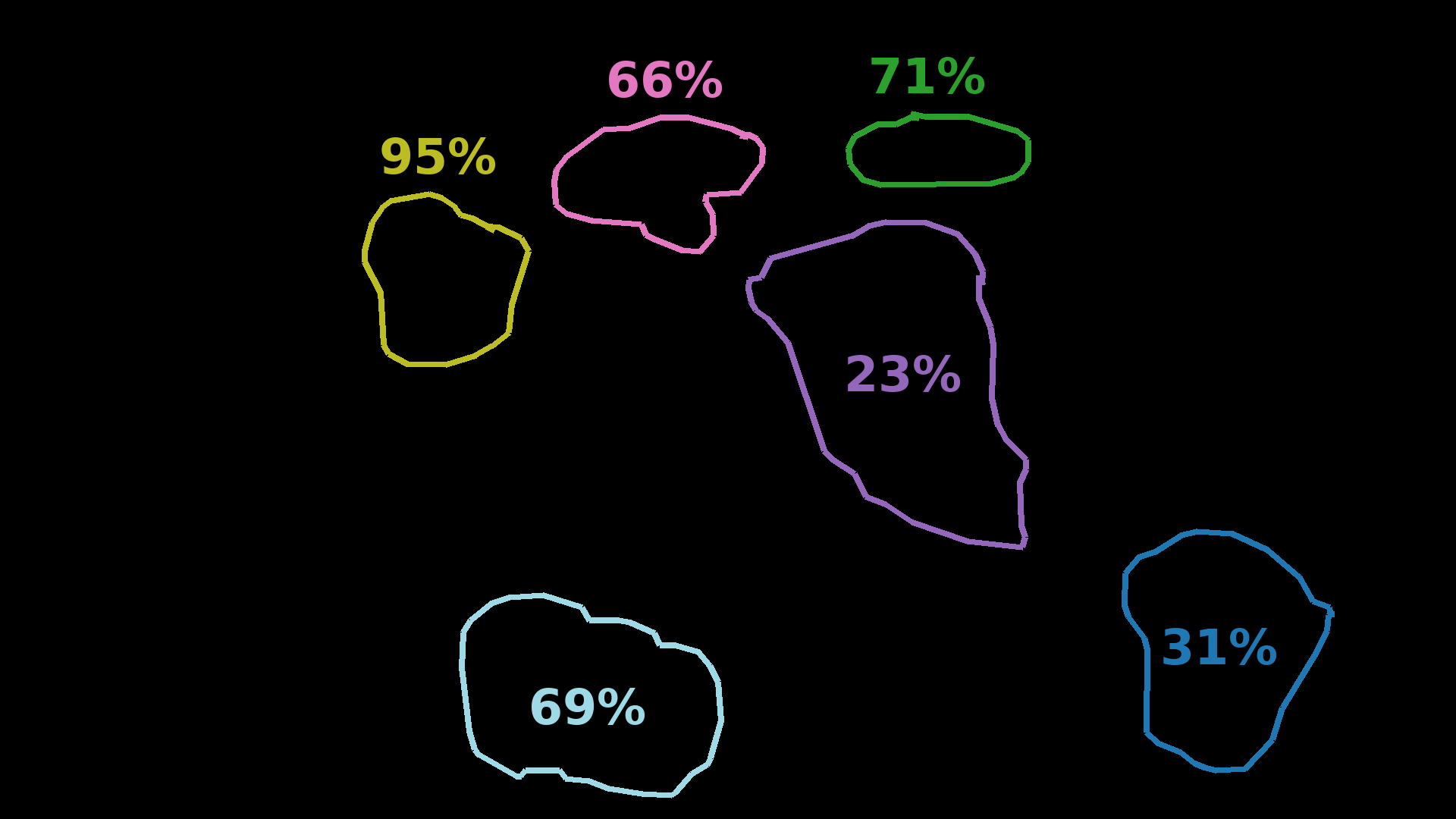}
    \includegraphics[width=0.2\linewidth]{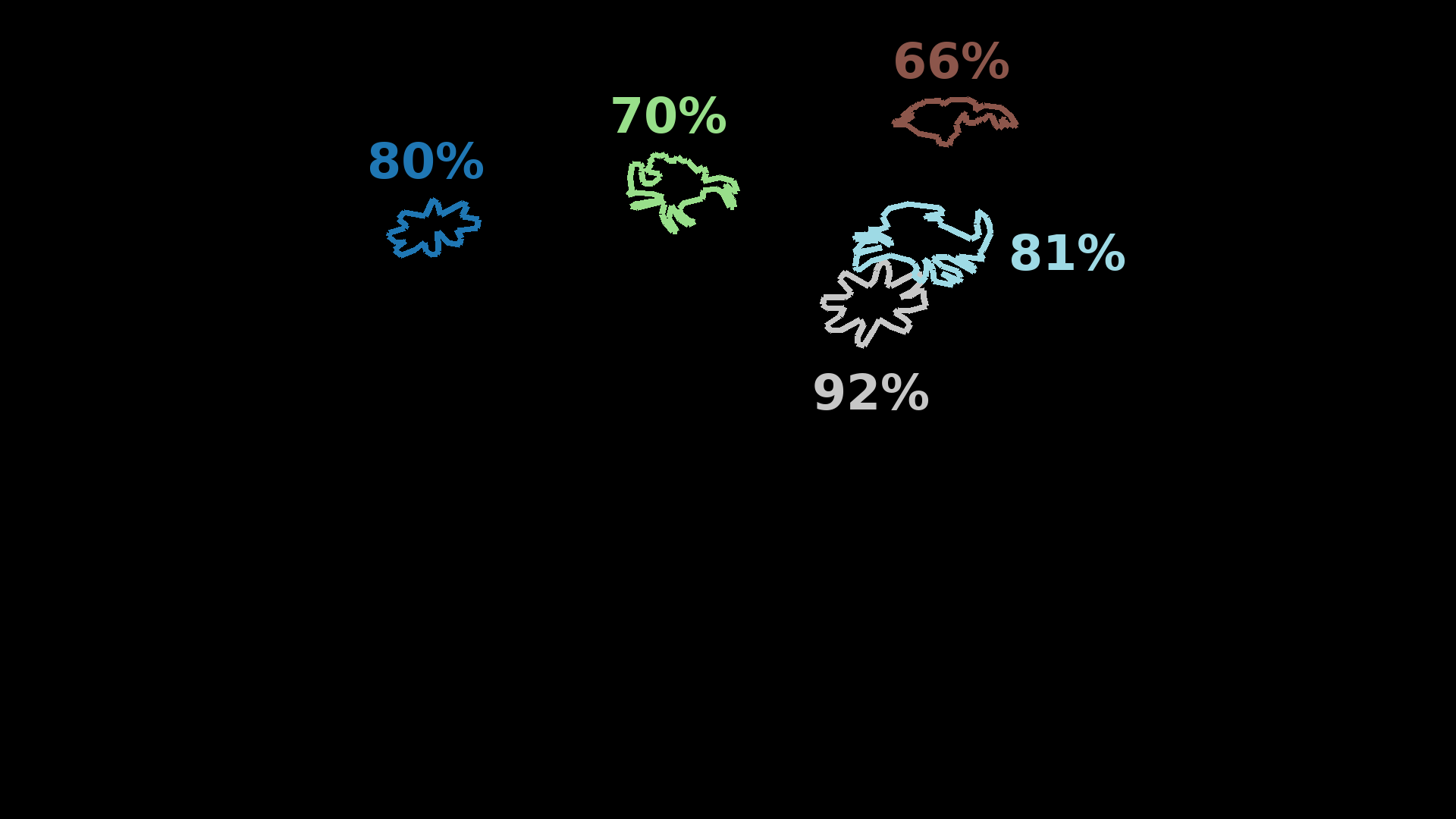}\includegraphics[width=0.2\linewidth]{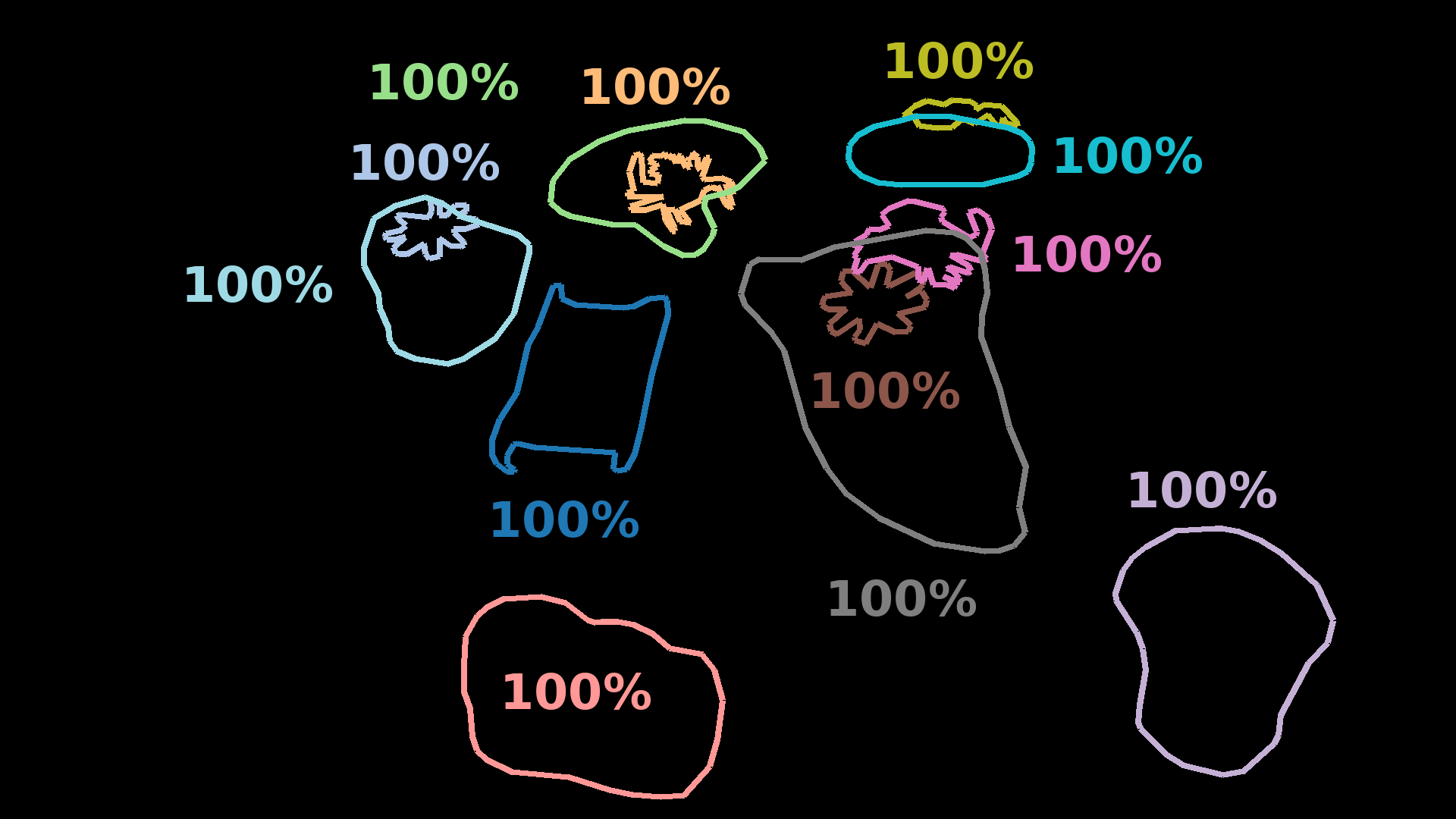}

    \includegraphics[width=0.2\linewidth]{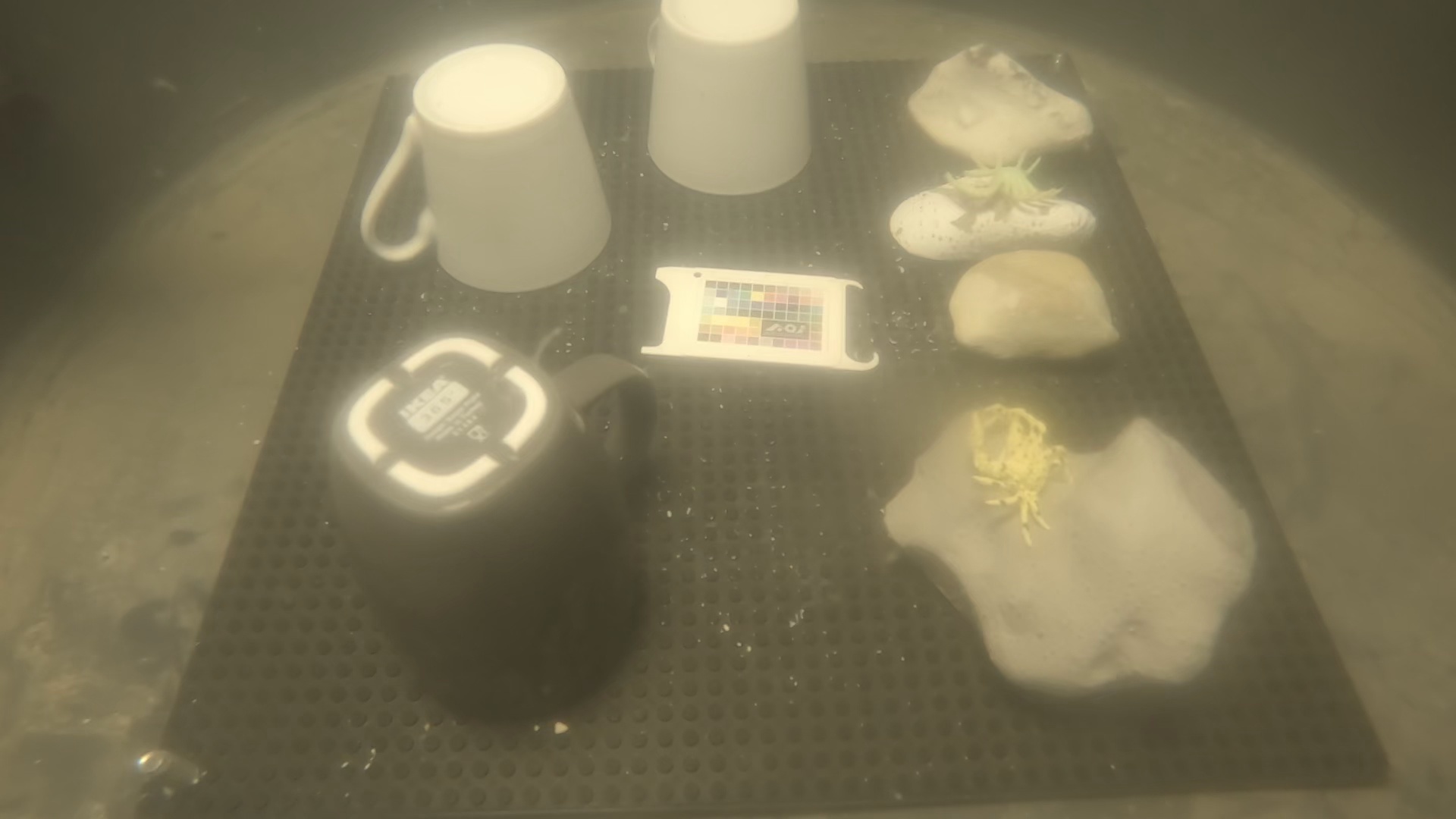}\includegraphics[width=0.2\linewidth]{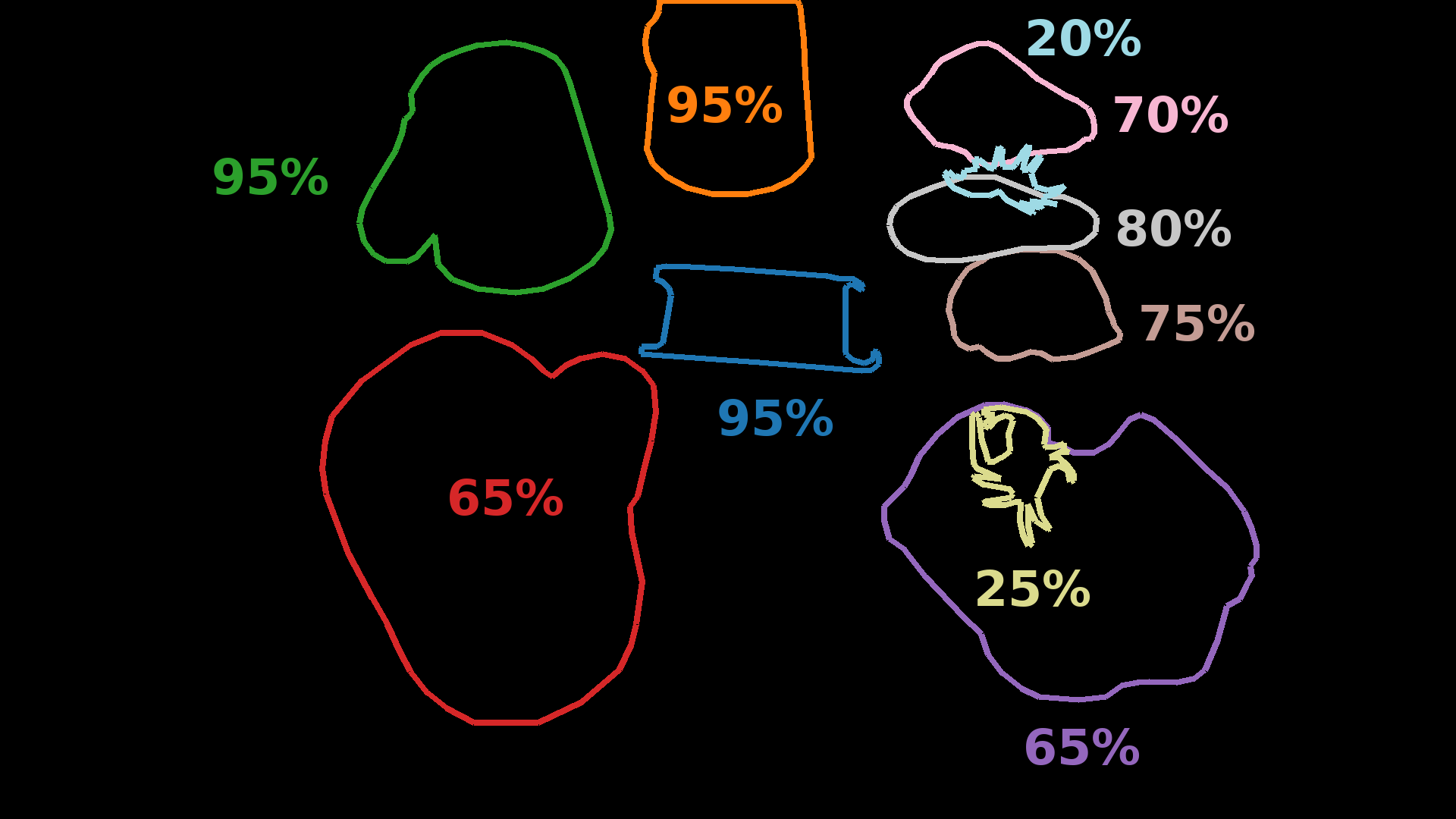}
    \includegraphics[width=0.2\linewidth]{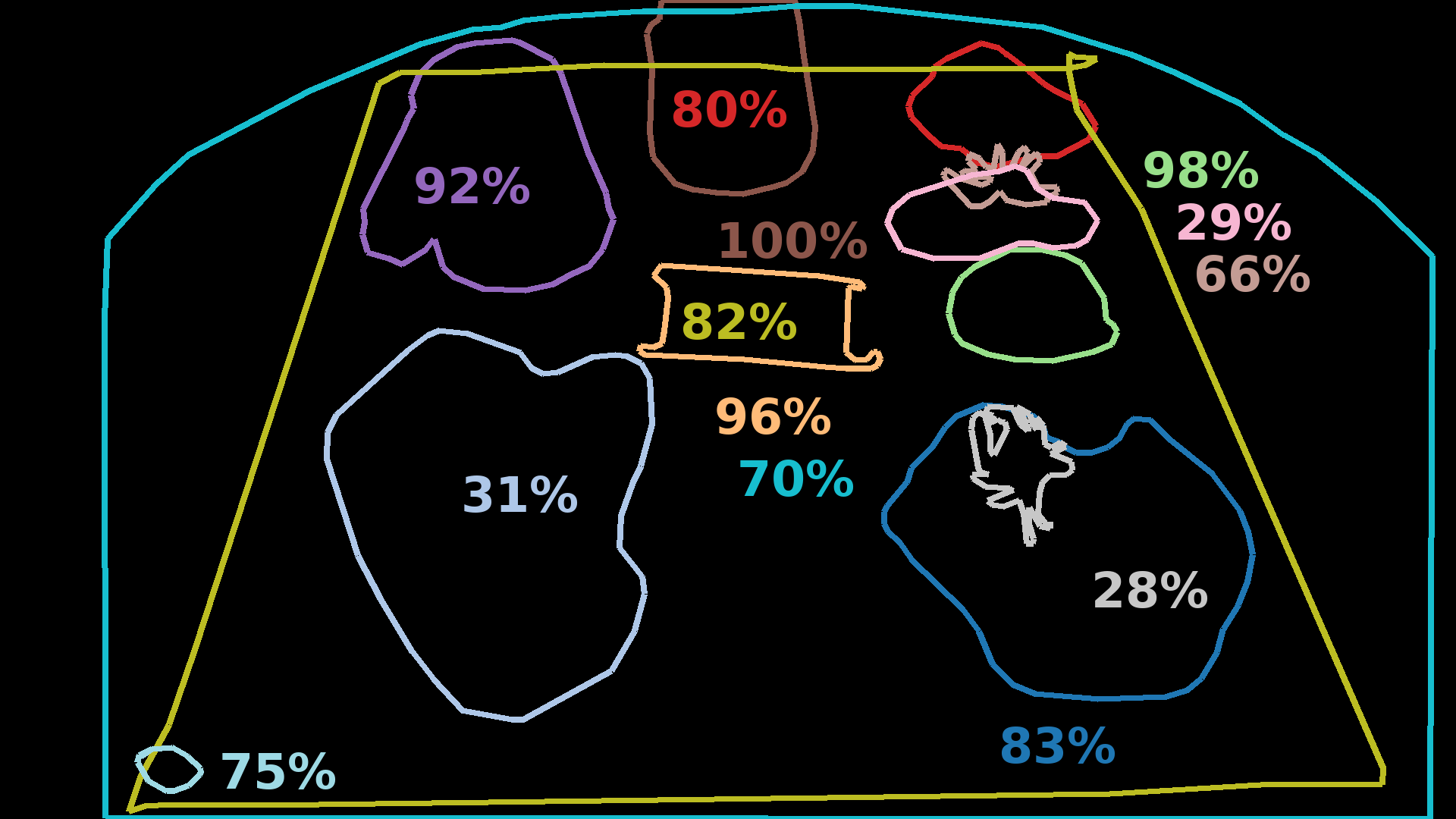}\includegraphics[width=0.2\linewidth]{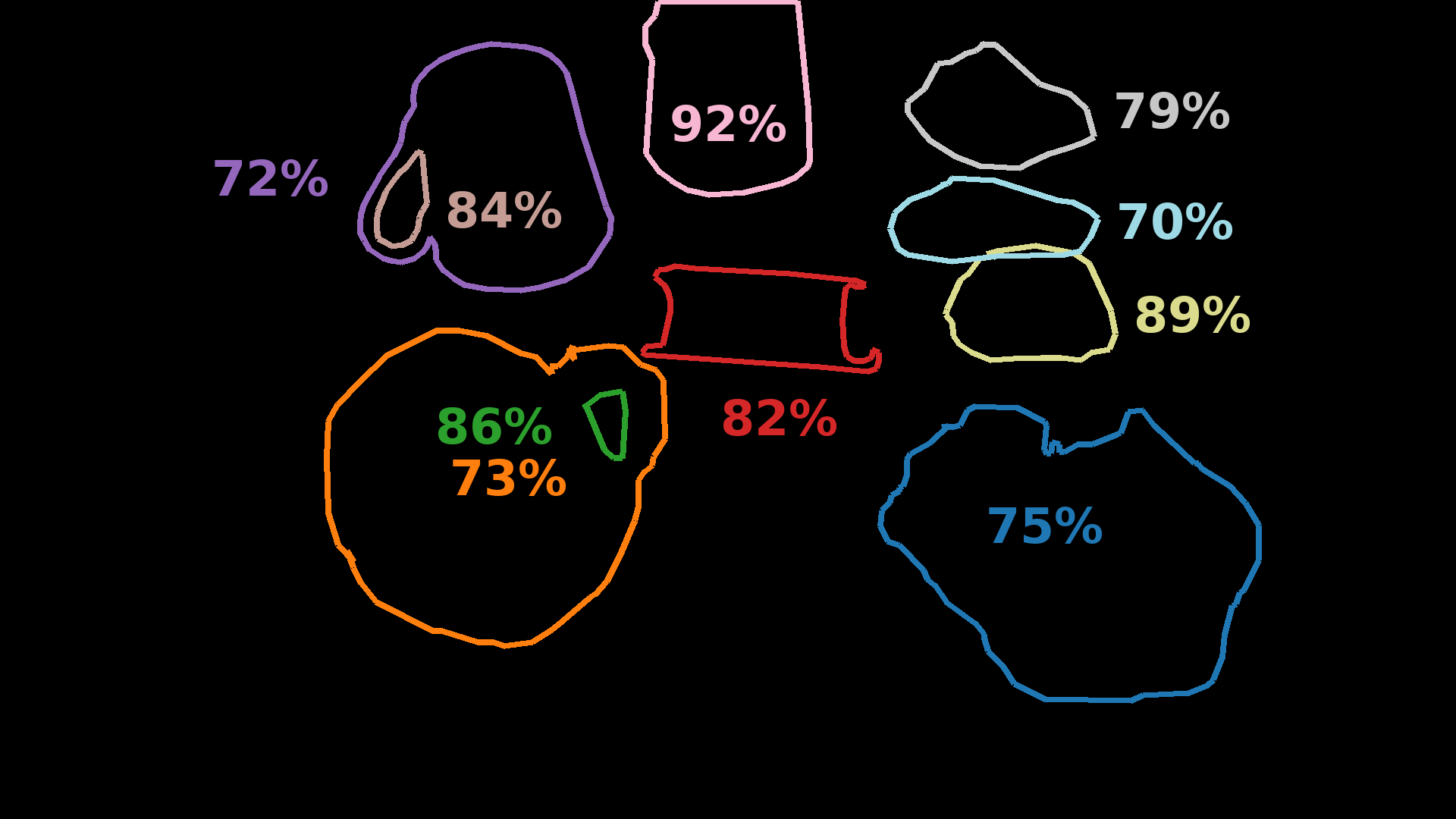}\includegraphics[width=0.2\linewidth]{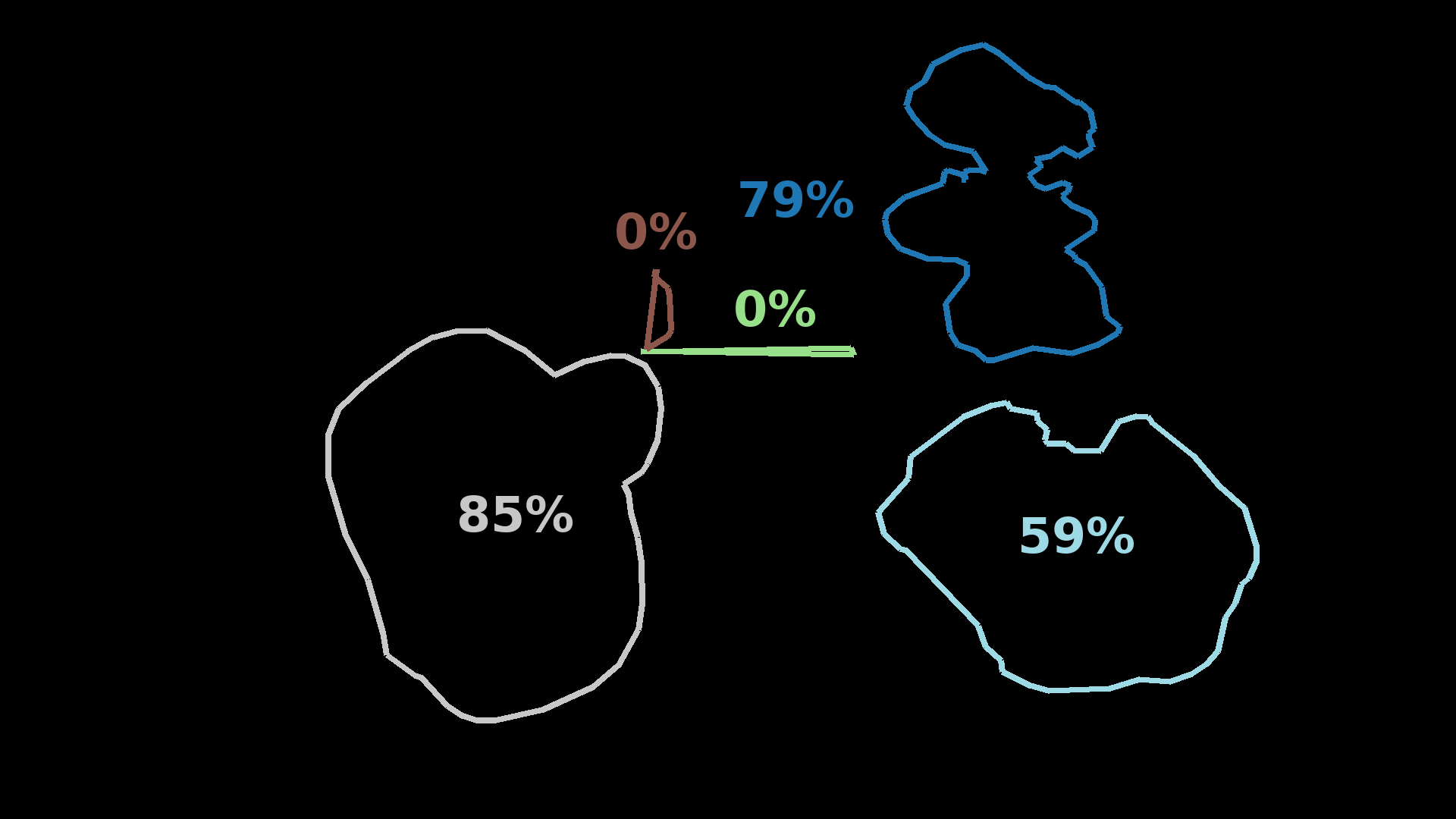}
    \caption{Clear-image annotations (object contours and confidence), selected via visual inspection to showcase inter-annotator diversity. More examples can be found in Appendix~\ref{app:indiv_annot}.}
    \label{fig:clear_annot_examples}

\end{figure}

Examining individual annotations in clear waters reveals substantial variation in how users interpret and perform the task, even when object visibility is high. Across the five scenes, we observe disagreement in several aspects: (1) the precision with which object contours are drawn, (2) whether holes within objects should be annotated (e.g., mug handles), (3) whether adjacent or overlapping objects should be treated as a single object, and (4) which objects are considered relevant enough to annotate in the first place. \cref{fig:checkerboard-annots} shows the diversity in annotation styles for a single object across different turbidity levels, while \cref{fig:clear_annot_examples} highlights the variation in full-scene annotations under clear-water conditions.

Taking a closer look at FP annotations, we find that, among the 79 users who annotated clear images, 12 chose to label the LEGO baseboard as an object, with 3 of these users additionally annotating the circular rim of the tub. A further 4 users included one or more of the small screws holding the baseboard. While such annotations are arguably reasonable, they are (perhaps unfairly) penalized when compared with the GT. In contrast, some users produced a large number of FN annotations by omitting objects that were clearly visible. For example, one user annotated only the largest rock in the scene, while two users exclusively annotated the plastic sea creatures - presumably because these objects appeared more "of interest", or representative of a typical underwater scene.

Erroneous groupings of multiple objects into a single annotation range from not annotating a seashell or crab stuck to the middle of the rock (quite frequent, understandable) to drawing a single contour around several neighboring rocks (somewhat questionable, much less frequent). These many-to-one groupings are not penalized by semantic segmentation metrics but negatively affect instance segmentation metrics.

\clearpage
\subsubsection{Effects of turbidity.}

\begin{figure}[tb]
\centering
    \includegraphics[width=\linewidth]{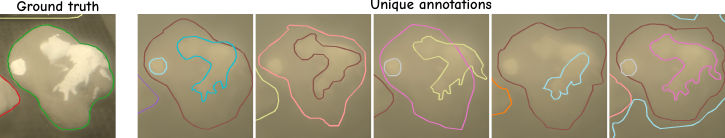}
    \caption{Example from a medium-turbidity scene (\texttt{Exp6\_Cam1}), where annotators separately contour different parts of the same rock. We do not observe this phenomenon in low-turbidity annotations. In high-turbidity images, annotators are more likely to roughly annotate the whole rock, miss it altogether, or only annotate the bright spots.}
    \label{fig:mediumturbid-rock}
\end{figure}

From qualitative comparisons like the one shown in \cref{fig:checkerboard-annots}, we observe a clear decline in annotation quality under turbidity-induced visibility degradation. When the object is clearly visible, roughly one-third of annotators draw meticulous contours. This proportion decreases under medium turbidity, despite some annotators remaining impressively thorough. While turbidity generally increases the number of FNs, interestingly, under medium turbidity, annotators sometimes significantly overestimate the number of objects present in certain areas. These are primarily one-to-many errors due to perceiving object sub-parts as separate objects (e.g., colorchecker patterns, bright spots on a rock) - cf.~\cref{fig:mediumturbid-rock,fig:checkerboard-annots}. In high turbidity, pixel-precise contours are an exception, only occurring for a small subset of the most salient objects.

\begin{figure}[tb]
    \centering
    \begin{minipage}{.51\textwidth}
    \centering
    \includegraphics[height=3cm]{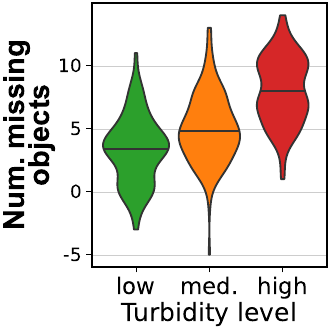}\hfill
    \includegraphics[height=3cm]{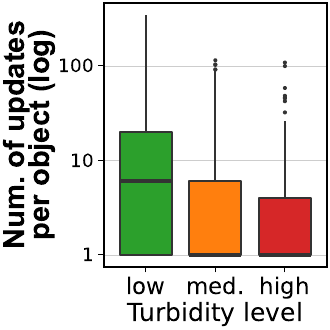}
    \caption{\textbf{Left:} Difference between the number of GT objects and the number of annotated objects (per user, per image). \textbf{Right:} User effort per object excluding deleted polygons.}
    \label{fig:object-num-turbidity}
    \end{minipage}\hfill
    \begin{minipage}{.46\textwidth}
    \centering
    \includegraphics[height=3cm]{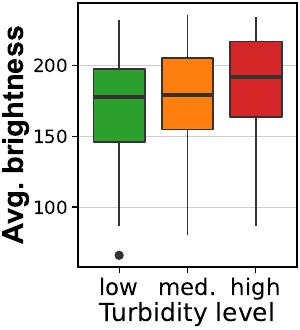}\hfill
    \includegraphics[height=3cm]{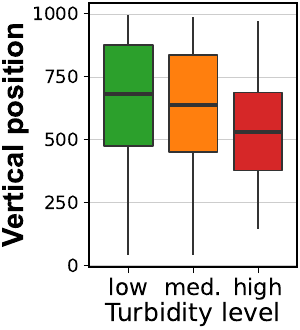}
    \caption{Appearance and location properties of annotated objects. The brightness is taken in reference to the clear image, and the vertical position is in pixels from the bottom.}
    \label{fig:object-properties-turbidity}
    \end{minipage}
\end{figure}

\begin{figure}[t!]
    \centering

    \centering
    \includegraphics[height=3cm]{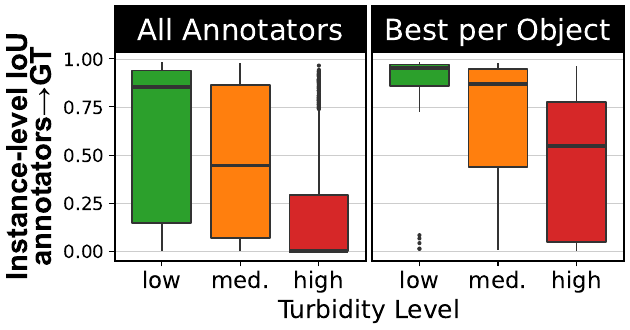}
    \hspace{2em}
    \includegraphics[height=3cm]{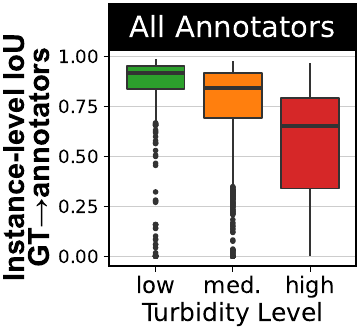}
    
    \caption{\textbf{Instance segmentation} IoU for different turbidity levels (one point per object), matching \textit{annotators}$\rightarrow$\textit{GT} (penalizing FNs, but not FPs) or \textit{GT}$\rightarrow$\textit{annotators} (penalizing FPs, but not FNs).}
    \label{fig:instance-perf-per-turbidity}

\end{figure}

These observations are also reflected quantitatively. \cref{fig:object-num-turbidity} shows that as turbidity increases, users miss more objects and make fewer instance-level adjustments. Objects that are brighter or located closer to the camera (appearing lower in the image) are more likely to be annotated, as illustrated in \cref{fig:object-properties-turbidity}. The highest-IoU contour in each of the five high-turbidity images ranges from 89\% to 97\%, and notably, almost always corresponds to objects located in the lower half of the image (cf. Appendix~\ref{app:results-qual}).

Finally, in~\cref{fig:instance-perf-per-turbidity}, we quantify the effect of turbidity on annotations in terms of instance-level segmentation performance. The substantial variation across annotators and objects, even for clear images, suggests that much of the label noise arises from factors unrelated to turbidity or visual quality. Annotation error is greatly reduced when only the best annotator per GT object is considered (middle, median IoU of 95.2\%) or when non-annotated objects are not penalized (right, median IoU of 91.5\%). Nevertheless, the fact that near-perfect performance across \textit{all} GT objects cannot be achieved in low turbidity highlights limitations in the task definition itself and the GT used for comparison. Orthogonally, turbidity introduces an additional systematic effect, with the quality of even the best matches deteriorating significantly as visibility decreases. The increased variance in the best-case high-turbidity annotations further confirms that the impact of turbidity is not uniform across objects and image regions.

\subsection{What helps and doesn't help?}

Given inter-annotator variability even in clear scenes and the compounding detrimental effect of turbidity, we investigate potential strategies for improving annotation quality.

\subsubsection{Can we filter out unreliable annotations?}\label{sec:analysis-confidence} Intuitively, user-reported confidence could be used to identify unreliable annotations at the individual level. At the same time, assigning a pool of users to the same image and looking at their annotations in aggregate, as an \textit{ensemble}, gives us a complementary measure of uncertainty/confidence: which proportion of users agree that a given pixel belongs to an object? 

\begin{figure}[t!]
    \newcommand{\emptyline}{\phantom{\tiny \textbf{Heatmap4}}}
    
    \centering
        \begin{tabular}{ccc}
            {\comicneue \scriptsize \textbf{(A) All individual confidence masks}}  &  {\comicneue \scriptsize \textbf{(B) Best users confidence masks}} &  {\comicneue \scriptsize \textbf{(C) Ensemble from binary masks}} \\
            \includegraphics[height=2cm,trim={0 0 3cm 0},clip]{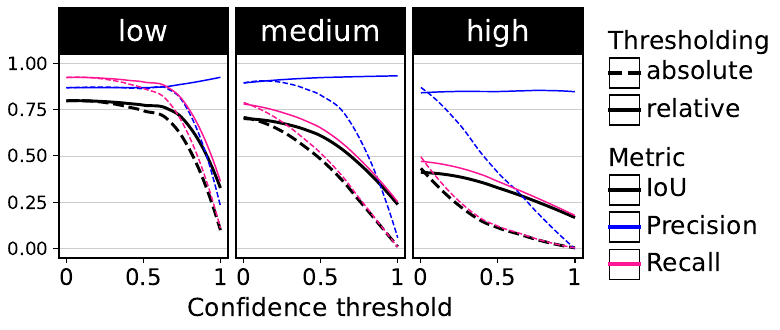} & \includegraphics[height=2cm,trim={0 0 3cm 0},clip]{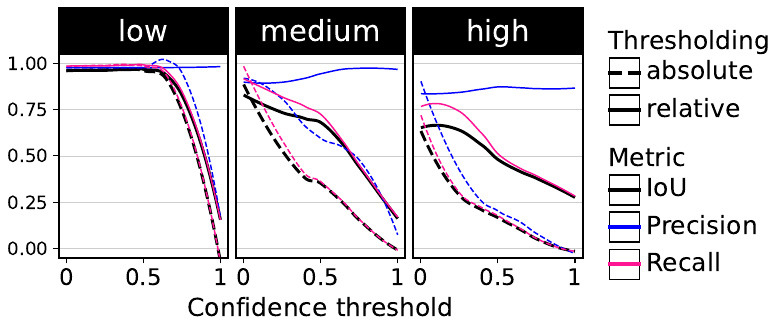} & \hfill\includegraphics[height=2cm]{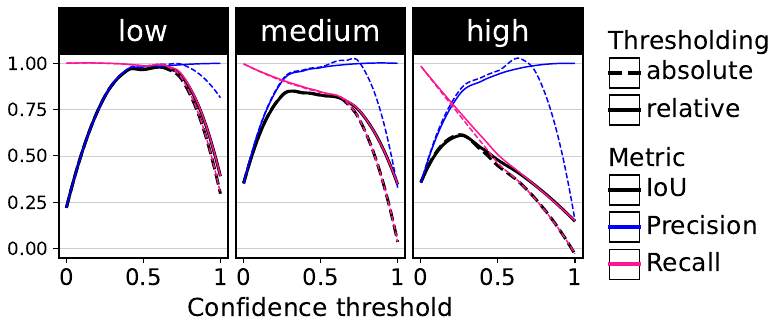}  \\
        \end{tabular}
    \hfill
    \caption{Average \textbf{semantic segmentation} performance across individual or ensemble masks, as a function of the confidence threshold. (B) only keeps the best annotator per image. With absolute thresholding, a pixel with confidence $c$ is considered an object if $c\geq T$. With relative thresholding, a pixel is considered an object if $c \geq c_{max} \cdot T$, where $c_{max}$ is the maximum confidence in the image.}
    \label{fig:threshold-conf-curves}

\end{figure}

We first evaluate these two uncertainty measures in relation to segmentation performance in~\cref{fig:threshold-conf-curves}, varying the threshold from 0 (all annotated objects included in the binary mask) to 1 (only the most confident contours included). Across individual annotators (A), no single threshold (whether absolute or relative) outperforms simply taking all annotations ($T=0$). Only considering the top-annotator per image (B) brings significant performance gains, but increasing the threshold still provides little to no benefit, with many valid annotations being discarded. This indicates that individual annotator confidence is not a reliable signal for whether an object should have been annotated. In contrast, confidence arising from inter-annotator agreement (C) is a much more informative signal, matching or outperforming top-annotators with the optimal choice of threshold (peak IoU). However, due to annotators systematically reporting lower confidence under increasing turbidity, the optimal threshold inevitably shifts as visibility degrades and thus should not be set uniformly across turbidity levels.

\begin{figure}[t!]

    \centering
        \includegraphics[height=3.5cm]{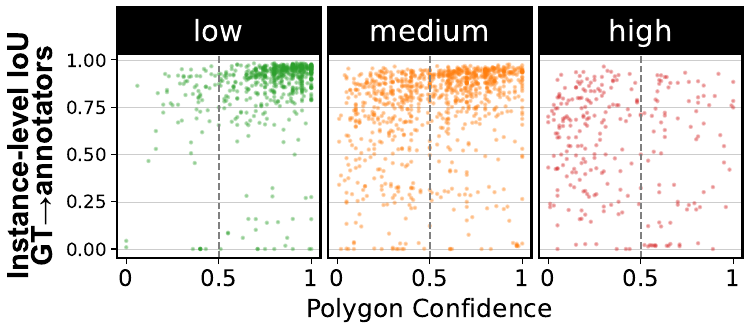}
    \caption{\textbf{Instance segmentation} IoU of individual annotated objects (FNs not penalized) and their corresponding confidence. See Appendix~\ref{app:results-plots} for a binned version.}
    \label{fig:threshold-conf-curves-instance}
\end{figure}

\begin{figure}[t!]
    \centering
    \includegraphics[width=\linewidth]{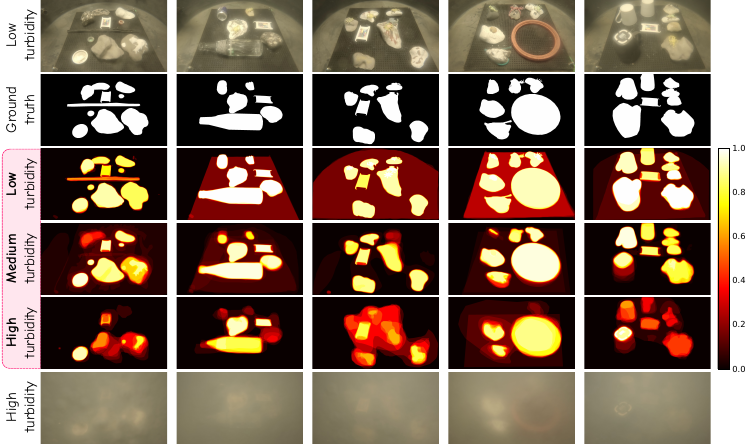}
    \caption{All five scenes with ground truth annotations and multi-annotator heatmaps (scale on the right). The low and high turbidity images are included for reference.}
    \label{fig:all_heatmaps}
\end{figure}

\cref{fig:threshold-conf-curves-instance} gives a complementary view, this time relating the instance-level IoU of individual annotations (\textit{GT}$\rightarrow$annotators) to their confidence, without penalizing missed objects. In low and medium turbidity, we observe a weak correlation between the two (Kendall's $\tau=0.26$, $p<10^{-20}$), suggesting that confidence can be a rough but noisy indicator of how well an annotated contour matches with the closest GT contour. In high turbidity, this relationship no longer holds at all (Kendall's $\tau=-0.01$, $p=0.08$). Note that as turbidity increases, setting a confidence threshold (e.g., at 0.5, corresponding to the vertical line in the plot) discards a growing proportion of reasonable annotations (IoU>0.75), echoing the drop in recall observed in~\cref{fig:threshold-conf-curves}.

\cref{fig:all_heatmaps} visualizes all the annotations collected in our experiments as multi-annotator heatmaps. Qualitatively, we find that multi-annotator ensembling helps smooth out discrepancies in task interpretation and annotation style; in clear images, the crowd recovers all the objects in all 5 scenes with precise contours. An impressive number of objects can be recovered in highly turbid scenes.
However, the ensemble is subject to group bias. When all or almost all annotators fail to identify an object, ensembling cannot compensate. Thus, ensembling brings diminishing returns under reduced visibility: turbidity adds systematic and consistent errors across individual annotators, especially False Negatives (missing objects) in the upper half of the image.

Zooming in on self-reported confidence, we observe widely different practices across annotators - some assign consistently high confidence to all clear-image objects (top right in ~\cref{fig:clear_annot_examples}), while others use the full range of confidences in a single image, despite drawing accurate contours (bottom left in~\cref{fig:clear_annot_examples}). Different types of objects are also associated with different average levels of confidence, with superimposed and/or non-standard items (e.g., nondescript beach trash) being among the least confidently annotated in clear images. Scene clutter also seems to play a role. For instance, the same color checkerboard appears in 4 scenes but is assigned 78.03\% confidence in scene Exp17\_Cam1 (least complex scene), 69.08\% in Exp18\_Cam3, 68.18\% in Exp30\_Cam1,  and 54.60\% in Exp6\_Cam1 (most complex scene). 
Clustering users by group or experience level reveals further diverging confidence patterns, which we explore below.

\begin{figure}[b!]
    \centering
    \includegraphics[width=0.47\linewidth]{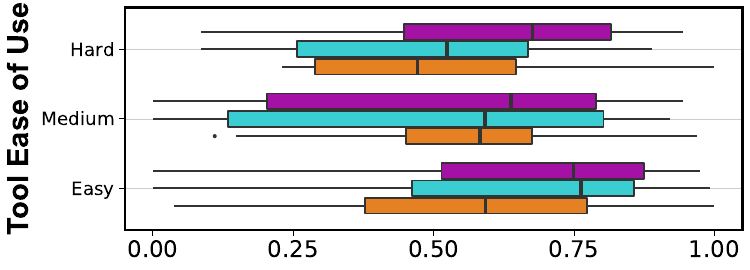}\hfill\includegraphics[width=0.47\linewidth]{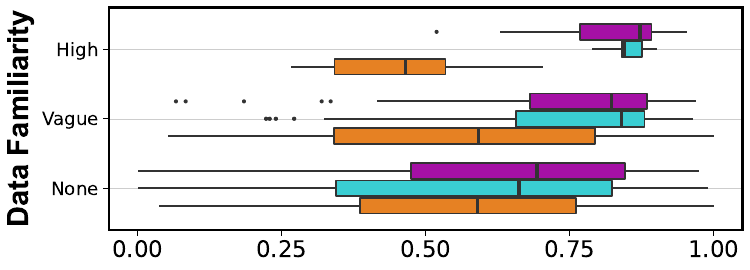}

    \includegraphics[width=0.47\linewidth]{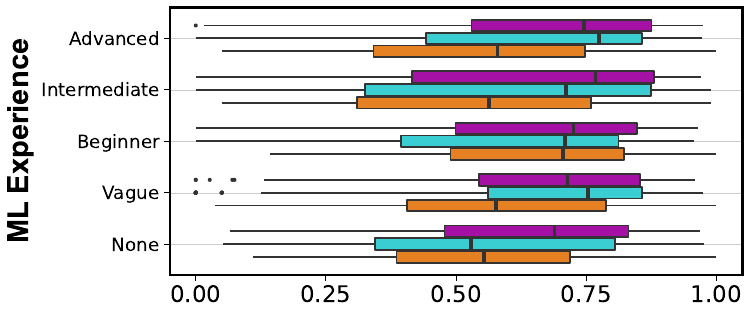}\hfill\includegraphics[width=0.47\linewidth]{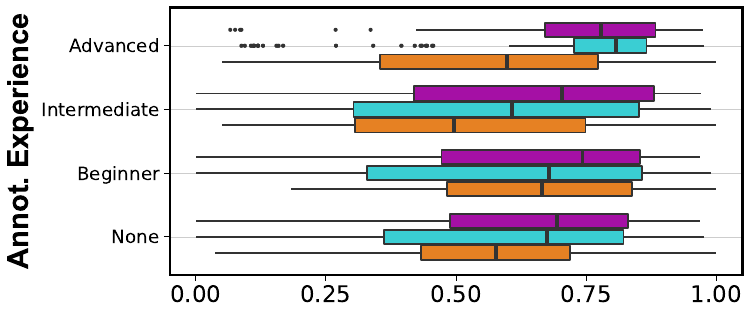}

    \includegraphics[width=\linewidth,trim={
    4cm 0 3cm 4cm}, clip]{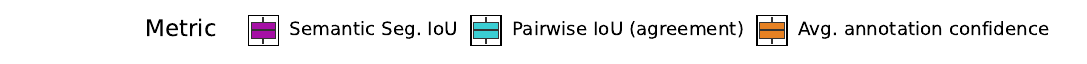}
    \caption{Annotation quality, inter-annotator agreement, and confidence of users with the same questionnaire answers. All metrics are computed at the image level.}
    \label{fig:iou_agreement_by_questionnaire_answers}
\end{figure}

\begin{figure}[t!]
\centering
\begin{tabular}{C{0.33\linewidth}C{0.33\linewidth}C{0.33\linewidth}}
\centering
{\comicneue \textbf{\scriptsize{\quad(A) Confidence}}} & {\comicneue \textbf{\quad\scriptsize{\quad(B) Agreement}}} & {\comicneue \textbf{\scriptsize{\quad\quad(C) Performance}}}\\
    \includegraphics[height=2.1cm]{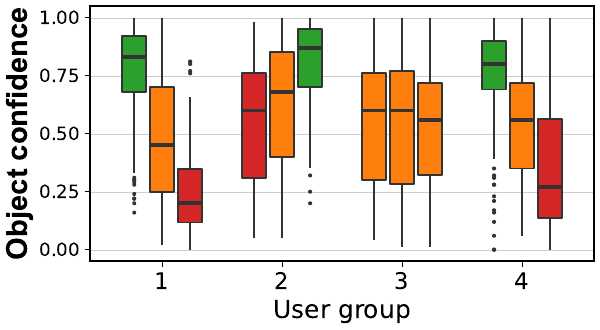} &
    \includegraphics[height=2.1cm]{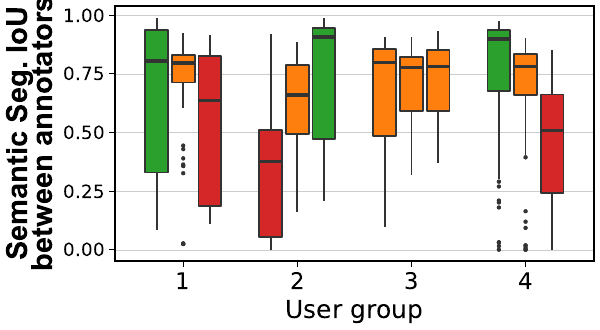} &
    \includegraphics[height=2.1cm]{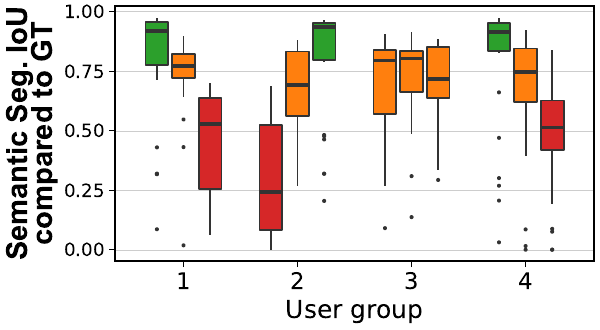}
\end{tabular}
    \caption{Interplay between annotations, user group, image order and turbidity. For each user group, the turbidity levels are shown in the order that they were shown in the study.}
    \label{fig:user_group_turbidity}

\end{figure}

\subsubsection{What role does background, proficiency or privileged information play?} Fig.~\ref{fig:iou_agreement_by_questionnaire_answers} looks at annotator performance, agreement and confidence based on questionnaire answers. We observe the highest level of agreement and fidelity between users who are very familiar with the dataset (top right of Fig.~\ref{fig:iou_agreement_by_questionnaire_answers}) and users who have annotated segmentation datasets before ("advanced" annotation experience). Users who reportedly struggled using the tool are the least likely to align in their annotations. Machine learning experience is the least informative indicator across all 3 metrics, with no significant differences between subgroups. Interestingly, out of the 15 top-annotators (14 unique people) identified in~\cref{fig:threshold-conf-curves}, only 1 was highly familiar with the TUB dataset; the majority had never seen TUB or related images before. Only 4 had experience annotating segmentation datasets, and 3 were not even familiar with the concept of annotation or ML in general. This suggests that lack of prior knowledge or experience is not necessarily a limiting factor for annotation.

Compared to general background knowledge, the order in which annotators are presented each image during the task seems to play a bigger role: the top-annotator of every high turbidity image in our study was part of \groupfour\ or \groupone. Both groups are shown increasingly turbid images and thus can first form an idea of what kind of objects to expect before losing visibility. We investigate how this affects annotation confidence, inter-annotator agreement and semantic segmentation performance in Fig~\ref{fig:user_group_turbidity}. We find that for turbid scenes, \groupfour\ and \groupone's prior information substantially raises their semantic segmentation performance and their agreement compared to \grouptwo, but also decreases their average confidence. When looking at medium turbidity annotations (image 2 for all groups), we do not see any notable differences between \groupthree\ and \groupfour. Interestingly, despite having the least amount of privileged information when annotating low visibility scenes, \grouptwo\ reports a higher confidence on average for every turbidity level, especially for high turbidity images (around 0.55 vs. around 0.3 for the other groups). This shows that the confidence assigned to manual annotations depends on many (sometimes counter-intuitive) factors which extend well beyond the content of the image itself.

\subsubsection{Do effort and time pay off?} 

\begin{figure}[htb]
    \centering
    \begin{subfigure}{.49\textwidth}
        \centering
        \includegraphics[height=3.3cm]{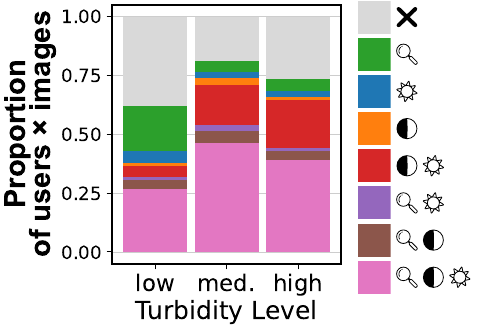}
        \caption{Types \& frequency of image adjustments.}
    \label{fig:boxplot_image_updates_by_turbidity_level_and_event}
    \end{subfigure}
    \hfill
    \begin{subfigure}{.49\textwidth}
    \centering
        \includegraphics[height=3.3cm]{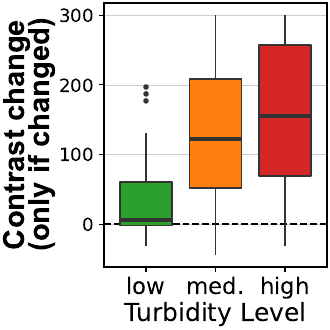}
        \caption{How much is contrast adjusted?}
    \label{fig:boxplot_image_settings_by_turbidity_level_and_event}
    \end{subfigure}
\par\bigskip 
    \begin{subfigure}{.49\textwidth}
    \centering
        \includegraphics[height=3.3cm]{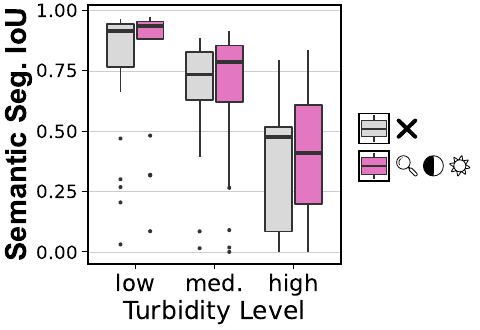}
        \caption{Do image adjustments help?}
    \label{fig:boxplot_image_settings_effort_by_turbidity_level}
    \end{subfigure}
    \hfill
    \begin{subfigure}{.49\textwidth}
        \centering
        \includegraphics[height=3.3cm]{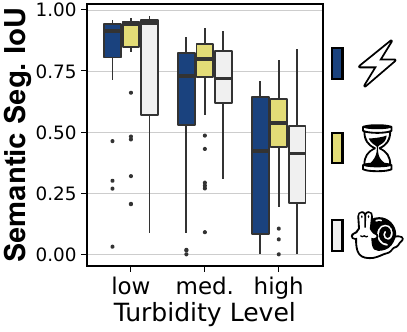}
        \caption{Does time per image help?}
    \label{fig:boxplot_time_iou_by_turbidity_level}
    \end{subfigure}
    \caption{Different facets of annotator behavior and effort under different turbidity levels.}
\end{figure}

 Looking at~\cref{fig:boxplot_image_updates_by_turbidity_level_and_event}, annotators frequently used one or more of the image adjustment sliders, especially when presented a medium turbidity scene. They were most likely to zoom in (\twemoji{mag}) when the image is clear. With increasing turbidity, they were more likely to adjust the brightness (\twemoji{sun}) and/or contrast (\twemoji{first_quarter_moon}), and \textit{when they did}, they opted for significantly higher contrast values (Fig.~\ref{fig:boxplot_image_settings_by_turbidity_level_and_event}), but similar brightness values (darker than default). At the same time, around 25\%  of submitted image annotations in our study were performed without using the sliders at all (\twemoji{x}). We compare the annotation quality of the lowest-effort \twemoji{x} group vs. the highest-effort \twemoji{mag}\twemoji{first_quarter_moon}\twemoji{sun} group in~\cref{{fig:boxplot_image_settings_effort_by_turbidity_level}}, and find that the effectiveness of image adjustments becomes less pronounced overall as visibility decreases, although it does slightly raise the best-case performance. When it comes to general user effort, we observe a similar pattern, where meticulous annotations are not necessarily better. We find that performing more frequent edits per contour can make them more precise (higher median \textit{GT}$\rightarrow$\textit{annotators} IoU), but does not make a difference in high turbidity. Furthermore, \cref{fig:boxplot_time_iou_by_turbidity_level} shows that there is a non-linear relation between time spent per image and segmentation performance, with a sweet spot achieved by medium-speed annotation (\twemoji{23f3}). Details about these experiments can be found in Appendix~\ref{app:results-plots}. Given the potential of ensembling, it would be interesting to investigate the trade-off between individual effort and number of annotators per image.

\section{Discussion}

\subsubsection{Limitations}
This work constitutes a preliminary study based on a relatively small dataset and a specific participant sample. As such, it remains to be investigated how well the findings generalize to other underwater or low-visibility settings. Some of the observed disagreement is likely influenced by the annotation instructions: we intentionally omitted descriptions of specific scenes or objects to avoid biasing annotators, but this left room for interpretation. Additionally, our participant pool consisted of volunteers and may not reflect the behavior of paid crowd workers or domain experts. A further limitation is the use of clear-reference manual annotations as ground truth. While these provide a useful upper bound for annotation quality under reduced visibility, they are themselves imperfect, are rarely available in practice, and may set an unrealistically high bar when image quality is highly degraded. More generally, without annotator explanations, it is difficult to disentangle the sources of annotation error and uncertainty.

\subsubsection{Future work}

While we have limited our analysis to the final annotations and confidence submitted by each user, future work could analyse the full timestamped annotation behaviour and its relation to annotation uncertainty/reliability. The link between object\allowbreak characteristics, depth, turbidity measurements, structural degradation, and their effect on annotation, should also be further investigated.
Furthermore, given that ensembles help mitigate some sources of disagreement, another promising direction is to explore effective ways of aggregating annotations at the instance level. 
More broadly, our findings raise questions about how vision models should be evaluated: if even the most dedicated annotators do not fully agree on the ground truth, evaluating models against a single annotation may provide a misleading and somewhat arbitrary estimate of performance.

\section{Conclusion}

In this work, we present the first multi-annotator study of image segmentation in turbid underwater images with more than 100 participants. We observe that annotators often approach and interpret the task differently, leading to substantial variation in annotation quality, object selection, and self-reported confidence, even in clear images. We find that image order matters: privileged information from clear scenes (or lack thereof) can have a significant effect.  As visibility decreases, performance drops systematically with annotators missing objects entirely, splitting objects into separate instances, and grouping multiple objects together. It also affects participants' ability to assess their own annotations: the calibration of self-reported confidence degrades alongside annotation quality. 

We identify annotator ensembles (the "wisdom of the crowd") as a promising avenue for mitigating the inter-annotator variability arising from different styles, task interpretations and levels of effort, as well as for recovering many barely-discernible objects in low visibility scenes. At the same time, ensembling reveals the systematic bias and errors arising under increasing turbidity, and the limits of manual annotation in the underwater domain. These issues warrant further interest in the computer vision community, as they inevitably propagate throughout ML training and evaluation pipelines.

\section*{Acknowledgements}
This work is supported by the Pioneer Centre for AI, DNRF grant number P1. This work is also supported by the Grundfos Foundation (“REPAI”, grant no. 83648813). We would also like to thank all the participants in the study, whose annotations enabled this work, and Roman Jurowetzki for valuable practical input on webapp deployment.

%
%


\bibliographystyle{splncs04}
\bibliography{main}

@String(CVPR= {Proceedings of the Computer Vision and Pattern Recognition Conference (CVPR)})

@String(ICCV= {Proceedings of the IEEE/CVF International Conference on Computer Vision (ICCV)})

@String(ECCV= {Computer Vision -- ECCV})

@String(NIPS= {Advances in Neural Information Processing Systems})

@String(ICML = {International Conference on Machine Learning })

@String(AAAI = {Proceedings of the AAAI Conference on Artificial Intelligence})

@String(WACV = {Proceedings of the IEEE/CVF Winter Conference on Applications of Computer Vision (WACV)})

@String(MICCAI = {Med. Image. Comput. Comput. Assist. Interv. (MICCAI)})

@String(IROS = {IEEE/RSJ International Conference on Intelligent Robots and Systems (IROS)})

@inproceedings{
labels-got-style_2023,
title={That Label's got Style: Handling Label Style Bias for Uncertain Image Segmentation},
author={Kilian Zepf and Eike Petersen and Jes Frellsen and Aasa Feragen},
booktitle={The Eleventh International Conference on Learning Representations },
year={2023},
url={https://openreview.net/forum?id=wZ2SVhOTzBX}
}

@InProceedings{jambo_2024,
author="G. Humblot-Renaux
and Johansen, Anders Skaarup
and Schmidt, Jonathan Eichild
and Irlind, Amanda Frederikke
and Madsen, Niels
and Moeslund, Thomas B.
and Pedersen, Malte",

title="Underwater Uncertainty: A Multi-annotator Image Dataset for Benthic Habitat Classification",
booktitle="Computer Vision -- ECCV 2024 Workshops",
year="2025",
publisher="Springer Nature Switzerland",
address="Cham",
pages="87--104",
isbn="978-3-031-92387-6",
doi = "10.1007/978-3-031-92387-6_6",
url = "https://link.springer.com/chapter/10.1007/978-3-031-92387-6_6"
}

@article{multi-annotator-benthic_2015,
    doi = {10.1371/journal.pone.0130312},
    author = {Beijbom, Oscar AND Edmunds, Peter J. AND Roelfsema, Chris AND Smith, Jennifer AND Kline, David I. AND Neal, Benjamin P. AND Dunlap, Matthew J. AND Moriarty, Vincent AND Fan, Tung-Yung AND Tan, Chih-Jui AND Chan, Stephen AND Treibitz, Tali AND Gamst, Anthony AND Mitchell, B. Greg AND Kriegman, David},
    journal = {PLOS ONE},
    publisher = {Public Library of Science},
    title = {Towards Automated Annotation of Benthic Survey Images: Variability of Human Experts and Operational Modes of Automation},
    year = {2015},
        volume = {10},
    url = {https://doi.org/10.1371/journal.pone.0130312},
    pages = {1-22},
    number = {7},

}

@inproceedings{arctique-label-noise_2024,
 author = {Franzen, Jannik and Winklmayr, Claudia and Guarino, Vanessa E. and Karg, Christoph and Yu, Xiaoyan and Koreuber, Nora and Albrecht, Jan P. and Bischoff, Philip and Kainmueller, Dagmar},
 booktitle = {Advances in Neural Information Processing Systems},
 doi = {10.52202/079017-2295},
 editor = {A. Globerson and L. Mackey and D. Belgrave and A. Fan and U. Paquet and J. Tomczak and C. Zhang},
 pages = {71855--71867},
 publisher = {Curran Associates, Inc.},
 title = {Arctique: An artificial histopathological dataset unifying realism and controllability for uncertainty quantification},
 url = {https://proceedings.neurips.cc/paper_files/paper/2024/file/840ba425392fbfede5cf50c755c608c6-Paper-Datasets_and_Benchmarks_Track.pdf},
 volume = {37},
 year = {2024}
}

@article{labeling-instructions-matter-biomed_2023,
  title={Labelling instructions matter in biomedical image analysis},
  author={R{\"a}dsch, Tim and Reinke, Annika and Weru, Vivienn and Tizabi, Minu D and Schreck, Nicholas and Kavur, A Emre and Pekdemir, B{\"u}nyamin and Ro{\ss}, Tobias and Kopp-Schneider, Annette and Maier-Hein, Lena},
  journal={Nature Machine Intelligence},
  volume={5},
  number={3},
  pages={273--283},
  year={2023},
  publisher={Nature Publishing Group UK London}
}

@article{study-annotator-agreement-eval_2016,
   title={An Empirical Study Into Annotator Agreement, Ground Truth Estimation, and Algorithm Evaluation},
   volume={25},
   ISSN={1941-0042},
   url={http://dx.doi.org/10.1109/TIP.2016.2544703},
   DOI={10.1109/tip.2016.2544703},
   number={6},
   journal={IEEE Transactions on Image Processing},
   publisher={Institute of Electrical and Electronics Engineers (IEEE)},
   author={Lampert, Thomas A. and Stumpf, Andre and Gancarski, Pierre},
   year={2016},
   month={June}, pages={2557–2572} }

@inproceedings{joly_lifeclef_2026,
    address = {Cham},
    title = {{LifeCLEF} 2026 {Teaser}: {AI} {Challenges} for {Biodiversity} {Understanding} and {Ecosystem} {Management}},
    isbn = {978-3-032-21321-1},
    shorttitle = {{LifeCLEF} 2026 {Teaser}},
    doi = {10.1007/978-3-032-21321-1_39},
    language = {en},
    booktitle = {Advances in {Information} {Retrieval}},
    publisher = {Springer Nature Switzerland},
    author = {Joly, Alexis and Picek, Lukáš and Kahl, Stefan and Goëau, Hervé and Adam, Lukáš and Bossy, Robert and Papafitsoros, Kostas and Čermák, Vojtěch and Klinck, Holger and Vellinga, Willem-Pier and Planqué, Robert and Denton, Tom and Chrobak, Laura and Barnard, Kevin and Nédellec, Claire and Deléger, Louise and Courtin, Marine and Martellucci, Giulio and Vinatier, Fabrice and Bonnet, Pierre},
    editor = {Campos, Ricardo and Jatowt, Adam and Lan, Yanyan and Aliannejadi, Mohammad and Bauer, Christine and MacAvaney, Sean and Anand, Avishek and Ren, Zhaochun and Verberne, Suzan and Bai, Nan and Mansoury, Masoud},
    year = {2026},
    pages = {287--296},
}

@article{blushtein-livnon_performance_2025,
    title = {Performance of {Human} {Annotators} in {Object} {Detection} and {Segmentation} of {Remotely} {Sensed} {Data}},
    volume = {63},
    issn = {1558-0644},
    url = {https://ieeexplore.ieee.org/document/10943264/},
    doi = {10.1109/TGRS.2025.3555235},
    urldate = {2026-06-16},
    journal = {IEEE Transactions on Geoscience and Remote Sensing},
    author = {Blushtein-Livnon, Roni and Svoray, Tal and Dorman, Michael},
    year = {2025},
    pages = {1--16},
}

@article{joshi_vision_2025,
    title = {Vision {Transformer}-{Based} {Unhealthy} {Tree} {Crown} {Detection} in {Mixed} {Northeastern} {US} {Forests} and {Evaluation} of {Annotation} {Uncertainty}},
    volume = {17},
    copyright = {http://creativecommons.org/licenses/by/3.0/},
    issn = {2072-4292},
    url = {https://www.mdpi.com/2072-4292/17/6/1066},
    doi = {10.3390/rs17061066},
    language = {en},
    number = {6},
    urldate = {2026-06-16},
    journal = {Remote Sensing},
    publisher = {Multidisciplinary Digital Publishing Institute},
    author = {Joshi, Durga and Witharana, Chandi},
    month = jan,
    year = {2025},
    pages = {1066},
}

@article{awad_ruod-r_2026,
    title = {{RUOD}-{R}: {A} {High}-{Fidelity} {Re}-{Annotated} {Benchmark} for {Underwater} {Object} {Detection}},
    volume = {14},
    issn = {2169-3536},
    shorttitle = {{RUOD}-{R}},
    url = {https://ieeexplore.ieee.org/document/11483160/},
    doi = {10.1109/ACCESS.2026.3685121},
    urldate = {2026-06-24},
    journal = {IEEE Access},
    author = {Awad, Ali and Saleem, Ashraf and Aljnadi, Yaman and Lucas, Evan and Paheding, Sidike and Havens, Timothy C.},
    year = {2026},
    pages = {60030--60046},
}

@article{wyatt_signal_2025,
    title = {Signal or noise? {Minimising} errors in image-based {AI} for marine ecosystem monitoring},
    volume = {41},
    issn = {1572-9761},
    shorttitle = {Signal or noise?},
    url = {https://doi.org/10.1007/s10980-025-02271-1},
    doi = {10.1007/s10980-025-02271-1},
    language = {en},
    number = {1},
    urldate = {2026-06-24},
    journal = {Landscape Ecology},
    author = {Wyatt, Mathew and Vercelloni, Julie and Faubel, Cal and Colquhuon, Jamie and Wakeford, Mary and Wilson, Shaun and Fulton, Christopher J.},
    month = dec,
    year = {2025},
    pages = {16},
}

@article{durden_comparison_2016,
    title = {Comparison of image annotation data generated by multiple investigators for benthic ecology},
    volume = {552},
    issn = {1616-1599, 0171-8630},
    url = {https://www.int-res.com/journals/meps/articles/meps11775},
    doi = {10.3354/meps11775},
    language = {en-US},
    urldate = {2026-06-24},
    journal = {Marine Ecology Progress Series},
    author = {Durden, Jennifer M. and Bett, B. J. and Schoening, Timm and Morris, Kirsty J. and Nattkemper, Tim W. and Ruhl, Henry A.},
    month = jun,
    year = {2016},
    pages = {61--70},
}

@article{schoening_recomiarecommendations_2016,
    title = {{RecoMIA}—{Recommendations} for {Marine} {Image} {Annotation}: {Lessons} {Learned} and {Future} {Directions}},
    volume = {3},
    issn = {2296-7745},
    shorttitle = {{RecoMIA}—{Recommendations} for {Marine} {Image} {Annotation}},
    url = {https://www.frontiersin.org/journals/marine-science/articles/10.3389/fmars.2016.00059/full},
    doi = {10.3389/fmars.2016.00059},
    language = {English},
    urldate = {2026-06-24},
    journal = {Frontiers in Marine Science},
    publisher = {Frontiers},
    author = {Schoening, Timm and Osterloff, Jonas and Nattkemper, Tim W.},
    month = apr,
    year = {2016},
}

@article{elsaser_seagrassfinder_2025,
    title = {{SeagrassFinder}: {Deep} learning for eelgrass detection and coverage estimation in the wild},
    volume = {90},
    issn = {1574-9541},
    shorttitle = {{SeagrassFinder}},
    url = {https://www.sciencedirect.com/science/article/pii/S1574954125002092},
    doi = {10.1016/j.ecoinf.2025.103200},
    urldate = {2026-06-24},
    journal = {Ecological Informatics},
    author = {Elsäßer, Jannik and Weihl, Laura and Cheplygina, Veronika and Nielsen, Lisbeth Tangaa},
    month = dec,
    year = {2025},
    pages = {103200},
}

@article{sakaridis_map-guided_2022,
    title = {Map-{Guided} {Curriculum} {Domain} {Adaptation} and {Uncertainty}-{Aware} {Evaluation} for {Semantic} {Nighttime} {Image} {Segmentation}},
    volume = {44},
    issn = {1939-3539},
    url = {https://ieeexplore.ieee.org/document/9298962/},
    doi = {10.1109/TPAMI.2020.3045882},
    number = {6},
    urldate = {2026-06-25},
    journal = {IEEE Transactions on Pattern Analysis and Machine Intelligence},
    author = {Sakaridis, Christos and Dai, Dengxin and Van Gool, Luc},
    month = jun,
    year = {2022},
    pages = {3139--3153},
}

@article{gurari_investigating_2016,
    title = {Investigating the {Influence} of {Data} {Familiarity} to {Improve} the {Design} of a {Crowdsourcing} {Image} {Annotation} {System}},
    volume = {4},
    copyright = {Copyright (c) 2016 Proceedings of the AAAI Conference on Human Computation and Crowdsourcing},
    issn = {2769-1349},
    url = {https://ojs.aaai.org/index.php/HCOMP/article/view/13294},
    doi = {10.1609/hcomp.v4i1.13294},
    language = {en},
    urldate = {2026-06-25},
    journal = {Proceedings of the AAAI Conference on Human Computation and Crowdsourcing},
    author = {Gurari, Danna and Sameki, Mehrnoosh and Betke, Margrit},
    month = sep,
    year = {2016},
    pages = {59--68},
}

@article{hernandez-giron_low_2015,
    title = {Low contrast detectability performance of model observers based on {CT} phantom images: {kVp} influence},
    volume = {31},
    issn = {1120-1797},
    shorttitle = {Low contrast detectability performance of model observers based on {CT} phantom images},
    url = {https://www.sciencedirect.com/science/article/pii/S1120179715001039},
    doi = {10.1016/j.ejmp.2015.04.012},
    number = {7},
    urldate = {2026-06-25},
    journal = {Physica Medica},
    author = {Hernandez-Giron, I. and Calzado, A. and Geleijns, J. and Joemai, R. M. S. and Veldkamp, W. J. H.},
    month = nov,
    year = {2015},
    pages = {798--807},
}

@article{kraff_uncertainties_2020,
    title = {Uncertainties of {Human} {Perception} in {Visual} {Image} {Interpretation} in {Complex} {Urban} {Environments}},
    volume = {13},
    issn = {2151-1535},
    url = {https://ieeexplore.ieee.org/document/9146616/},
    doi = {10.1109/JSTARS.2020.3011543},
    urldate = {2026-06-25},
    journal = {IEEE Journal of Selected Topics in Applied Earth Observations and Remote Sensing},
    author = {Kraff, Nicolas Johannes and Wurm, Michael and Taubenböck, Hannes},
    year = {2020},
    pages = {4229--4241},
}

@InProceedings{what-can-we-learn-annotator-var-skin-seg_2026,
author="Abhishek, Kumar
and Kawahara, Jeremy
and Hamarneh, Ghassan",
editor="Celebi, M. Emre
and M{\"u}ller, Johanna Paula
and Barata, Catarina
and Halpern, Allan
and Tschandl, Philipp
and Combalia, Marc
and Liu, Yuan
and Abhishek, Kumar
and Jaworek-Korjakowska, Joanna
and Yap, Moi Hoon
and Breininger, Katharina
and Lindholz, Maximilian
and Hutter, Jana
and Ruppel, Richard
and Tripathy, Smiti
and Mathis-Ullrich, Franziska
and Burghaus, Stefanie
and May, Matthias",
title="What Can We Learn from Inter-Annotator Variability in Skin Lesion Segmentation?",
booktitle="Skin Image Analysis, and Computer-Aided Pelvic Imaging for Female Health",
year="2026",
publisher="Springer Nature Switzerland",
address="Cham",
pages="23--33",
isbn="978-3-032-05825-6",
doi={10.1007/978-3-032-05825-6_3}
}

@ARTICLE{agreement-heatmaps-medseg_2023,
  author={Yang, Feng and Zamzmi, Ghada and Angara, Sandeep and Rajaraman, Sivaramakrishnan and Aquilina, André and Xue, Zhiyun and Jaeger, Stefan and Papagiannakis, Emmanouil and Antani, Sameer K.},
  journal={IEEE Access}, 
  title={Assessing Inter-Annotator Agreement for Medical Image Segmentation}, 
  year={2023},
  volume={11},
  number={},
  pages={21300-21312},
  doi={10.1109/ACCESS.2023.3249759}}

@inproceedings{weber-genzel_varierr_2024,
    address = {Bangkok, Thailand},
    title = {{VariErr} {NLI}: {Separating} {Annotation} {Error} from {Human} {Label} {Variation}},
    shorttitle = {{VariErr} {NLI}},
    url = {https://aclanthology.org/2024.acl-long.123/},
    doi = {10.18653/v1/2024.acl-long.123},
    urldate = {2026-06-30},
    booktitle = {Proceedings of the 62nd {Annual} {Meeting} of the {Association} for {Computational} {Linguistics} ({Volume} 1: {Long} {Papers})},
    publisher = {Association for Computational Linguistics},
    author = {Weber-Genzel, Leon and Peng, Siyao and De Marneffe, Marie-Catherine and Plank, Barbara},
    editor = {Ku, Lun-Wei and Martins, Andre and Srikumar, Vivek},
    month = aug,
    year = {2024},
    pages = {2256--2269},
}

@article{sylolypavan_impact_2023,
    title = {The impact of inconsistent human annotations on {AI} driven clinical decision making},
    volume = {6},
    copyright = {2023 The Author(s)},
    issn = {2398-6352},
    url = {https://www.nature.com/articles/s41746-023-00773-3},
    doi = {10.1038/s41746-023-00773-3},
    language = {en},
    number = {1},
    urldate = {2026-06-30},
    journal = {npj Digital Medicine},
    publisher = {Nature Publishing Group},
    author = {Sylolypavan, Aneeta and Sleeman, Derek and Wu, Honghan and Sim, Malcolm},
    month = feb,
    year = {2023},
    pages = {26},
}

@inproceedings{tschirschwitz_kalos_2026,
    title = {{KaLOS} finds {Consensus}: {A} {Meta}-{Algorithm} for {Evaluating} {Inter}-{Annotator} {Agreement} in {Complex} {Vision} {Tasks}},
    shorttitle = {{KaLOS} finds {Consensus}},
    url = {https://openaccess.thecvf.com/content/CVPR2026/html/Tschirschwitz_KaLOS_finds_Consensus_A_Meta-Algorithm_for_Evaluating_Inter-Annotator_Agreement_in_CVPR_2026_paper.html},
    booktitle = {Proceedings of the IEEE/CVF Conference on Computer Vision and Pattern Recognition (CVPR)},
    language = {en},
    urldate = {2026-06-30},
    author = {Tschirschwitz, David and Rodehorst, Volker},
    year = {2026},
    pages = {38554--38563},
}

@misc{ribeiro_handling_2019,
    title = {Handling {Inter}-{Annotator} {Agreement} for {Automated} {Skin} {Lesion} {Segmentation}},
    url = {http://arxiv.org/abs/1906.02415},
    doi = {10.48550/arXiv.1906.02415},
    urldate = {2026-06-30},
    publisher = {arXiv},
    author = {Ribeiro, Vinicius and Avila, Sandra and Valle, Eduardo},
    month = jun,
    year = {2019},
    note = {arXiv:1906.02415 [cs.CV]},
}

@article{uma_learning_2021,
    title = {Learning from {Disagreement}: {A} {Survey}},
    volume = {72},
    copyright = {Copyright (c)},
    issn = {1076-9757},
    shorttitle = {Learning from {Disagreement}},
    url = {https://www.jair.org/index.php/jair/article/view/12752},
    doi = {10.1613/jair.1.12752},
    language = {en},
    urldate = {2026-06-30},
    journal = {Journal of Artificial Intelligence Research},
    author = {Uma, Alexandra N. and Fornaciari, Tommaso and Hovy, Dirk and Paun, Silviu and Plank, Barbara and Poesio, Massimo},
    month = dec,
    year = {2021},
    pages = {1385--1470},
}

@article{Raine2026,
  title = {AI-Driven Marine Robotics: Emerging Trends in Underwater Perception and Ecosystem Monitoring},
  volume = {40},
  ISSN = {2159-5399},
  url = {http://dx.doi.org/10.1609/aaai.v40i48.42133},
  DOI = {10.1609/aaai.v40i48.42133},
  number = {48},
  journal = {Proceedings of the AAAI Conference on Artificial Intelligence},
  publisher = {Association for the Advancement of Artificial Intelligence (AAAI)},
  author = {Raine,  Scarlett and Fischer,  Tobias},
  year = {2026},
  month = Mar,
  pages = {40981–40989}
}

@inproceedings{test_set_errors-2021,
 author = {Northcutt, Curtis and Athalye, Anish and Mueller, Jonas},
 booktitle = {Proceedings of the Neural Information Processing Systems Track on Datasets and Benchmarks},
  pages = {},
 title = {Pervasive Label Errors in Test Sets Destabilize Machine Learning Benchmarks},
 url = {https://datasets-benchmarks-proceedings.neurips.cc/paper_files/paper/2021/file/f2217062e9a397a1dca429e7d70bc6ca-Paper-round1.pdf},
 volume = {1},
 year = {2021}
}

@article{yan_learning_2014,
    title = {Learning from multiple annotators with varying expertise},
    volume = {95},
    issn = {1573-0565},
    url = {https://doi.org/10.1007/s10994-013-5412-1},
    doi = {10.1007/s10994-013-5412-1},
    language = {en},
    number = {3},
    urldate = {2026-04-28},
    journal = {Machine Learning},
    author = {Yan, Yan and Rosales, Rómer and Fung, Glenn and Subramanian, Ramanathan and Dy, Jennifer},
    month = jun,
    year = {2014},
    pages = {291--327},
}

@inproceedings{imagenet-labelling-issues-2020,
author = {Tsipras, Dimitris and Santurkar, Shibani and Engstrom, Logan and Ilyas, Andrew and M\k{a}dry, Aleksander},
title = {From ImageNet to image classification: contextualizing progress on benchmarks},
year = {2020},
publisher = {JMLR.org},
booktitle = ICML,
articleno = {892},
numpages = {11},
series = {ICML'20},
url={https://proceedings.mlr.press/v119/tsipras20a.html}
}

@article{commowick_multiple_2021,
    title = {Multiple sclerosis lesions segmentation from multiple experts: {The} {MICCAI} 2016 challenge dataset},
    volume = {244},
    issn = {1053-8119},
    shorttitle = {Multiple sclerosis lesions segmentation from multiple experts},
    url = {https://www.sciencedirect.com/science/article/pii/S1053811921008624},
    doi = {10.1016/j.neuroimage.2021.118589},
    urldate = {2026-07-01},
    journal = {NeuroImage},
    author = {Commowick, Olivier and Kain, Michaël and Casey, Romain and Ameli, Roxana and Ferré, Jean-Christophe and Kerbrat, Anne and Tourdias, Thomas and Cervenansky, Frédéric and Camarasu-Pop, Sorina and Glatard, Tristan and Vukusic, Sandra and Edan, Gilles and Barillot, Christian and Dojat, Michel and Cotton, Francois},
    month = dec,
    year = {2021},
    pages = {118589},
}

@InProceedings{impact-annotation-error-seg_2022,
  title = 	 {An Analysis of the Impact of Annotation Errors on the Accuracy of Deep Learning for Cell Segmentation},
  author =       {V\u{a}dineanu, \c{S}erban and Pelt, Dani{\"e}l Maria and Dzyubachyk, Oleh and Batenburg, Kees Joost},
  booktitle = 	 {Proceedings of The 5th International Conference on Medical Imaging with Deep Learning},
  pages = 	 {1251--1267},
  year = 	 {2022},
  editor = 	 {Konukoglu, Ender and Menze, Bjoern and Venkataraman, Archana and Baumgartner, Christian and Dou, Qi and Albarqouni, Shadi},
  volume = 	 {172},
  series = 	 {Proceedings of Machine Learning Research},
  month = 	 {06--08 Jul},
  publisher =    {PMLR},
  url = 	 {https://proceedings.mlr.press/v172/vadineanu22a.html}
}

@article{erf-dataset-rain_2025, title={ERF: A Benchmark Dataset for Robust Semantic Segmentation Under Extreme Rainfall Conditions}, volume={39}, url={https://ojs.aaai.org/index.php/AAAI/article/view/33007}, DOI={10.1609/aaai.v39i9.33007}, abstractNote={As climate change reshapes global weather patterns, the increasing frequency and intensity of extreme rainfall events have amplified the safety imperatives for autonomous driving systems. During such events, rainfall can escalate from heavy to violent, as defined by the World Meteorological Organization, severely impairing images with diverse and significant degradations. Many existing semantic segmentation models perform well under light to heavy rain, but there is a notable absence of datasets addressing violent rain conditions for these models to validate and learn from. In this paper, we introduce the Extreme RainFall (ERF) dataset for semantic segmentation in both image and video tasks under violent rain conditions. Our dataset comprises 14,757 unlabeled frames and 100 labeled frames, all captured during four different violent rainfall periods. We use our dataset to evaluate the robustness of various methods against violent rainfall, focusing on four approaches: 1) image-based foundation models, 2) image-based domain generalization methods, 3) image-based domain adaptation methods, and 4) video-based methods. The results reveal that none of the existing models tested is capable of withstanding the extreme challenges posed by violent rainfall conditions. By analyzing the results, we offer insights and suggestions for developing more robust models under extreme rainfall events.}, number={9}, journal={Proceedings of the AAAI Conference on Artificial Intelligence}, author={Yang, Xin and Zhang, Xin and Wang, Xinchao}, year={2025}, month={Apr.}, pages={9301-9309} }

@inproceedings{gurari_how_2015,
    title = {How to {Collect} {Segmentations} for {Biomedical} {Images}? {A} {Benchmark} {Evaluating} the {Performance} of {Experts}, {Crowdsourced} {Non}-experts, and {Algorithms}},
    issn = {1550-5790},
    shorttitle = {How to {Collect} {Segmentations} for {Biomedical} {Images}?},
    url = {https://ieeexplore.ieee.org/abstract/document/7046014},
    doi = {10.1109/WACV.2015.160},
    urldate = {2026-07-01},
    booktitle = {2015 {IEEE} {Winter} {Conference} on {Applications} of {Computer} {Vision}},
    author = {Gurari, Danna and Theriault, Diane and Sameki, Mehrnoosh and Isenberg, Brett and Pham, Tuan A. and Purwada, Alberto and Solski, Patricia and Walker, Matthew and Zhang, Chentian and Wong, Joyce Y. and Betke, Margrit},
    month = jan,
    year = {2015},
    note = {ISSN: 1550-5790},
    pages = {1169--1176},
}

@article{foggy-cityscapes_2018,
  author = {Sakaridis, Christos and Dai, Dengxin and Van Gool, Luc},
  title = {Semantic Foggy Scene Understanding with Synthetic Data},
  journal = {International Journal of Computer Vision},
  year = {2018},
  month = {Sep},
  volume = {126},
  number = {9},
  pages = {973--992},
  url = {https://doi.org/10.1007/s11263-018-1072-8}
}

@INPROCEEDINGS{acdc-dataset_2021,
  author={Sakaridis, Christos and Dai, Dengxin and Van Gool, Luc},
  booktitle={2021 IEEE/CVF International Conference on Computer Vision (ICCV)}, 
  title={ACDC: The Adverse Conditions Dataset with Correspondences for Semantic Driving Scene Understanding}, 
  year={2021},
  volume={},
  number={},
  pages={10745-10755},
  doi={10.1109/ICCV48922.2021.01059}}

@article{study-weak-noisy-annot_2025,
  title={How to efficiently annotate images for best-performing deep learning-based segmentation models: An empirical study with weak and noisy annotations and segment anything model},
  author={Zhang, Yixin and Zhao, Shen and Gu, Hanxue and Mazurowski, Maciej A},
  journal={Journal of Imaging Informatics in Medicine},
  volume={38},
  number={5},
  pages={3235--3247},
  year={2025},
  publisher={Springer},
  url = {https://link.springer.com/article/10.1007/s10278-025-01408-7},
  doi = {doi.org/10.1007/s10278-025-01408-7}
}

@Article{effect-label-noise-remoteseg_2022,
AUTHOR = {Maiti, A. and Oude Elberink, S. J. and Vosselman, G.},
TITLE = {EFFECT OF LABEL NOISE IN SEMANTIC SEGMENTATION OF HIGH RESOLUTION AERIAL IMAGES AND HEIGHT DATA},
JOURNAL = {ISPRS Annals of the Photogrammetry, Remote Sensing and Spatial Information Sciences},
VOLUME = {V-2-2022},
YEAR = {2022},
PAGES = {275--282},
URL = {https://isprs-annals.copernicus.org/articles/V-2-2022/275/2022/},
DOI = {10.5194/isprs-annals-V-2-2022-275-2022}
}

@INPROCEEDINGS{synthia-dataset_2016,
  author={Ros, German and Sellart, Laura and Materzynska, Joanna and Vazquez, David and Lopez, Antonio M.},
  booktitle={2016 IEEE Conference on Computer Vision and Pattern Recognition (CVPR)}, 
  title={The SYNTHIA Dataset: A Large Collection of Synthetic Images for Semantic Segmentation of Urban Scenes}, 
  year={2016},
  volume={},
  number={},
  pages={3234-3243},
  doi={10.1109/CVPR.2016.352}}

@ARTICLE{jaffe2015,
  author={Jaffe, Jules S.},
  journal={IEEE Journal of Oceanic Engineering}, 
  title={Underwater Optical Imaging: The Past, the Present, and the Prospects}, 
  year={2015},
  volume={40},
  number={3},
  pages={683-700},
  doi={10.1109/JOE.2014.2350751}}

@inbook{Islam2024,
  title = {Computer Vision Applications in Underwater Robotics and Oceanography},
  ISBN = {9781003328957},
  url = {http://dx.doi.org/10.1201/9781003328957-9},
  DOI = {10.1201/9781003328957-9},
  booktitle = {Computer Vision},
  publisher = {Chapman and Hall/CRC},
  author = {Islam,  Md Jahidul and Li,  Alberto Quattrini and Girdhar,  Yogesh A and Rekleitis,  Ioannis},
  year = {2024},
  month = May,
  pages = {173–204}
}

@inproceedings{lian_watermask_2023,
    address = {Paris, France},
    title = {{WaterMask}: {Instance} {Segmentation} for {Underwater} {Imagery}},
    copyright = {https://doi.org/10.15223/policy-029},
    isbn = {979-8-3503-0718-4},
    shorttitle = {{WaterMask}},
    url = {https://ieeexplore.ieee.org/document/10376692/},
    doi = {10.1109/ICCV51070.2023.00126},
    language = {en},
    urldate = {2026-07-03},
    booktitle = {2023 {IEEE}/{CVF} {International} {Conference} on {Computer} {Vision} ({ICCV})},
    publisher = {IEEE},
    author = {Lian, Shijie and Li, Hua and Cong, Runmin and Li, Suqi and Zhang, Wei and Kwong, Sam},
    month = oct,
    year = {2023},
    pages = {1305--1315},
}

@inproceedings{islam_semantic_2020,
    title = {Semantic {Segmentation} of {Underwater} {Imagery}: {Dataset} and {Benchmark}},
    issn = {2153-0866},
    shorttitle = {Semantic {Segmentation} of {Underwater} {Imagery}},
    url = {https://ieeexplore.ieee.org/abstract/document/9340821},
    doi = {10.1109/IROS45743.2020.9340821},
    urldate = {2026-07-03},
    booktitle = {2020 {IEEE}/{RSJ} {International} {Conference} on {Intelligent} {Robots} and {Systems} ({IROS})},
    author = {Islam, Md Jahidul and Edge, Chelsey and Xiao, Yuyang and Luo, Peigen and Mehtaz, Muntaqim and Morse, Christopher and Enan, Sadman Sakib and Sattar, Junaed},
    month = oct,
    year = {2020},
    note = {ISSN: 2153-0866},
    pages = {1769--1776},
}

@misc{ismiroglou_beyond_2026,
    title = {Beyond {Aesthetics}: {Quantifying} {Information} {Loss} in {Turbid} {Scenes}},
    shorttitle = {Beyond {Aesthetics}},
    url = {http://arxiv.org/abs/2606.26295},
    doi = {10.48550/arXiv.2606.26295},
    urldate = {2026-07-03},
    publisher = {arXiv},
    author = {Ismiroglou, Vasiliki and Bengtson, Stefan H. and Benos, Tasos and Moeslund, Thomas B. and Pedersen, Malte},
    month = jun,
    year = {2026},
    note = {arXiv:2606.26295 [cs.CV]},
}

@inproceedings{bauchwitz_task_2025,
    title = {Task {Configuration} {Impacts} {Annotation} {Quality} and {Model} {Training} {Performance} in {Crowdsourced} {Image} {Segmentation}},
    url = {https://openaccess.thecvf.com/content/WACV2025/html/Bauchwitz_Task_Configuration_Impacts_Annotation_Quality_and_Model_Training_Performance_in_WACV_2025_paper.html},
    language = {en},
    urldate = {2026-07-01},
    booktitle = {Proceedings of the {Winter} {Conference} on {Applications} of {Computer} {Vision} ({WACV})},
    author = {Bauchwitz, Benjamin R. and Cummings, Mary},
    year = {2025},
    pages = {6646--6656},
}

@inproceedings{sauder_coralscapes_2025,
    title = {The {Coralscapes} {Dataset}: {Semantic} {Scene} {Understanding} in {Coral} {Reefs}},
    shorttitle = {The {Coralscapes} {Dataset}},
    url = {https://openaccess.thecvf.com/content/ICCV2025W/CVAUI\%20\%26\%20AAMVEM/html/Sauder_The_Coralscapes_Dataset_Semantic_Scene_Understanding_in_Coral_Reefs_ICCVW_2025_paper.html},
    language = {en},
    urldate = {2026-07-03},
    booktitle = {Proceedings of the {IEEE}/{CVF} {International} {Conference} on {Computer} {Vision} ({ICCV}) {Workshops}},
    author = {Sauder, Jonathan and Domazetoski, Viktor and Banc-Prandi, Guilhem and Perna, Gabriela and Meibom, Anders and Tuia, Devis},
    year = {2025},
    pages = {2136--2143},
}

@article{schumann_consensus_2023,
    address = {Red Hook, NY, USA},
    series = {{NIPS} '23},
    title = {Consensus and {Subjectivity} of {Skin} {Tone} {Annotation} for {ML} {Fairness}},
    language = {en},
    journal = {Proceedings of the 37th International Conference on Neural Information Processing Systems},
    publisher = {Curran Associates Inc.},
    author = {Schumann, Candice and Olanubi, Gbolahan O and Wright, Auriel and Monk, Ellis and Heldreth, Courtney and Ricco, Susanna},
    year = {2023},
    pages = {30},
    url = {https://proceedings.neurips.cc/paper_files/paper/2023/hash/60d25b3210c92f5ba2002a8e1f1adf1c-Abstract-Datasets_and_Benchmarks.html}
}

@inproceedings{pedersen_detection_2019,
    title = {Detection of {Marine} {Animals} in a {New} {Underwater} {Dataset} with {Varying} {Visibility}},
    url = {https://openaccess.thecvf.com/content_CVPRW_2019/html/AAMVEM/Pedersen_Detection_of_Marine_Animals_in_a_New_Underwater_Dataset_with_CVPRW_2019_paper.html},
    urldate = {2026-07-03},
    booktitle = {Proceedings of the {IEEE}/{CVF} {Conference} on {Computer} {Vision} and {Pattern} {Recognition} ({CVPR}) {Workshops}},
    author = {Pedersen, Malte and Bruslund Haurum, Joakim and Gade, Rikke and Moeslund, Thomas B.},
    year = {2019},
    pages = {18--26},
}

@inproceedings{lucas_underwater_2025,
    title = {Underwater {Image} {Enhancement} and {Object} {Detection}: {Are} {Poor} {Object} {Detection} {Results} {On} {Enhanced} {Images} {Due} to {Missing} {Human} {Labels}?},
    shorttitle = {Underwater {Image} {Enhancement} and {Object} {Detection}},
    url = {https://openaccess.thecvf.com/content/WACV2025W/MaCVi/html/Lucas_Underwater_Image_Enhancement_and_Object_Detection_Are_Poor_Object_Detection_WACVW_2025_paper.html},
    language = {en},
    urldate = {2026-06-15},
    booktitle = {Proceedings of the {Winter} {Conference} on {Applications} of {Computer} {Vision} ({WACV}) {Workshops}},
    author = {Lucas, Evan and Awad, Ali and Geglio, Anthony and Moradi, Shadi and Saleem, Ashraf and Havens, Timothy and Galloway, Angus and Paheding, Sidike},
    year = {2025},
    pages = {1520--1525},
}

@article{schoening_semi-automated_2012,
    title = {Semi-{Automated} {Image} {Analysis} for the {Assessment} of {Megafaunal} {Densities} at the {Arctic} {Deep}-{Sea} {Observatory} {HAUSGARTEN}},
    volume = {7},
    url = {https://doi.org/10.1371/journal.pone.0038179},
    doi = {10.1371/journal.pone.0038179},
    number = {6},
    journal = {PLOS ONE},
    publisher = {Public Library of Science},
    author = {Schoening, Timm and Bergmann, Melanie and Ontrup, Jörg and Taylor, James and Dannheim, Jennifer and Gutt, Julian and Purser, Autun and Nattkemper, Tim W.},
    month = jun,
    year = {2012},
    pages = {1--14},
}

@inproceedings{li_exploring_2026,
    title = {Exploring the {Underwater} {World} {Segmentation} without {Extra} {Training}},
    url = {https://openaccess.thecvf.com/content/CVPR2026/html/Li_Exploring_the_Underwater_World_Segmentation_without_Extra_Training_CVPR_2026_paper.html},
    booktitle={Proceedings of the IEEE/CVF Conference on Computer Vision and Pattern Recognition (CVPR)},
    language = {en},
    urldate = {2026-07-03},
    author = {Li, Bingyu and Huo, Tao and Zhang, Da and Zhao, Zhiyuan and Gao, Junyu and Li, Xuelong},
    year = {2026},
    pages = {39879--39889},
}

@article{jahanbakht_semi-supervised_2023,
    title = {Semi-supervised and weakly-supervised deep neural networks and dataset for fish detection in turbid underwater videos},
    volume = {78},
    issn = {15749541},
    url = {https://linkinghub.elsevier.com/retrieve/pii/S1574954123003321},
    doi = {10.1016/j.ecoinf.2023.102303},
    language = {en},
    urldate = {2026-02-06},
    journal = {Ecological Informatics},
    author = {Jahanbakht, Mohammad and Rahimi Azghadi, Mostafa and Waltham, Nathan J.},
    month = dec,
    year = {2023},
    pages = {102303},
}

@article{jansi_rani_novel_2024,
    title = {A novel automated approach for fish biomass estimation in turbid environments through deep learning, object detection, and regression},
    volume = {81},
    issn = {1574-9541},
    url = {https://www.sciencedirect.com/science/article/pii/S157495412400205X},
    doi = {10.1016/j.ecoinf.2024.102663},
    urldate = {2026-07-03},
    journal = {Ecological Informatics},
    author = {Jansi Rani, S. V. and Ioannou, Iacovos and Swetha, R. and Dhivya Lakshmi, R. M. and Vassiliou, Vasos},
    month = jul,
    year = {2024},
    pages = {102663},
}

@inproceedings{zhou_treasure_2023,
    title = {The {Treasure} {Beneath} {Multiple} {Annotations}: {An} {Uncertainty}-{Aware} {Edge} {Detector}},
    shorttitle = {The {Treasure} {Beneath} {Multiple} {Annotations}},
    url = {https://openaccess.thecvf.com/content/CVPR2023/html/Zhou_The_Treasure_Beneath_Multiple_Annotations_An_Uncertainty-Aware_Edge_Detector_CVPR_2023_paper.html},
    booktitle={Proceedings of the IEEE/CVF Conference on Computer Vision and Pattern Recognition (CVPR)},
    language = {en},
    urldate = {2026-07-03},
    author = {Zhou, Caixia and Huang, Yaping and Pu, Mengyang and Guan, Qingji and Huang, Li and Ling, Haibin},
    year = {2023},
    pages = {15507--15517},
}

@article{schmarje_is_2022,
    title = {Is one annotation enough? - {A} data-centric image classification benchmark for noisy and ambiguous label estimation},
    volume = {35},
    shorttitle = {Is one annotation enough?},
    url = {https://proceedings.neurips.cc/paper_files/paper/2022/hash/d6c03035b8bc551f474f040fe8607cab-Abstract-Datasets_and_Benchmarks.html},
    language = {en},
    urldate = {2026-07-03},
    journal = {Advances in Neural Information Processing Systems},
    author = {Schmarje, Lars and Grossmann, Vasco and Zelenka, Claudius and Dippel, Sabine and Kiko, Rainer and Oszust, Mariusz and Pastell, Matti and Stracke, Jenny and Valros, Anna and Volkmann, Nina and Koch, Reinhard},
    month = dec,
    year = {2022},
    pages = {33215--33232},
}

@inproceedings{rottmann_automated_2023,
    title = {Automated {Detection} of {Label} {Errors} in {Semantic} {Segmentation} {Datasets} via {Deep} {Learning} and {Uncertainty} {Quantification}},
    url = {https://openaccess.thecvf.com/content/WACV2023/html/Rottmann_Automated_Detection_of_Label_Errors_in_Semantic_Segmentation_Datasets_via_WACV_2023_paper.html},
    booktitle={Proceedings of the IEEE/CVF Winter Conference on Applications of Computer Vision (WACV)},
    language = {en},
    urldate = {2026-07-03},
    author = {Rottmann, Matthias and Reese, Marco},
    year = {2023},
    pages = {3214--3223},
}

@article{
multi-annotator-dl-framework_2023,
title={Multi-annotator Deep Learning: A Probabilistic Framework for Classification},
author={Marek Herde and Denis Huseljic and Bernhard Sick},
journal={Transactions on Machine Learning Research},
issn={2835-8856},
year={2023},
url={https://openreview.net/forum?id=MgdoxzImlK},
note={}
}

@INPROCEEDINGS{suim-dataset_2020,
  author={Islam, Md Jahidul and Edge, Chelsey and Xiao, Yuyang and Luo, Peigen and Mehtaz, Muntaqim and Morse, Christopher and Enan, Sadman Sakib and Sattar, Junaed},
  booktitle=IROS, 
  title={Semantic Segmentation of Underwater Imagery: Dataset and Benchmark}, 
  year={2020},
  volume={},
  number={},
  pages={1769-1776},
  doi={10.1109/IROS45743.2020.9340821}}

@misc{imperfect-thesis,
title = "Imperfect labels and imperfect classifiers",
author = "Galadrielle Humblot-Renaux",
year = "2025",
doi = "10.54337/aau812590654",
language = "English",
series = "Ph.d.-serien for Det Tekniske Fakultet for IT og Design, Aalborg Universitet",
publisher = "Aalborg University Open Publishing",
}

@article{review-cv-benthos_2025,
title = {Surveying the deep: A review of computer vision in the benthos},
journal = {Ecological Informatics},
volume = {86},
pages = {102989},
year = {2025},
issn = {1574-9541},
doi = {https://doi.org/10.1016/j.ecoinf.2024.102989},
url = {https://www.sciencedirect.com/science/article/pii/S1574954124005314},
author = {Cameron Trotter and Huw J. Griffiths and Rowan J. Whittle}
}

@article{survey-underwater-cv_2023,
author = {Gonz\'{a}lez-Sabbagh, Salma P. and Robles-Kelly, Antonio},
title = {A Survey on Underwater Computer Vision},
year = {2023},
issue_date = {December 2023},
publisher = {Association for Computing Machinery},
address = {New York, NY, USA},
volume = {55},
number = {13s},
issn = {0360-0300},
url = {https://doi.org/10.1145/3578516},
doi = {10.1145/3578516},
journal = {ACM Comput. Surv.},
month = jul,
articleno = {268},
numpages = {39}
}

\clearpage
\appendix

\noindent This is the supplementary for \textbf{A Multi-Annotator Study of Segmentation Noise and Uncertainty in Turbid Underwater Images}. Here we describe the design choices behind the annotation study (Section~\ref{app:studydesign}), participant statistics (Section~\ref{app:userstats}), evaluation metrics (Section~\ref{app:evalmetrics}) and additional results supporting the analysis in the main text (Section~\ref{app:results}). 

\section{Study design}\label{app:studydesign}

\subsubsection{Dataset}

\begin{figure}[htb]
    \includegraphics[width=\linewidth]{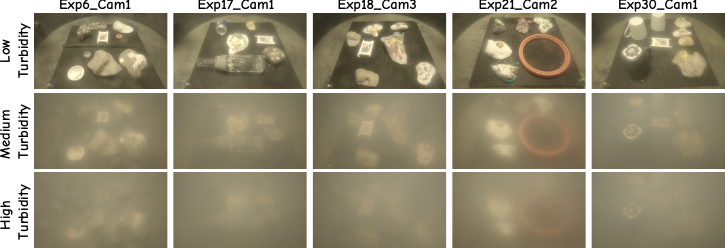} 
    \caption{The 15 TUB~\cite{ismiroglou_beyond_2026} images used in this study (5 scenes at 3 turbidity levels).}
    \label{fig:dataset-imgs}
\end{figure}

\begin{figure}[htb]
\centering
    \includegraphics[width=0.75\linewidth]{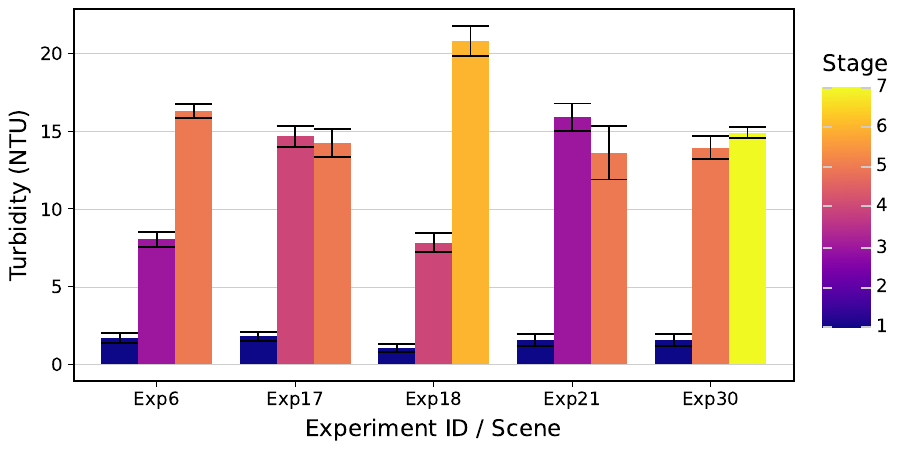} 
    \caption{Turbidity measurements for the 15 TUB~\cite{ismiroglou_beyond_2026} images used in this study (5 scenes at 3 turbidity levels). Measurements are based on TUB metadata, with avg$\pm$std turbidity per image.}
    \label{fig:dataset-ntu}
\end{figure}

\cref{fig:dataset-imgs} shows the images selected for our study. Every participants sees 3 unique images from this selection. \cref{fig:dataset-ntu} shows the turbidity measurements for each scene, which do not necessarily reflect the visual turbidity.

\paragraph{Why 5 scenes?}

For a fixed number of participants, a trade-off exists between the number of unique scenes included in the study and the number of annotations collected per image (3 images per scene). Owing to the group-based study design, the high- and low-turbidity images were expected to receive fewer annotations than the medium-turbidity images. The final selection of five scenes was made with the goal of collecting annotations from approximately 100 participants, corresponding to an expected minimum of roughly 13 annotations per image and about 25 participants per experimental group. This design was deemed sufficient to enable analyses of group-related effects, turbidity-related effects, and variability arising from scene-specific characteristics.

\paragraph{How are images selected?}

The selection of scenes and images was primarily qualitative, as we noticed that the visual appearance/visibility level across images assigned to the same Stage number or with the same turbidity measurement in the TUB~\cite{ismiroglou_beyond_2026} dataset was not necessarily consistent (some images appear much more clear than others). The goal was to ensure consistent visual turbidity levels across all selected scenes. This criterion excluded scenes in which the clearest image was not sufficiently clear or in which the most turbid image did not exhibit a sufficiently high degree of degradation. Additionally, we sought to include a diverse range of object types across the selected scenes.

Specifically, we first arbitrarily chose a reference scene based on the minimum and maximum turbidity requirements, and selected three images representing distinct turbidity levels:
\begin{itemize}
    \item The clearest image with minimal visual degradation.
    \item A moderately turbid image in which some object boundaries became ambiguous and more difficult to discern, but remain mostly identifiable. The visibility is high enough that the edge of the LEGO baseboard remains visible upon close inspection.
    \item A highly turbid image in which some objects remained visible, particularly when viewed at a reduced scale, but object boundaries were severely distorted and more distant objects were effectively obscured. The edge of the LEGO baseboard is no longer visible.
\end{itemize}

The remaining scenes and corresponding images were selected such that they visually matched the three turbidity levels established by the reference scene.

\subsubsection{Participant groups}

\begin{figure}[h]
    \centering
    \includegraphics[height=4cm]{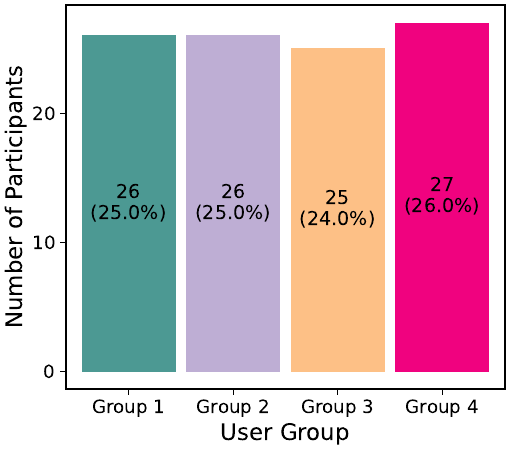}
    \caption{Distribution of the 104 participants across the 4 groups}
    \label{fig:user_group_distribution}
\end{figure}

\paragraph{How are participants assigned to a group?} Participants are assigned to a group in a sequential manner: when a participants starts the study, the number of completed submissions per group is automatically retrieved and the participant is automatically assigned to the group with the least number of submissions. In the case where multiple groups have the same lowest number of submissions, the group is selected randomly among this subset. This assignment procedure ensures an even distribution of completed submissions per group, as shown in Fig.~\ref{fig:user_group_distribution}.

\clearpage

\subsection{Collected data}\label{app:collecteddata}

Alongside the final annotations submitted for every image, we collect the following data:

\paragraph{Automatically collected metadata}

\begin{itemize}
    \item monitor resolution, color depth and window size
    \item clicks and interactions within the annotation interface (e.g. clicking on a button, moving a point, moving a slider)
    \item browser timezone and locale
\end{itemize}

\paragraph{Self-reported data}

\begin{figure}[h]
    \centering
    \includegraphics[width=0.9\linewidth]{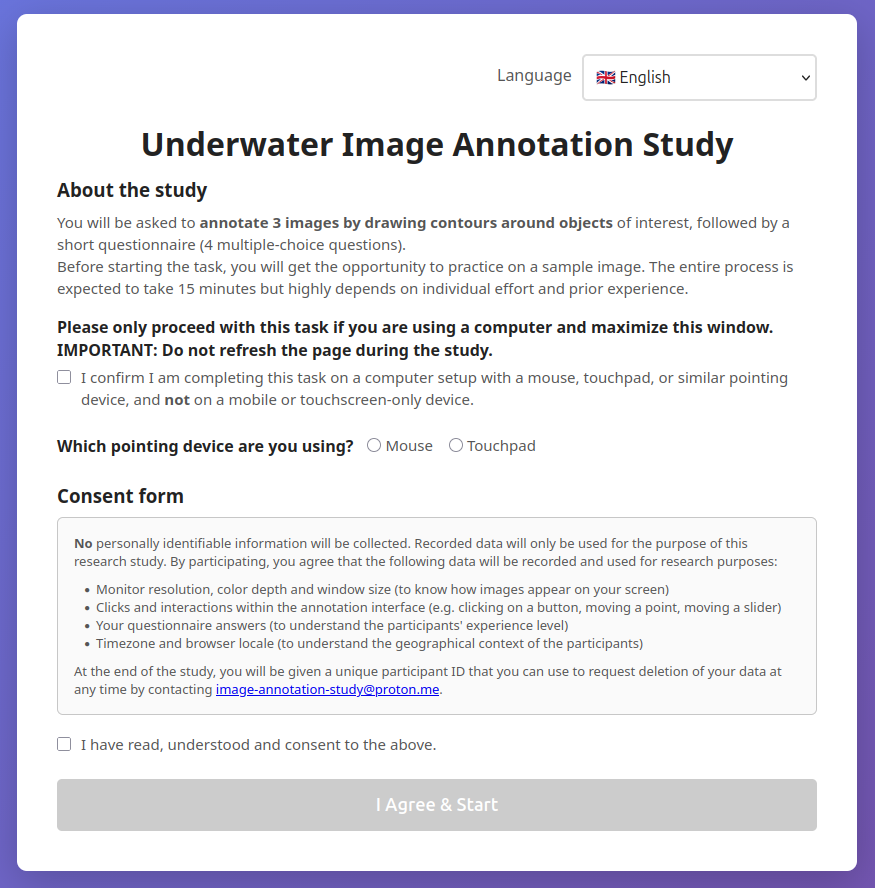}
    \caption{Screenshot of the starting screen}
    \label{fig:screenshot-start}
\end{figure}

\begin{figure}[h]
    \centering
    \includegraphics[width=0.9\linewidth]{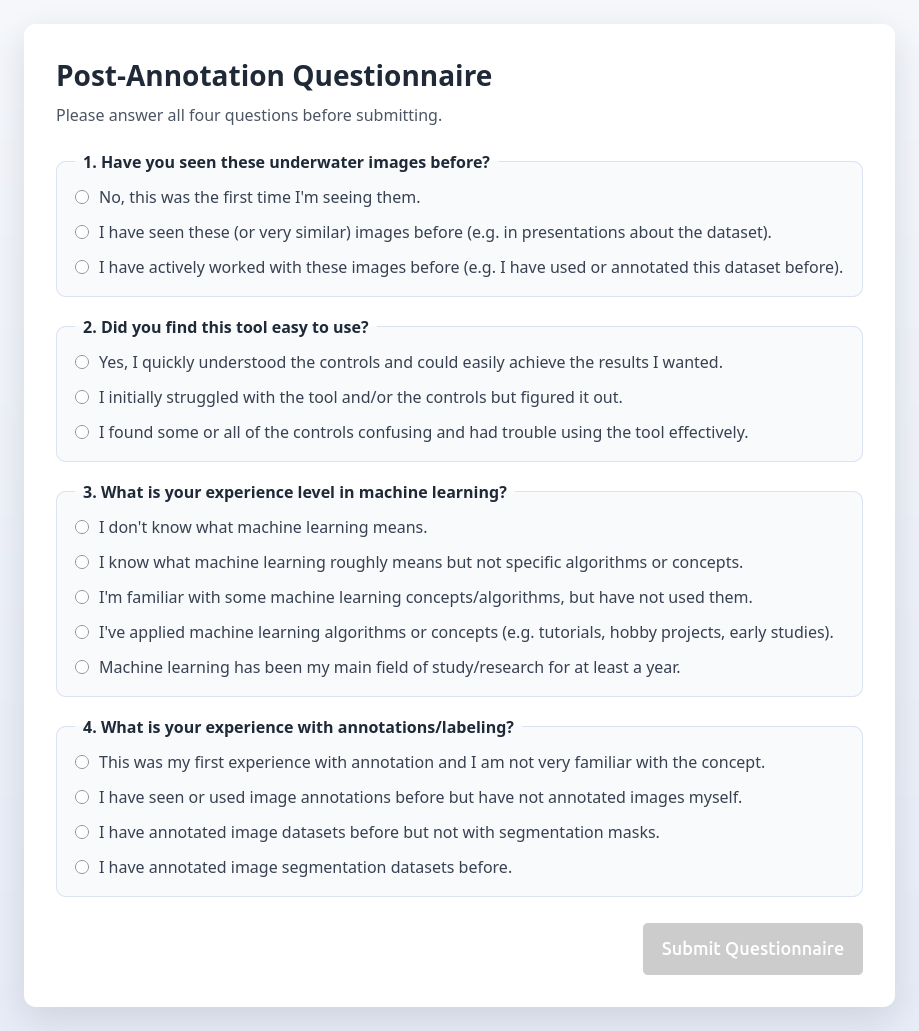}
    \caption{Screenshot of the post-annotation questionnaire}
    \label{fig:screenshot-questionnaire}
\end{figure}

\begin{itemize}
    \item selected language - cf. Fig.~\ref{fig:screenshot-start}
    \item pointing device (touchpad vs. mouse)  - cf. Fig.~\ref{fig:screenshot-start}
    \item post-annotation questionnaire answers - cf. Fig.~\ref{fig:screenshot-questionnaire}
\end{itemize}

\subsection{Practice image}\label{app:practice-page}

\begin{figure}[h]
    \centering

    \textit{The user is first introduced to the practice page and encouraged to play around with the tool. The first instruction box then automatically opens for the user to read.}
    
    \includegraphics[width=0.49\linewidth]{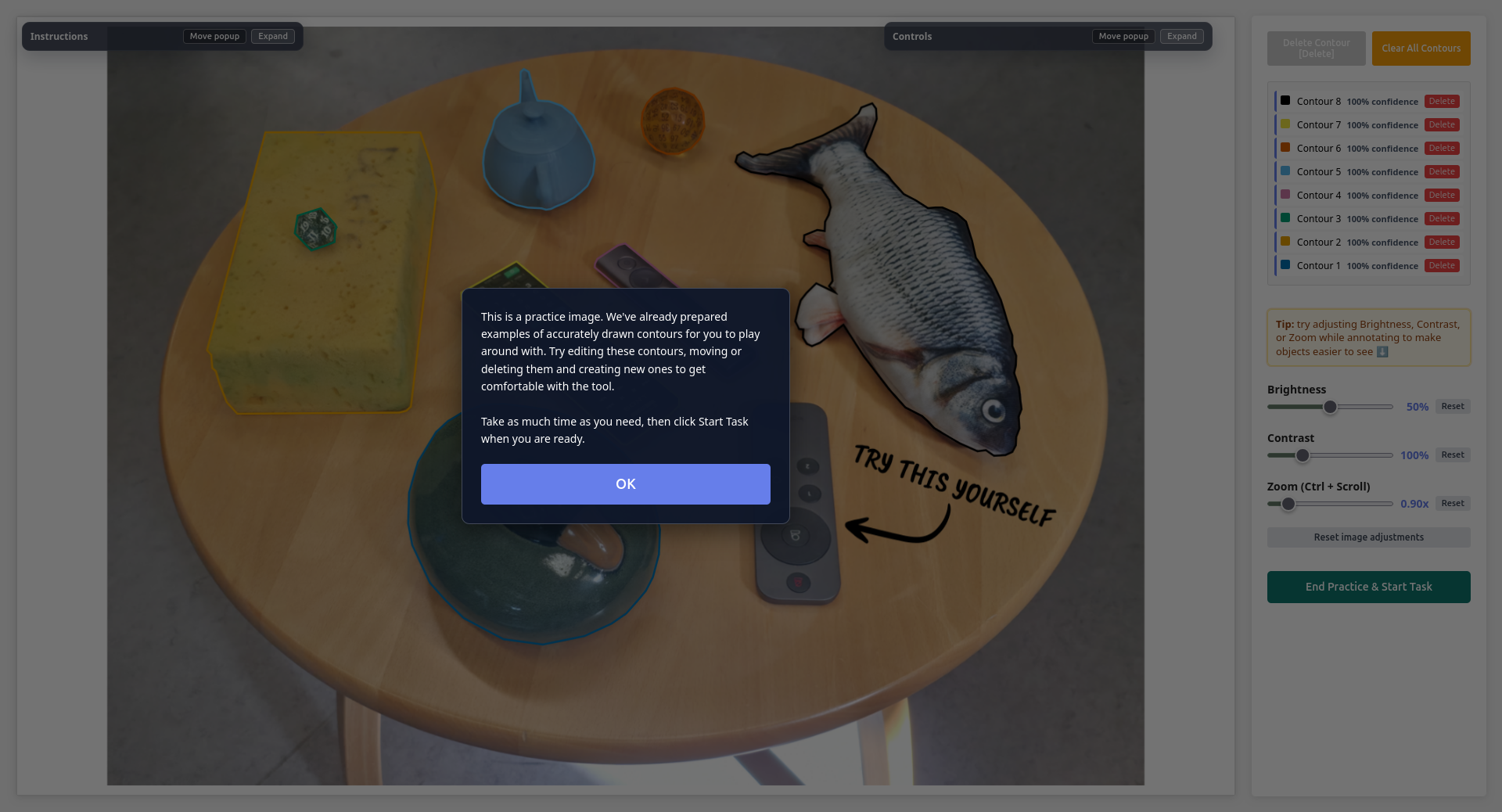}\hfill\includegraphics[width=0.49\linewidth]{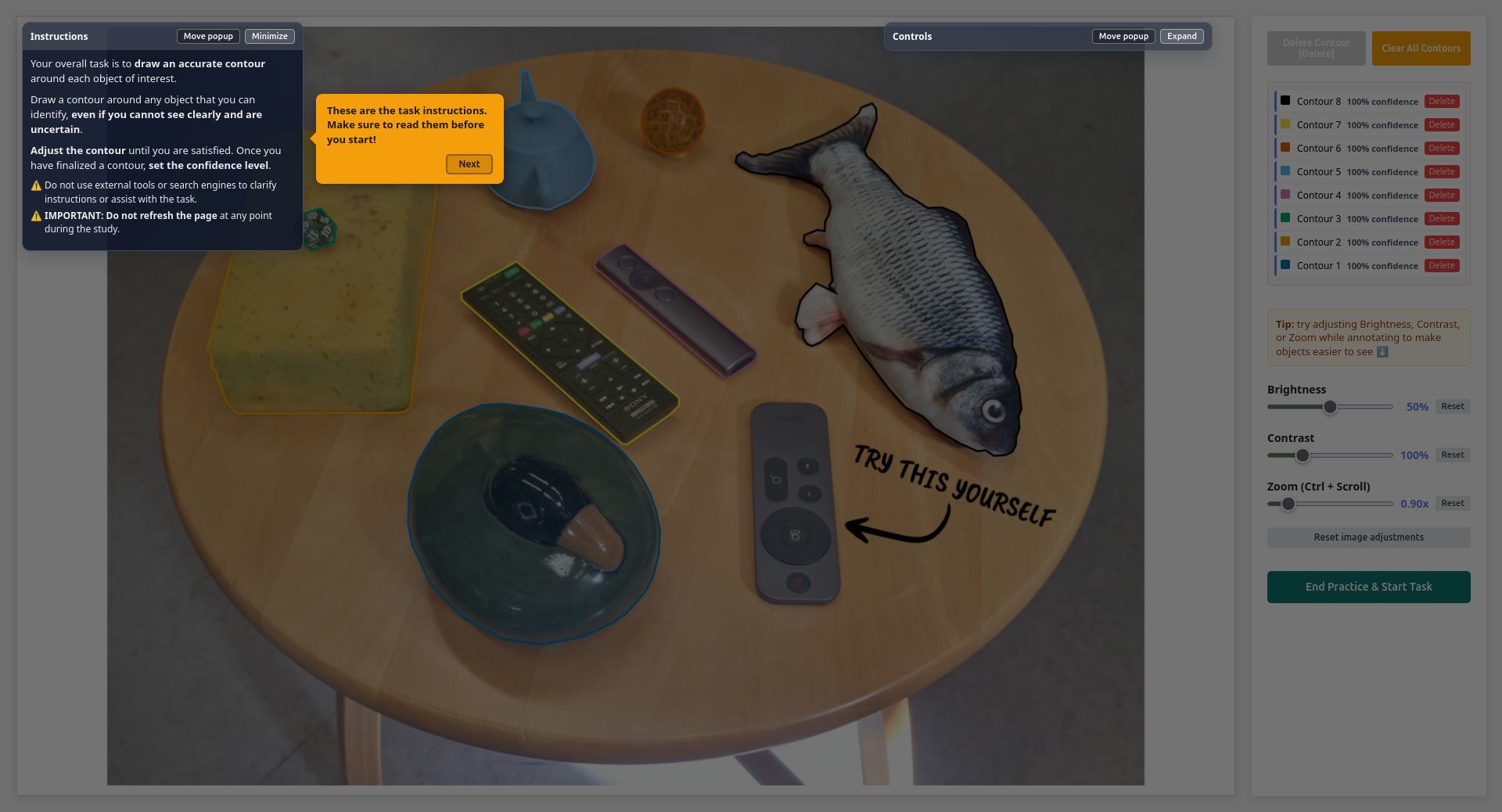}

    \vspace{2em}

    \textit{The second instruction box (explaining controls and shortcuts) then automatically opens as well, leaving both boxes visible. A final tooltip explains that these boxes can be minimized/maximized and moved at anytime.}

    \includegraphics[width=0.49\linewidth]{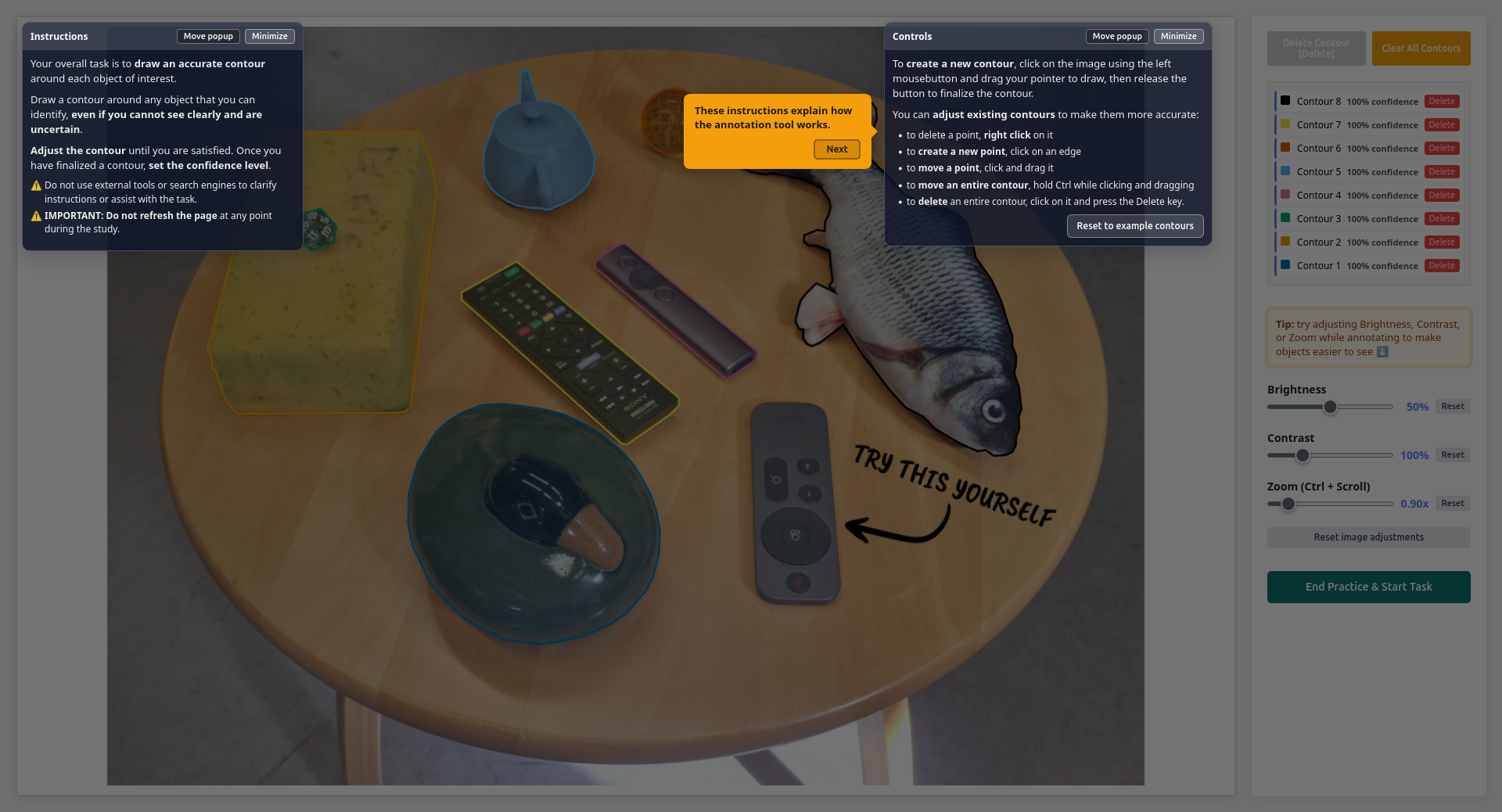}\hfill\includegraphics[width=0.49\linewidth]{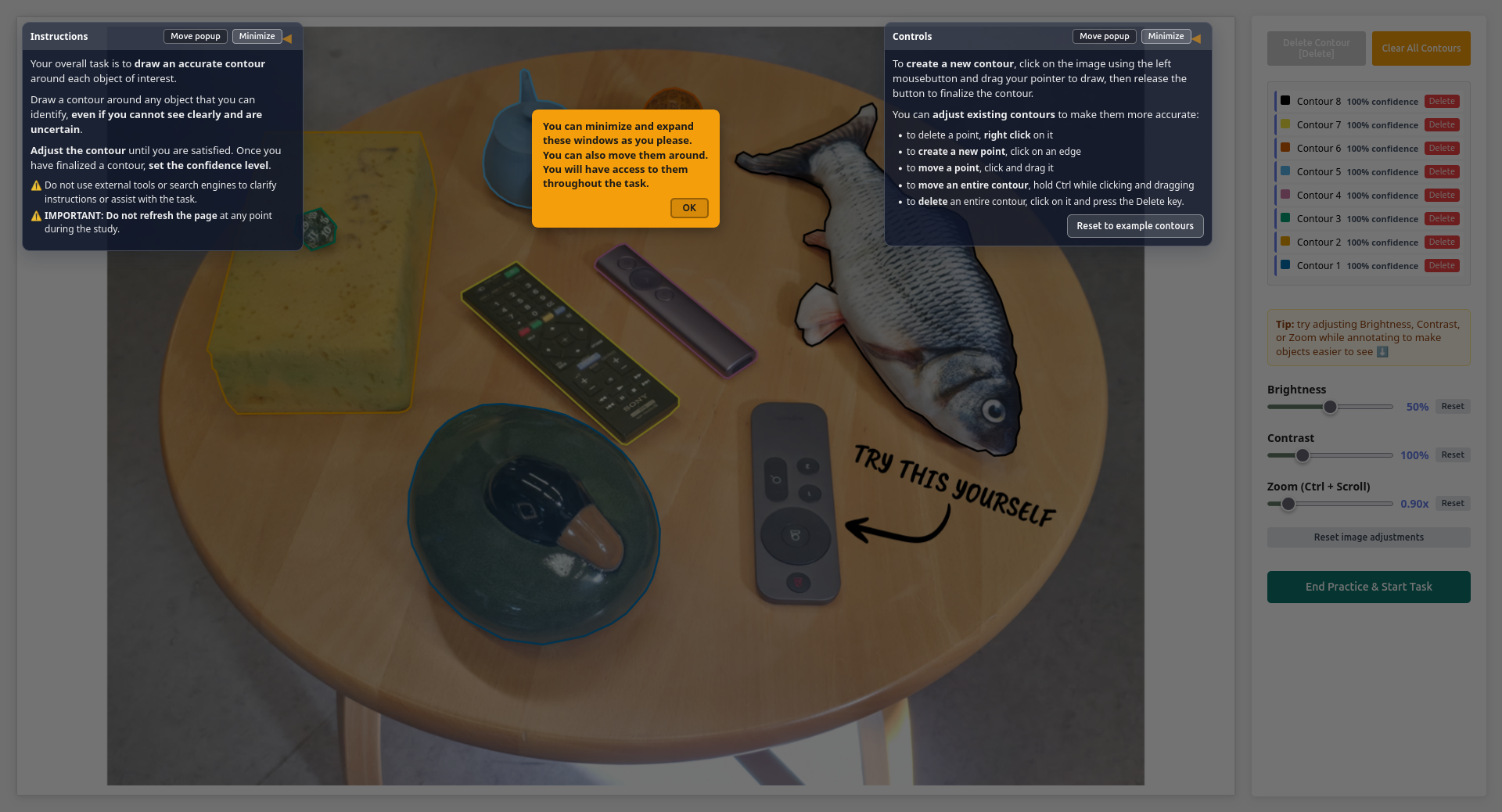}

    \vspace{2em}

    \textit{The user can then start using the tool as they please - either creating new annotations or editing/deleting the already provided ones.}

    \includegraphics[width=0.5\linewidth]{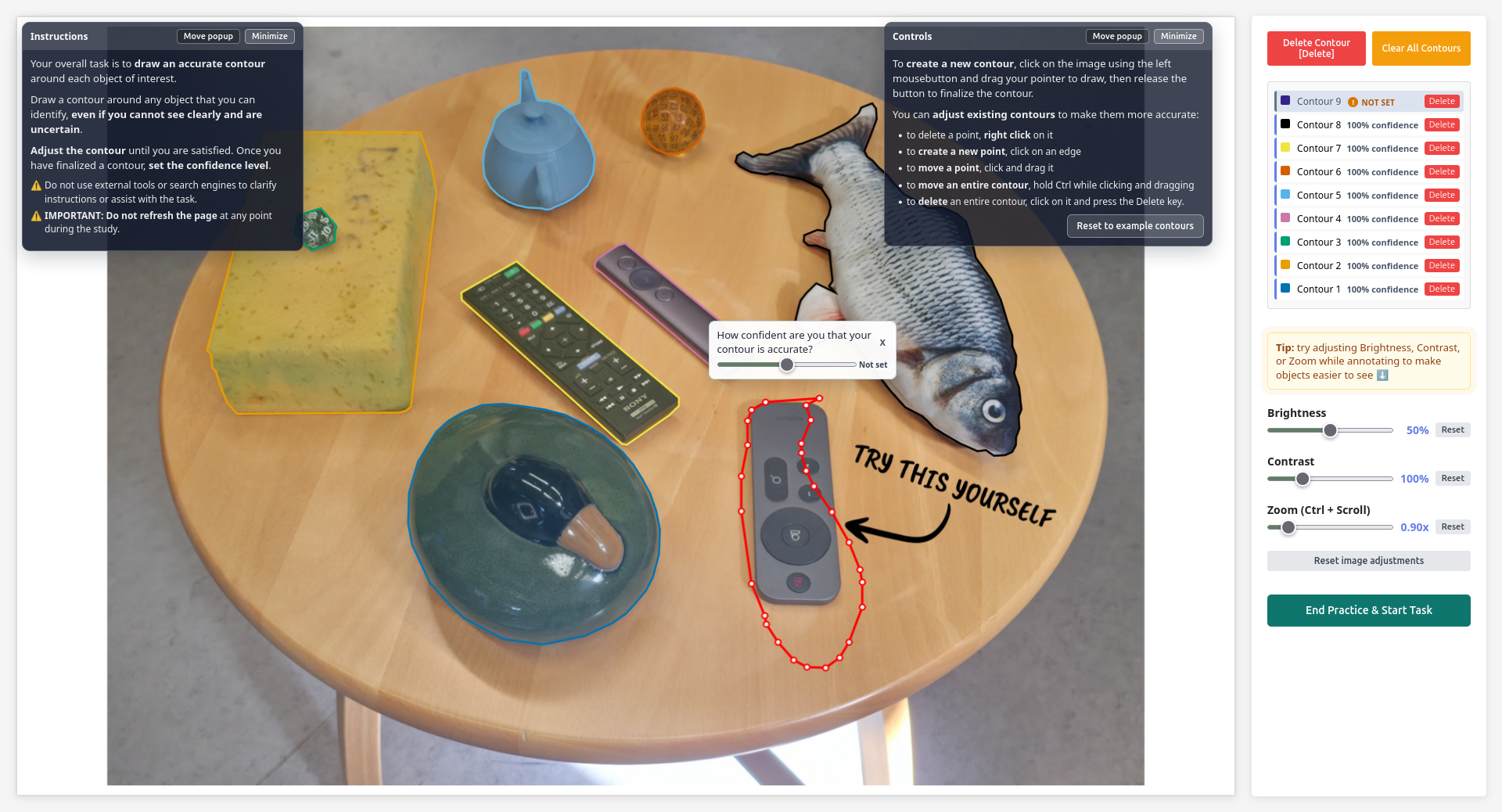}\hfill\includegraphics[width=0.5\linewidth]{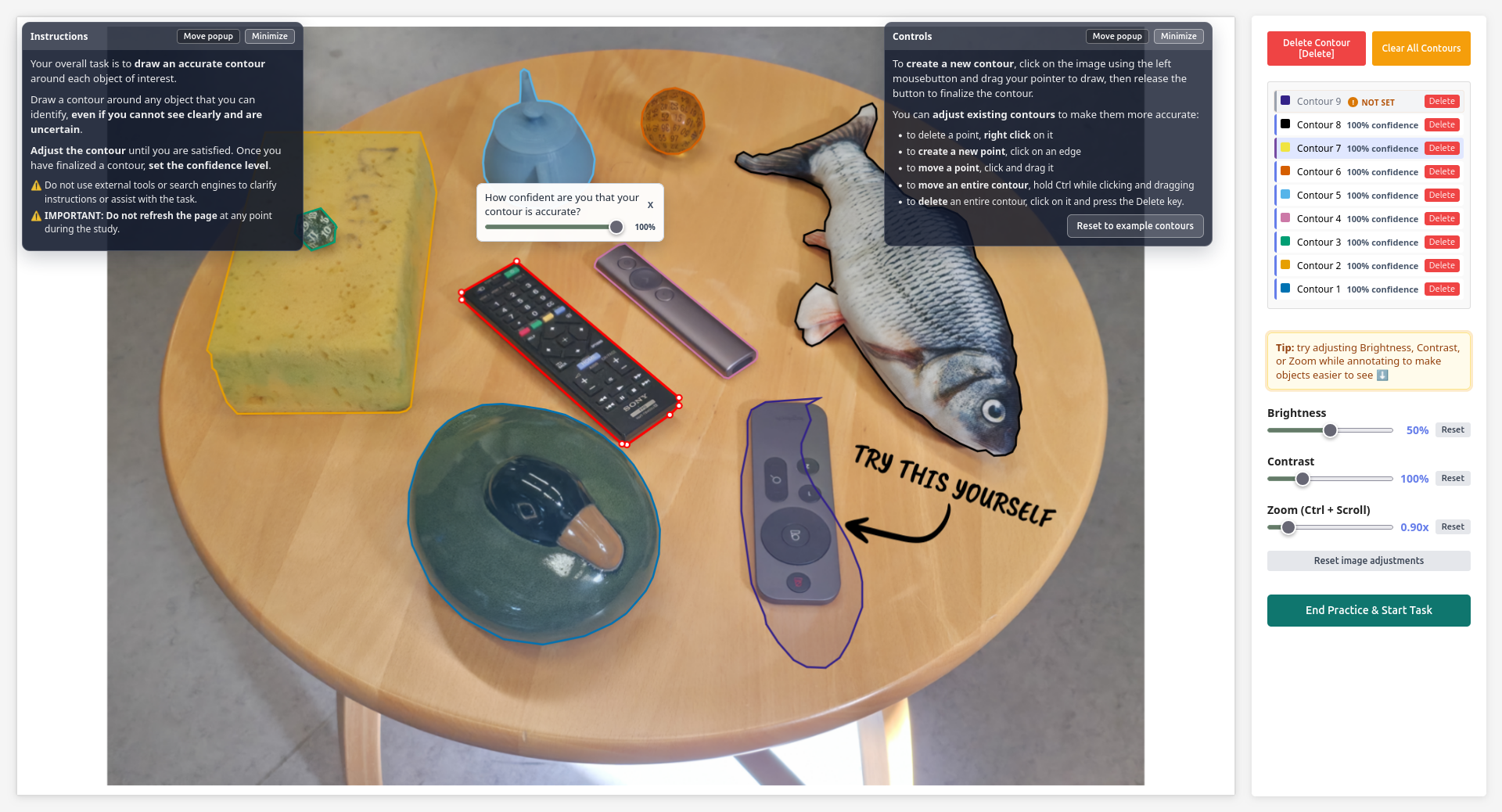}
    
    \caption{Ordered screenshots and explanations of the practice page flow.}
    \label{fig:practicepage-screenshots}
\end{figure}

Before the real annotation task begins (3 underwater images), all users are first presented with a practice page designed to help them become familiar with the interface, annotation tool, and task. As practice image, we chose to create a scene which shares some characteristics with the real task (random objects placed on a single large flat surface), but is taken above water and with different objects and different background. As reference for what is expected in terms of annotation, we annotated all the objects in the practice scene ourselves except one object which we explicitly leave for the user to try to annotate. When the practice page is opened, the user is nudged to read both instruction boxes and annotate this remaining object. Figure~\ref{fig:practicepage-screenshots} shows the information to the user at each step, along with two example interactions (annotating the suggested object, and selecting an existing annotation).

\clearpage

\subsection{Additional considerations and design choices}
The web application, its instructions, and all presented content were carefully designed. A key concern was that annotations of highly turbid scenes could be strongly influenced by prior expectations about scene structure or by assumptions regarding which objects should be present. To reduce such biases, participants were not provided with a predefined list of object categories, and were instead instructed to annotate “objects of interest” and "any objects they can identify". 

To further limit dataset-specific preconceptions (which would influence our analysis of image order), participants were not explicitly exposed to any images from the TUB dataset prior to the main task. Instead, a tutorial image with a similar structural composition was provided. Importantly, this image was captured in-air rather than underwater, ensuring that participants understood the annotation objective without being biased by the visual characteristics of turbid scenes. The tutorial also implicitly clarified the notion of “objects of interest” by including visible elements that were intentionally not meant to be annotated (e.g., background surfaces such as a table), mirroring the distinction we sought between salient objects and scene background in the TUB dataset. However, because the TUB dataset is publicly available, some participants were already familiar with it before the study. We therefore included an explicit question regarding prior exposure to the dataset in the post-annotation questionnaire.

We also considered which image-adjustment sliders should be included, and whether or not their values should be reset for each image. Many image adjustments may have a neutral effect on image visibility, making certain slider settings effectively arbitrary. Furthermore, the usefulness of a given adjustment can vary substantially across images. To encourage deliberate adjustments, all sliders were reset to their default values whenever a new image was presented. This design choice aimed to reduce carry-over effects from previous images and increase the likelihood that any applied adjustments reflected intentional user actions.

Additional design choices were considered for their potential influence on participant expectations. The study title explicitly referenced “underwater,” which may have primed participants to anticipate challenging visual conditions. A time estimate of 15 minutes was provided at the start of the study, although the level of detail in the tutorial annotations may have implicitly suggested a higher annotation effort. Finally, the inclusion of a confidence rating mechanism along with the instruction to annotate "even when one cannot see clearly and is uncertain" emphasized that uncertainty was expected in the task and potentially guided participants to annotate things they would not have otherwise.

\clearpage
\section{Participant statistics}\label{app:userstats}

The link to the study was shared widely through personal networks, professional channels and on social media, with users participating on a voluntary basis. Users were encouraged to participate regardless of their background or experience level. 

\begin{figure}[h]
    \centering
    \includegraphics[width=0.95\linewidth]{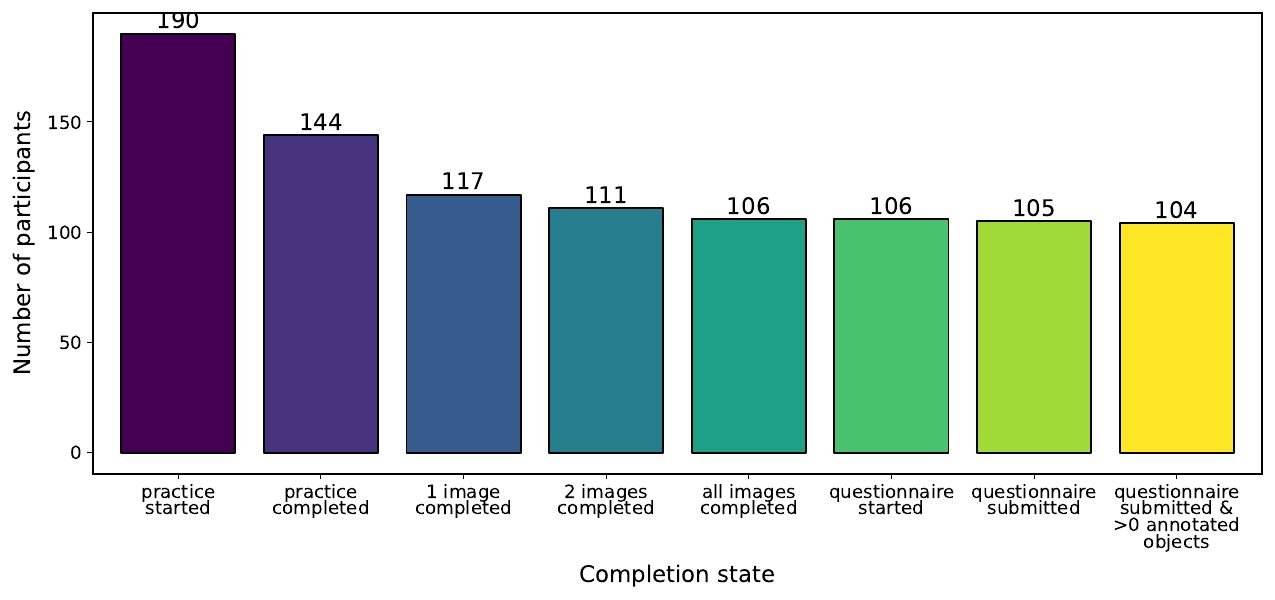}
    \caption{How far did participants get? 190 participants started the study, while 104 fully completed it with non-empty annotations. Results and analysis are based on these 104 submissions.}
    \label{fig:completion_state_barplot}
\end{figure}

As shown in Figure~\ref{fig:completion_state_barplot}, in total, 144 users submitted annotations for the practice image (Image 0), 117 for Image 1, 111 for Image 2 and 106 for Image 3. Ultimately, 106 completed the study by submitting answers to the post-annotation questionnaire. 1 user completed the study but submitted zero annotations for all 3 TUB images, and is thus excluded. Unless explicitly stated, our analysis is based on the 104 complete and non-empty submissions.

\begin{figure}[h]
    \centering
    \includegraphics[width=0.7\linewidth]{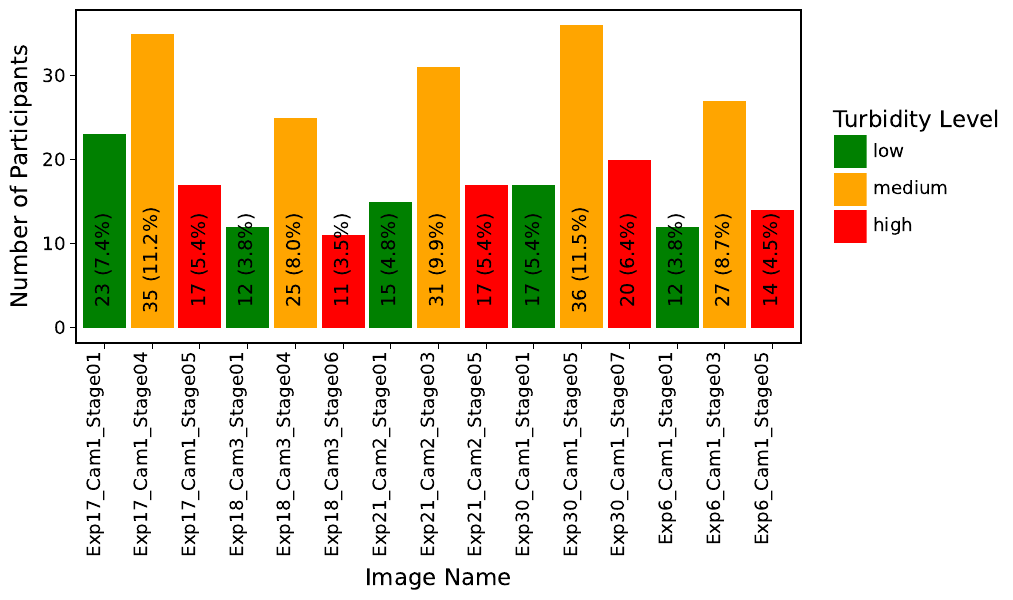}
    \caption{Number of annotations obtained for each image. Note that each annotator is assigned to three different images, and that \groupthree, sees 3 medium turbidity images (vs. one image per turbidity level for the other groups) - hence the higher of number of annotations per medium turbidity image.}
    \label{fig:image-name-dist}
\end{figure}

In total, we obtained 312 submitted annotated images (104 participants $\times$ 3 images per participant). The number of submissions per image is shown in Fig.~\ref{fig:image-name-dist}, and ranges from 11 to 36.

\begin{figure}[h]
    \centering
    
    \includegraphics[width=\linewidth]{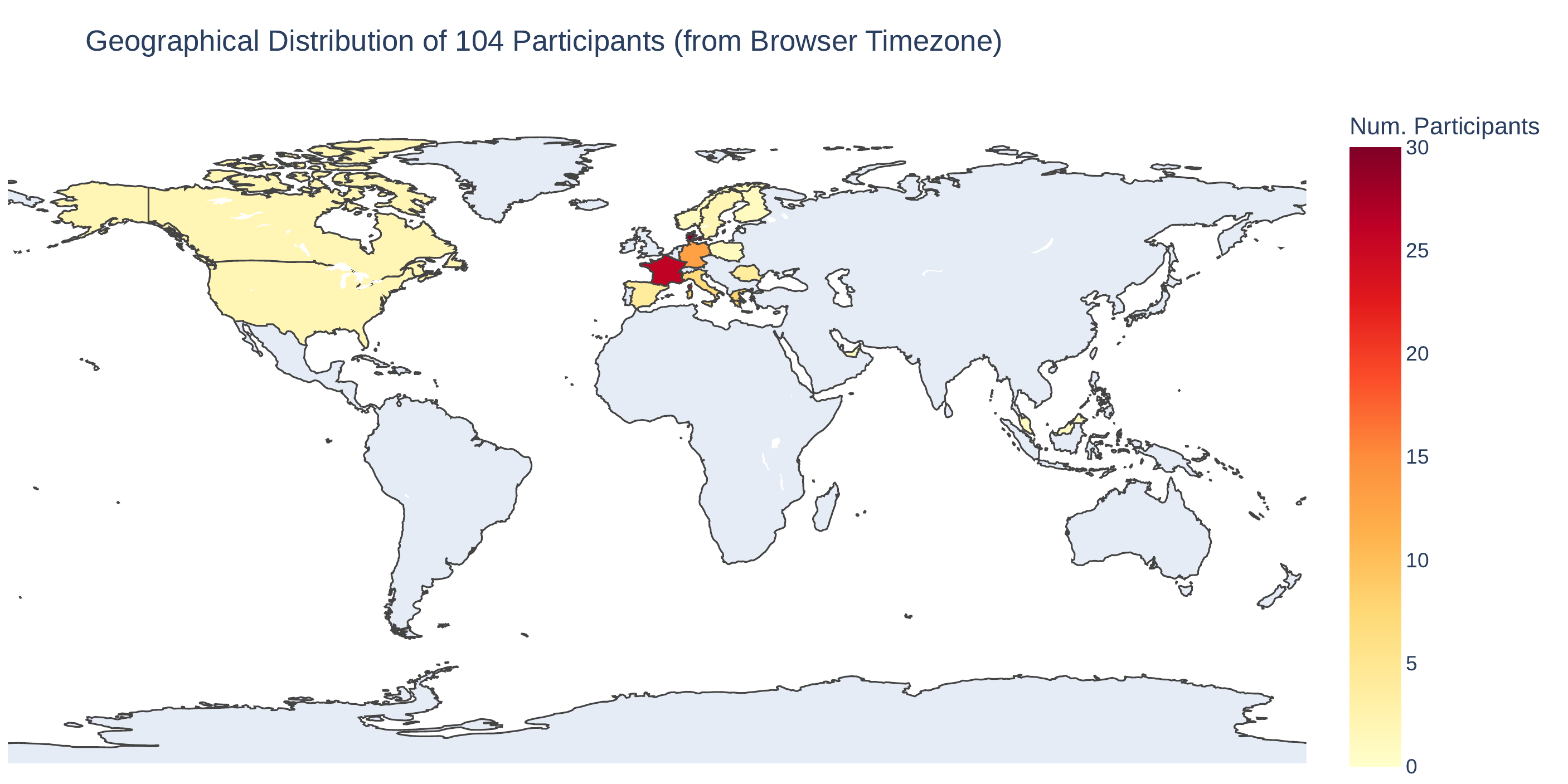}

    \vspace{1em}

    \includegraphics[width=\linewidth]{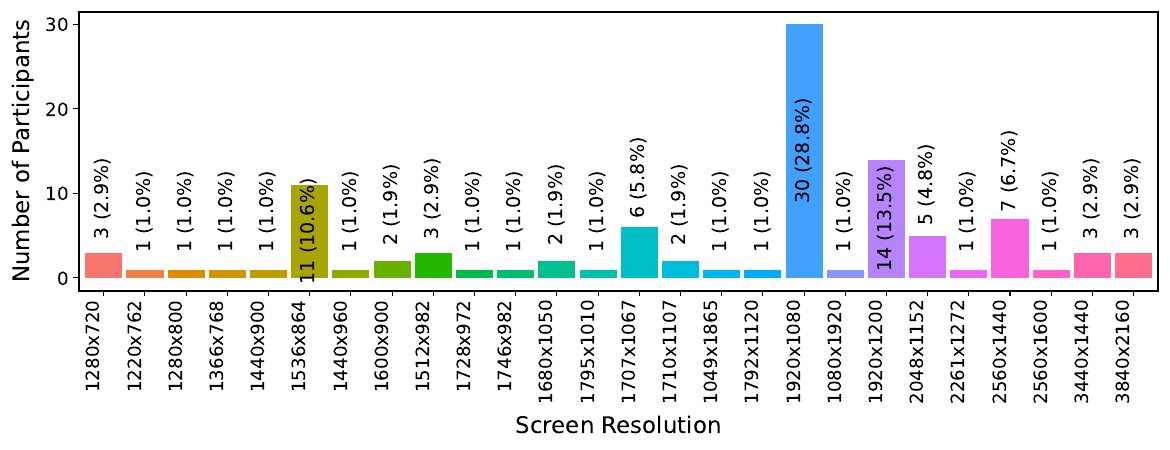}

    \caption{Participants' browser timezone and screen size  (104 participants).}
    \label{fig:userstats-timezone-screen}
\end{figure}

\begin{figure}[h]
    \centering

    \includegraphics[height=4cm]{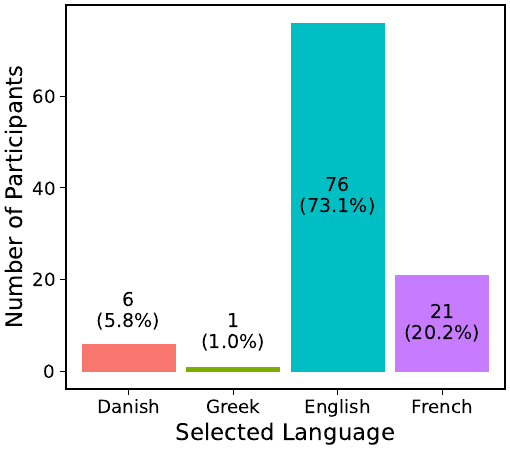}\hspace{2em}
    \includegraphics[height=4cm]{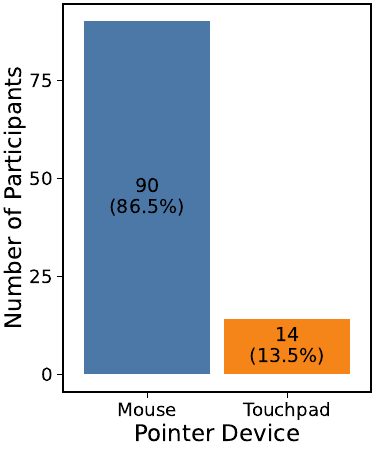}
    
    \vspace{1em}
    
    \includegraphics[height=4cm]{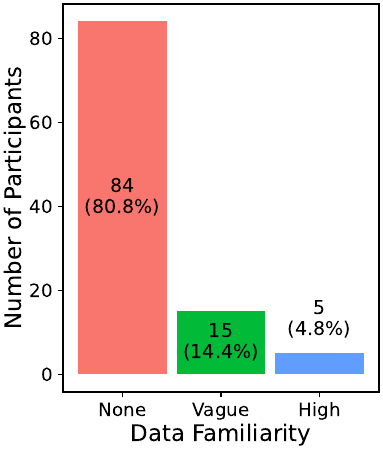}\hspace{2em}
    \includegraphics[height=4cm]{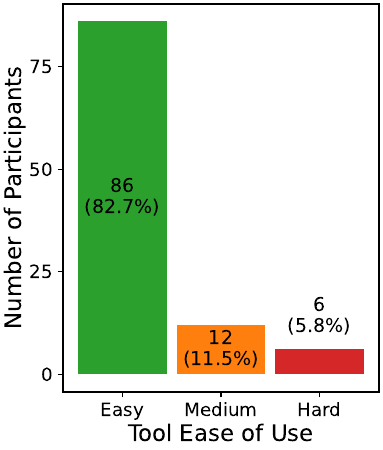}

    \vspace{1em}
    
    \includegraphics[height=4cm]{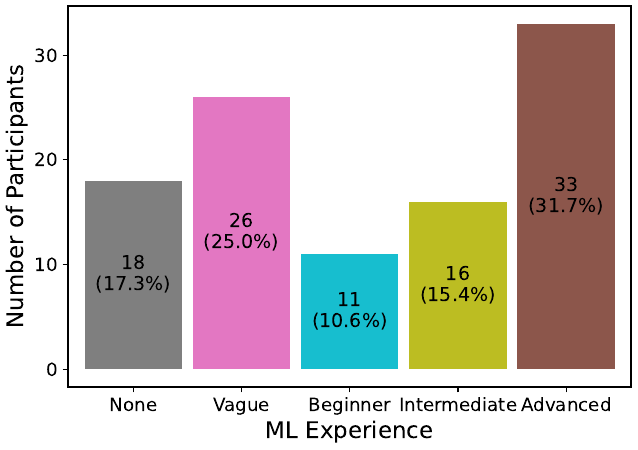}\hspace{2em}
    \includegraphics[height=4cm]{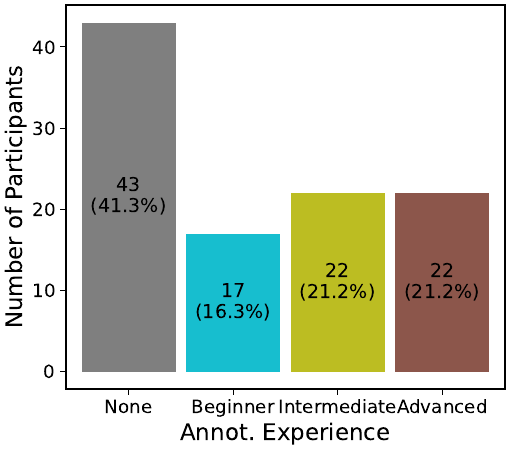}
    \caption{Participants' selected options (language, pointer device and post-annotation questionnaire) (104 participants).}
    \label{fig:userstats-mcq}
\end{figure}

As shown in Fig.~\ref{fig:userstats-timezone-screen} and Fig.~\ref{fig:userstats-mcq}, users participated in the study from 16 different countries (inferred from browser timezone) and using a variety of computer devices. 13.3\% reported using a touchpad vs. 86.7\% using a mouse, with screen resolutions varying from 1280$\times$720 to 3840$\times$2160.

Based on the post-annotation questionnaire, a large majority found the annotation tool easy to use (81.9\%), while 7 users reported having trouble using the tool effectively. In terms of prior knowledge and experience, 53.3\% reported having little to no experience with machine learning, while 32.4\% have machine learning as their main field of study/research. Similarly, 58.1\% reported having no prior experience with image annotation, while 21\% had annotated image segmentation datasets. Lastly, 81\% of participants had never seen the TUB (or similar) images before, and 5 had worked with TUB images before.

\clearpage

\section{Evaluation metrics}\label{app:evalmetrics}

\subsection{Instance segmentation} We evaluate annotated contours according to two matching protocols \textit{GT$\rightarrow$annotators} and \textit{annotators$\rightarrow$GT}. In both protocols, for a given annotated image (from a single annotator), annotated contours and GT contours from the corresponding image are first converted to individual binary masks:
\begin{equation}
     A={a_1,\ldots,a_m}, \qquad G={g_1,\ldots,g_n},
\end{equation}
Both matching protocols use the IoU between an annotated mask $A_i$ and a GT mask $G_j$ as matching criteria:
\begin{equation}
    \mathrm{IoU}(a_{i}, g_{j}) = \frac{|a_{i} \cap g_{j}|}{|a_{i} \cup g_{j}|}
\end{equation}

\subsubsection{\textit{GT$\rightarrow$annotators}} For each annotated mask $a_i\in A$, we compute its IoU with every GT mask $g\in G$. The matching GT mask is selected as
\begin{equation}
g^*=\arg\max_{g\in G}\mathrm{IoU}(a_i,g),
\end{equation}
and the corresponding score $\mathrm{IoU}(a_i,g^*)$ is assigned to $a_i$. Thus, \textbf{every annotated contour in the image receives an IoU score}.

\subsubsection{\textit{annotators$\rightarrow$GT}} For every GT mask $g_{j} \in G$, we compute its IoU with every annotated mask $a_i\in A$. The matching annotated mask is selected as
\begin{equation}
a^*=\arg\max_{a\in A}\mathrm{IoU}(a,g_j),
\end{equation}
and the corresponding score, $\mathrm{IoU}(a^*,g_j)$ is assigned to $g_{j}$. Thus, \textbf{every GT contour in the image receives an IoU score}.\\

\noindent These two matching protocols give us complementary measures. \textit{GT$\rightarrow$annotators} does not penalize annotators for missing an object is present in the GT. \textit{annotators$\rightarrow$GT} does not penalize annotators for annotating an object that is not present in the GT. 

When reporting these two metrics, we report them at the instance-level, without aggregating across contours, images or annotators.

\subsection{Semantic segmentation}

\begin{figure}[h]
    \centering
    \begin{tabular}{C{0.20\linewidth}C{0.05\linewidth}C{0.20\linewidth}C{0.05\linewidth}C{0.20\linewidth}C{0.05\linewidth}C{0.20\linewidth}}
        & &  $M_{A}$ & & $M_{GT}$ & & \\
        \includegraphics[width=\linewidth]{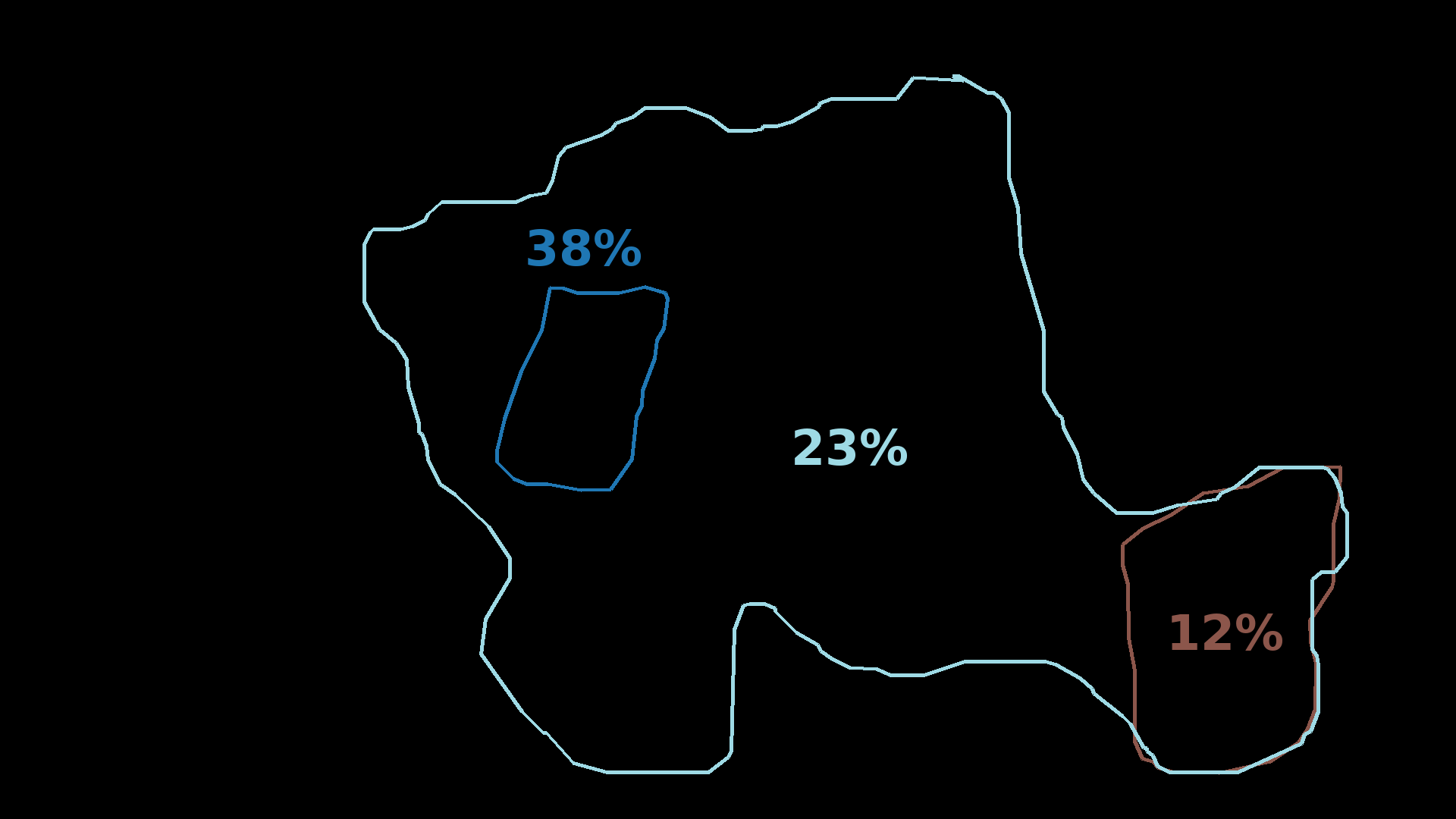} & $\rightarrow$ & \includegraphics[width=\linewidth]{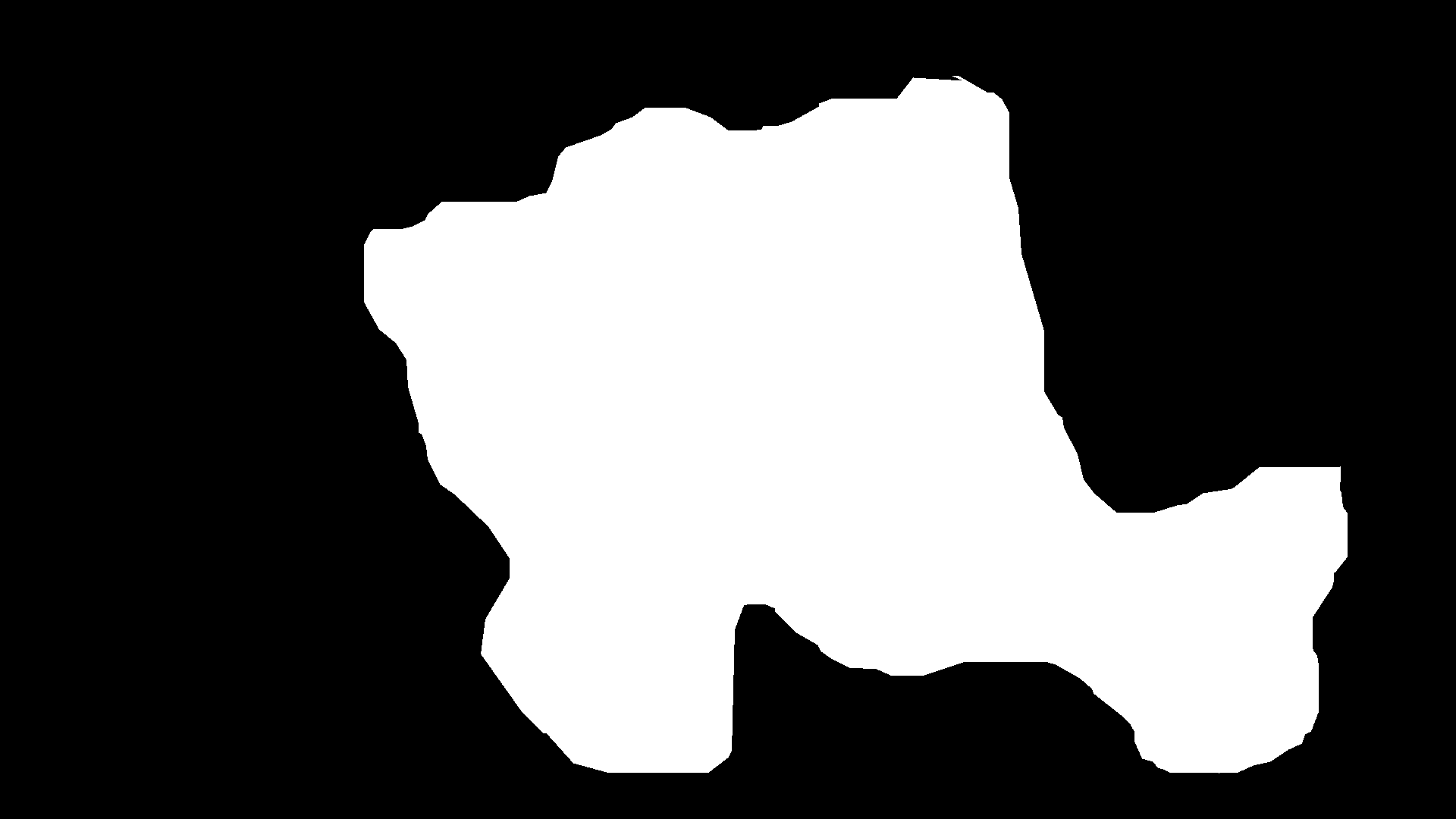} & \textbf{vs.} & \includegraphics[width=\linewidth]{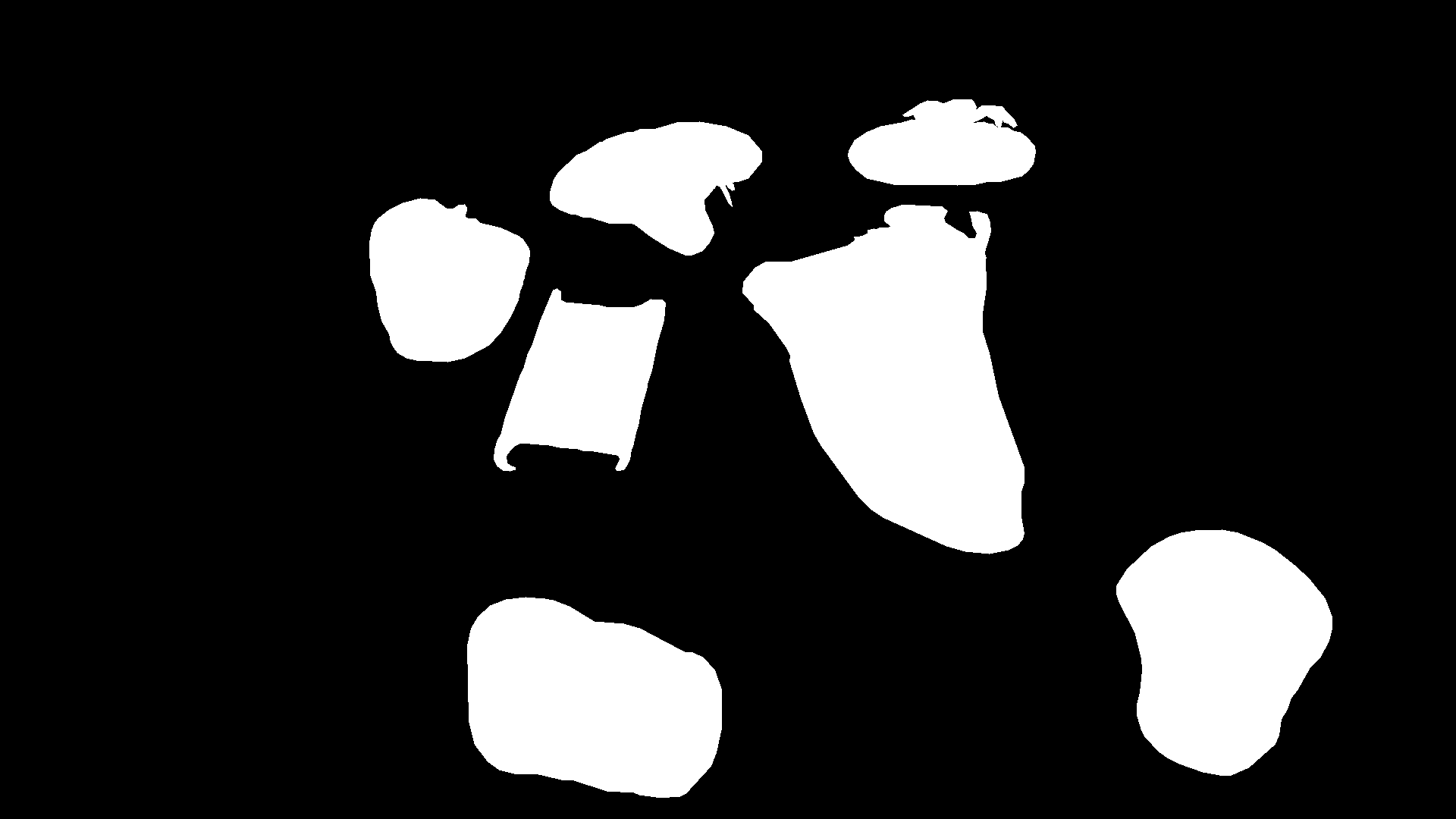} & $\leftarrow$ & \includegraphics[width=\linewidth]{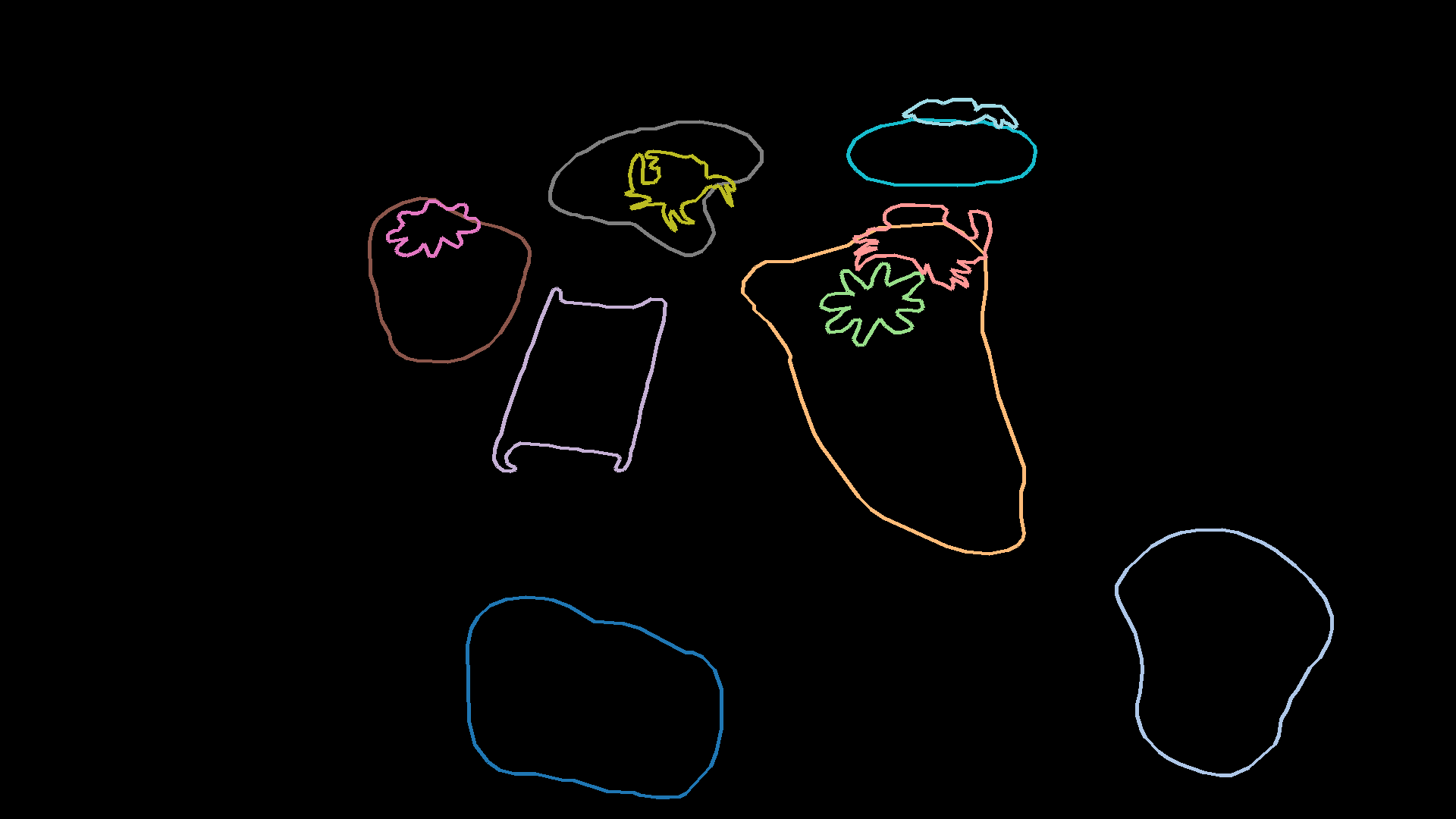} \\
    \end{tabular}
    \caption{Illustration of how an annotator's contours (left) are compared to GT contours (right) in terms of binary semantic segmentation, using an annotation of \texttt{Exp18\_Cam3\_Stage06} as an example.}
    \label{fig:binaryseg-iou-example}
\end{figure}

We also evaluate the annotation task as binary semantic segmentation, at the \textbf{image-level}. For a given annotated image, we map all contours with a confidence $\geq$ a certain threshold $T$ onto a single binary mask $M_A$, where foreground/positive pixels (white) correspond to an annotated shape. Similarly, we map the corresponding GT contours to a binary mask $M_{GT}$, as illustrated in \cref{fig:binaryseg-iou-example}.

As the main metric, we compute the IoU between the two - similarly to above but at the image-level:
\begin{equation}
    \mathrm{IoU}(M_{A}, M_{GT}) = \frac{|M_A \cap M_{GT}|}{|M_A \cup M_{GT}|}
\end{equation}
As secondary metrics, we also compute the Precision and Recall:
\begin{equation}
    \mathrm{Precision}(M_{A}, M_{GT}) = \frac{\mathrm{TP}}{\mathrm{TP}+\mathrm{FP}} = \frac{|M_A \cap M_{GT}|}{|M_A|}
\end{equation}
\begin{equation}
    \mathrm{Recall}(M_{A}, M_{GT}) = \frac{\mathrm{TP}}{\mathrm{TP}+\mathrm{FN}} = \frac{|M_A \cap M_{GT}|}{|M_{GT}|}
\end{equation}

We compute these metrics for every unique annotator per image ($104$ participants $\times$ 3 images per annotator $\rightarrow$ $312$ value for each metric). By default, we use a threshold of $T=0$ in the paper, meaning that \textbf{all} annotated contours in an image are included in the evaluation. We find that this is the best global threshold for individual annotators (cf. Figure~\ref*{fig:threshold-conf-curves} in the main text).

\subsubsection{Inter-annotator agreement}

To measure agreement, we similarly compute binary segmentation metrics, but between a pair of annotations of the same image rather than in reference to a GT mask. Given an image, we consider all $N$ annotators who were assigned this image during the study, $A_1, A_2, \ldots, A_{N}$. We generate an image-level binary mask for each of these annotators $M_{A_1}, M_{A_2} \ldots M_{A_N}$ as described above (confidence threshold $T=0$). For each pair of annotators $(A_j,A_k), j \neq k$ we compute the IoU between the two masks:
\begin{equation}
    \mathrm{IoU}(M_{A_j}, M_{A_k}) = \frac{|M_{A_j} \cap M_{A_k}|}{|M_{A_j} \cup M_{A_k}|}
\end{equation}
We compute this metric for every unique annotator pair per image.

\clearpage
\section{Results}\label{app:results}

\subsection{Qualitative examples}\label{app:results-qual}

\begin{figure}[h]
    \centering

    \includegraphics[width=\linewidth]{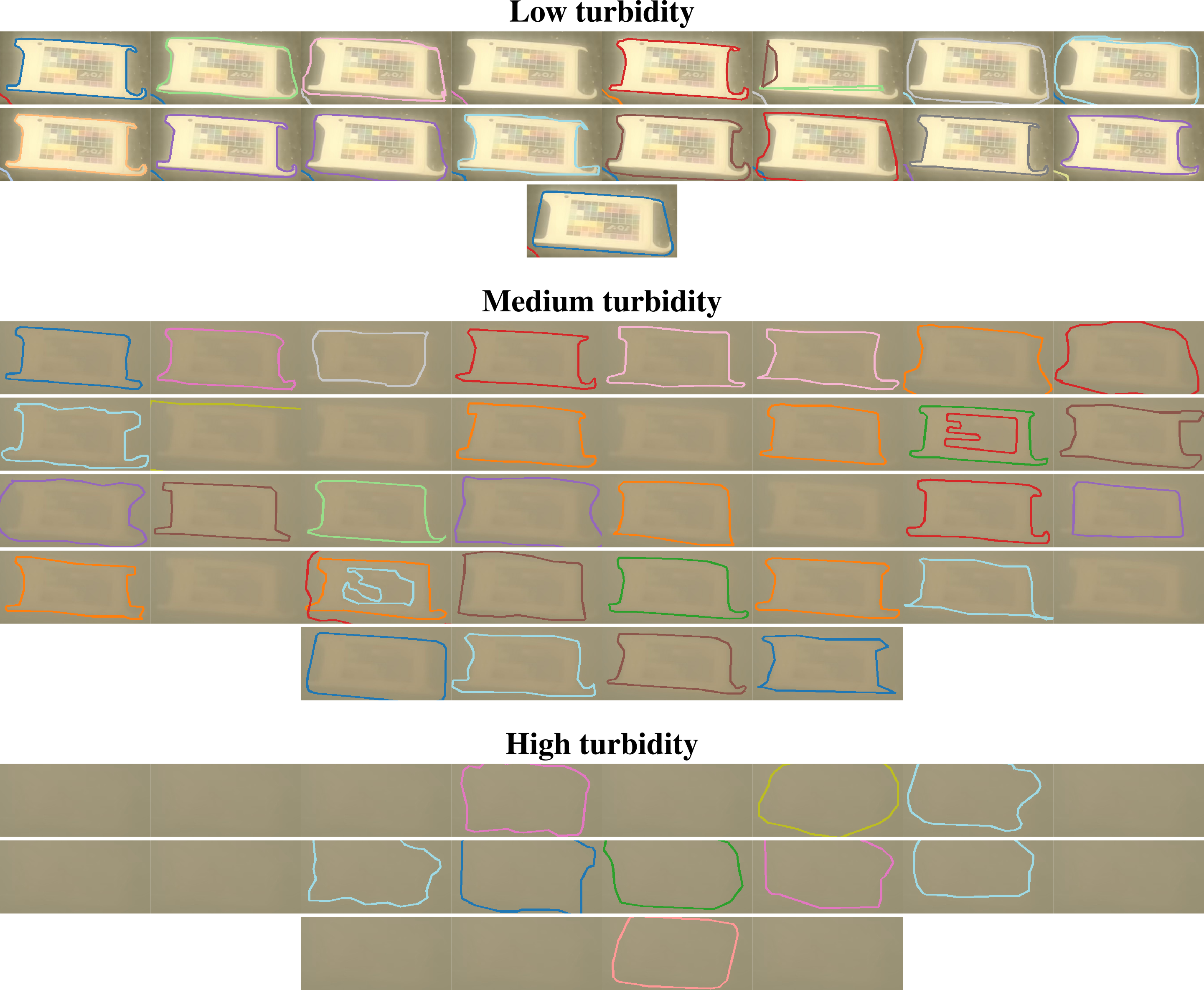}
     
        \caption{All annotations for the same object in the same scene (\texttt{Exp30\_Cam1}), varying turbidity. Some users fail to annotate the object altogether, especially as turbidity increases.}
        \label{fig:checkerboard-annots-more}
    

\vspace{1em}

    \centering
    \includegraphics[width=\linewidth]{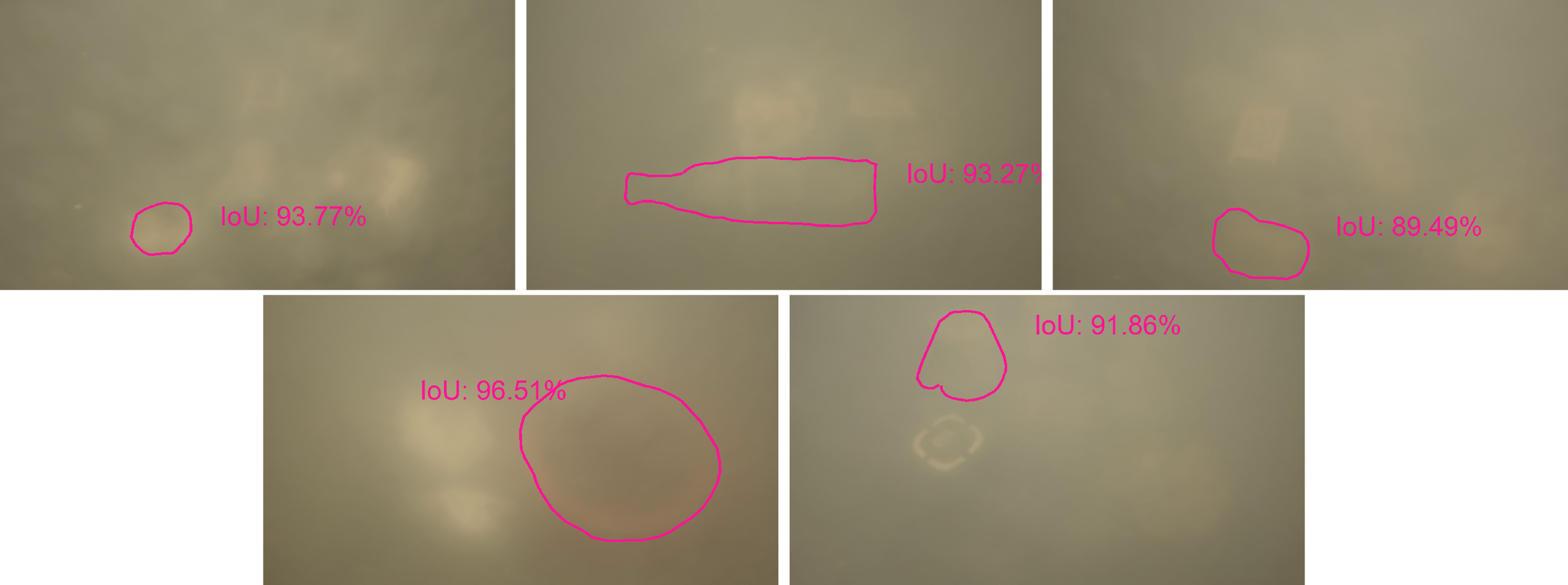}
    
    \caption{Best object annotation (in terms of IoU) per high-turbidity scene: white circular cap, bottle, light-gray rock, orange rubber seal, white mug.}
    \label{fig:highturbid_best}
\end{figure}

\cref{fig:checkerboard-annots-more} zooms in on the same example object as Figure~\ref*{fig:checkerboard-annots} in the main text but showing \textbf{all} annotations per turbidity level. 

\cref{fig:highturbid_best} gives an idea of best-case performance in high turbidity, with the top annotation per highly turbid scene. 

\subsection{Plots}\label{app:results-plots}

\subsubsection{Confidence histograms} 
\begin{figure}[tb]
    \centering
    \includegraphics[height=3cm]{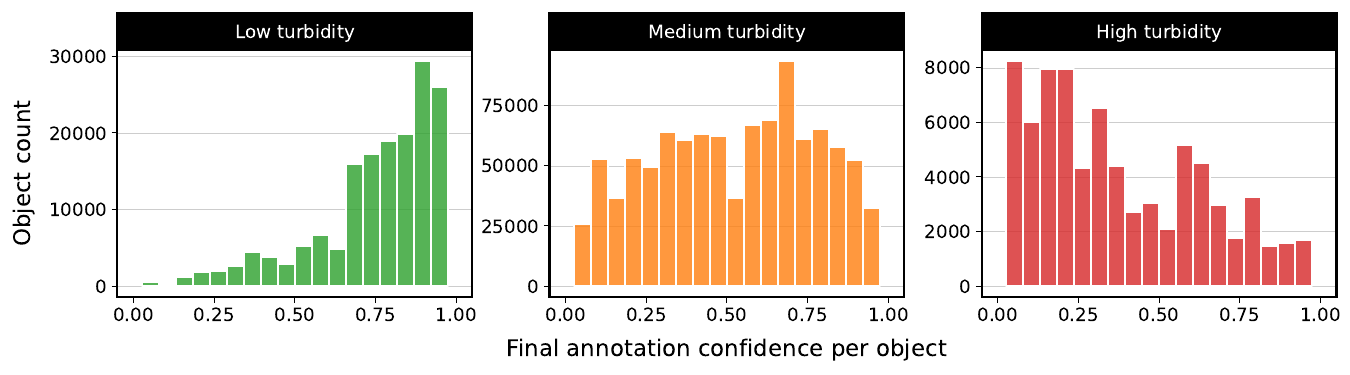}
    \caption{Distribution of object confidence for different turbidity levels.}
    \label{fig:histo_conf_turbidity}

    \vspace{2em}

    \centering
    {\comicneue \textbf{Low turbidity}}
    
    \includegraphics[height=3cm]{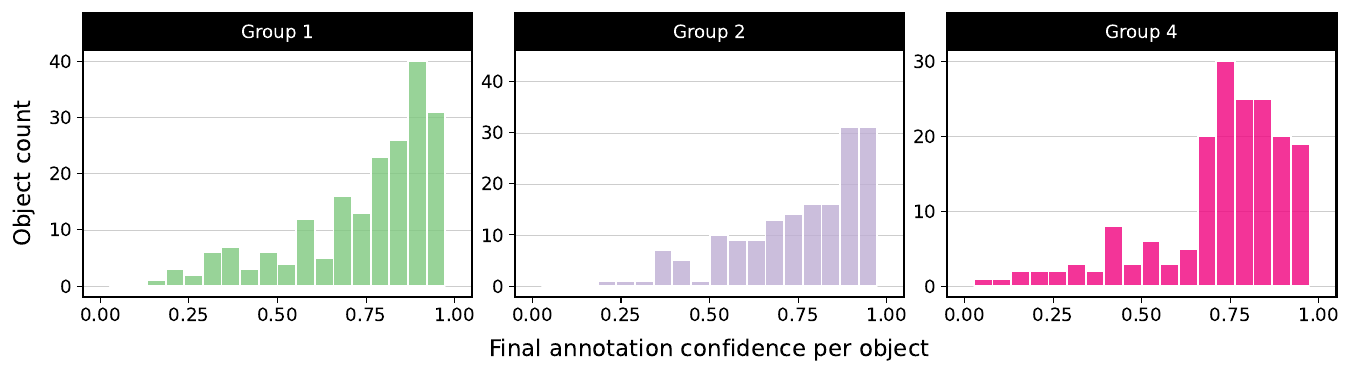}

    \vspace{1em}

    {\comicneue \textbf{Medium turbidity}}
    
    \includegraphics[width=\linewidth]{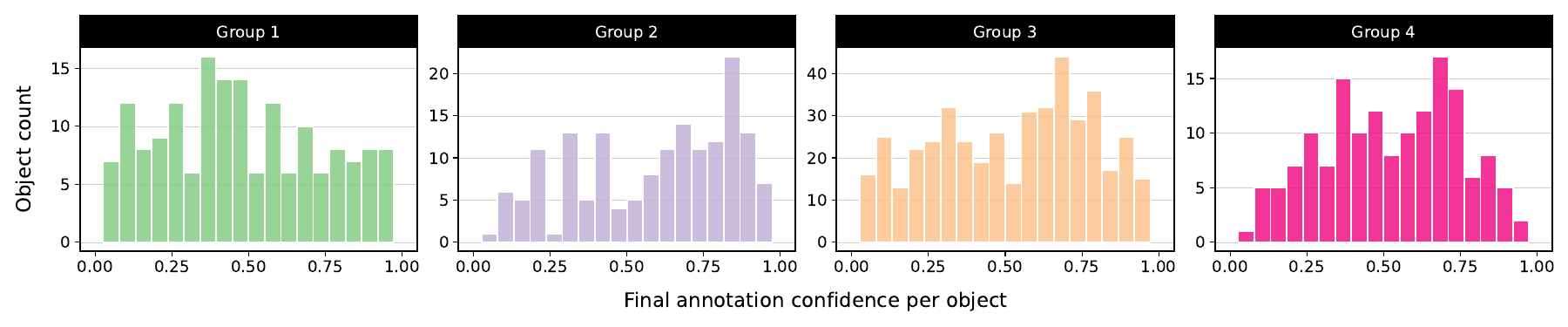}

    \vspace{1em}

    {\comicneue \textbf{High turbidity}}
    
    \includegraphics[height=3cm]{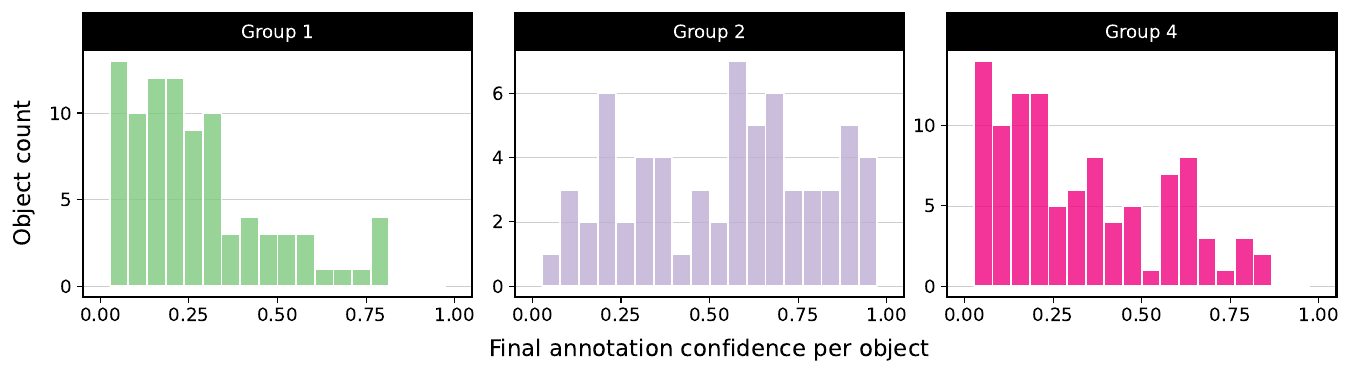}
    \caption{Distribution of object confidence for low, medium and high turbidity images, depending on user group. Note: \groupthree\ only sees medium turbidity images.}
    \label{fig:histo_conf_imagenum}
\end{figure}

\cref{fig:histo_conf_turbidity,fig:histo_conf_imagenum} show the distribution of object confidence as a function of turbidity level (across all groups) and as a function of image number (depending on the group).

\subsubsection{Confidence vs. performance} 
\begin{figure}[tb]
    \centering
    \includegraphics[height=4cm]{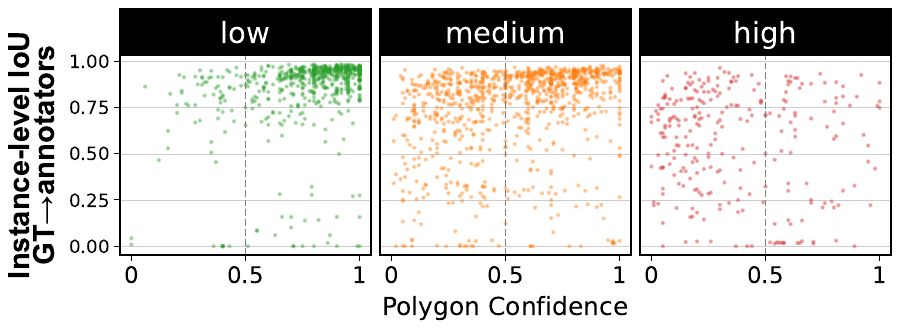}
    \includegraphics[height=4cm]{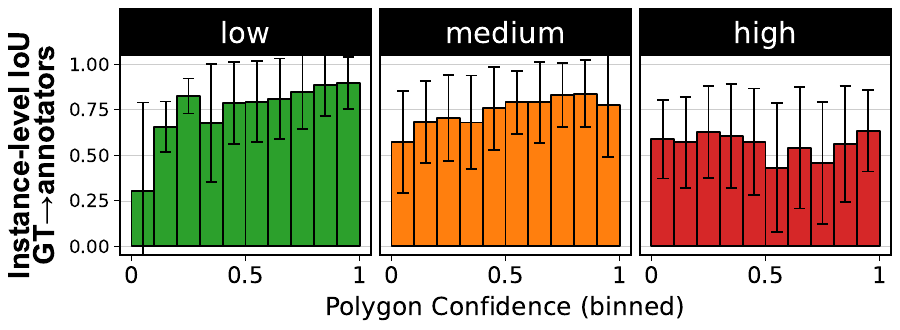}
    \caption{Every point represents an annotated contour/polygon. The top plot is the same as Figure~\ref*{fig:threshold-conf-curves-instance} (main text) but enlarged for convenience. The bottom plot is based on the same data points, but shows the average performance (with error bar) across 10 confidence bins.}
    \label{fig:bar_ious_wrt_annot_conf_by_turbidity}
\end{figure}

\cref{fig:bar_ious_wrt_annot_conf_by_turbidity} shows the instance-level performance as a function of confidence, similarly to Figure~\ref*{fig:threshold-conf-curves-instance} in the main text, but also showing performance for different confidence bins.

\subsubsection{Annotation effort and time}

\begin{figure}[tb]
    \centering
    \includegraphics[height=6cm]{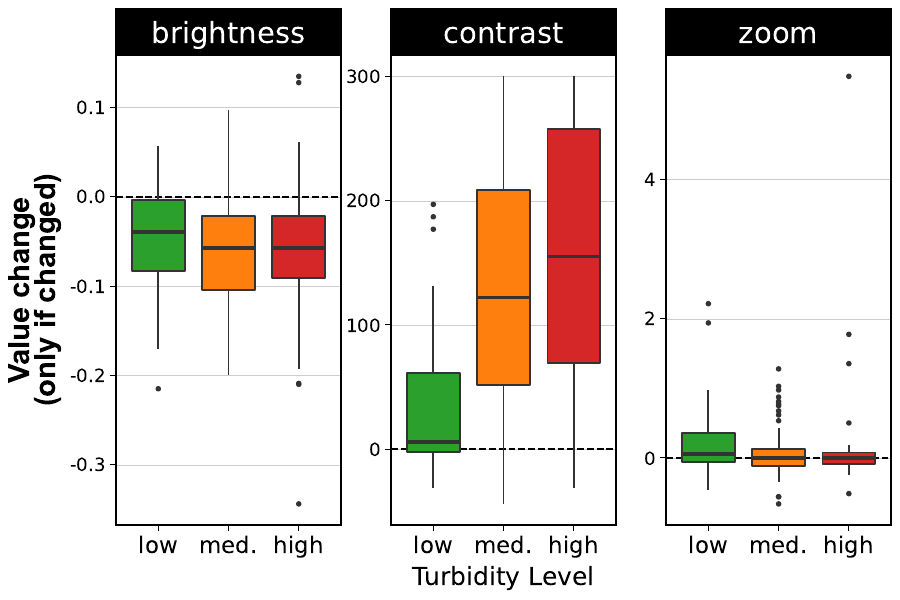}
    \caption{Final change in the value of image adjustment sliders (final value $-$ default value). 0 indicates no change. Each point represents an annotated image, and only images where the value was changed at least once during annotation are included in the plot. Note that the "contrast" sub-plot corresponds to Figure~\ref*{fig:boxplot_image_settings_by_turbidity_level_and_event} in the main text.}
    \label{fig:boxplot_image_settings_changes_by_turbidity_level_extra}
\end{figure}

\begin{figure}[tb]
    \centering
    \includegraphics[height=6cm]{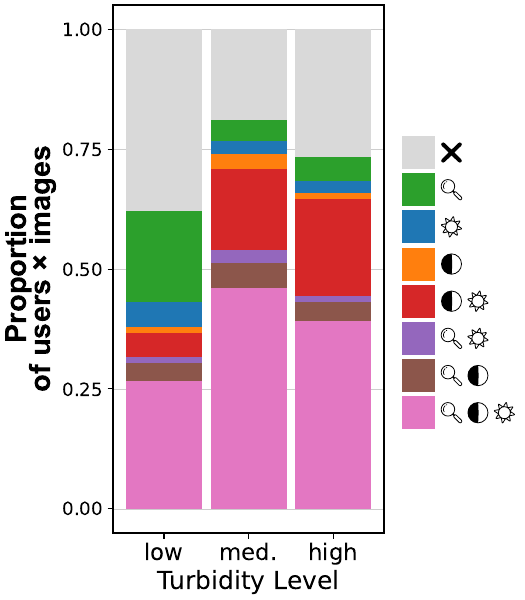}
    \caption{Type and frequency of image adjustments, individually and in combination.\\ \twemoji{x}: \textbf{no} sliders were adjusted. \\\twemoji{mag}: \textbf{only the zoom} was adjusted.\\ \twemoji{sun}: \textbf{only the brightness} was adjusted.\\ \twemoji{first_quarter_moon}: \textbf{only the contrast} was adjusted. \\\twemoji{first_quarter_moon}\twemoji{sun}: \textbf{both the contrast and brightness} were adjusted. And so on.\\ Note that this plot is the same as Figure~\ref*{fig:boxplot_image_updates_by_turbidity_level_and_event} in the main text, just enlarged.}
    \label{fig:barplot_image_settings_changes_by_turbidity_level_extra}
\end{figure}

\paragraph{Image adjustments} \cref{fig:boxplot_image_settings_changes_by_turbidity_level_extra} shows by \textit{how much} the brightness/contrast/zoom were adjusted in cases where they \textit{were} adjusted. \cref{fig:barplot_image_settings_changes_by_turbidity_level_extra} shows the proportion of annotated images for which each type of adjustment was made (ranging from no adjustment of any type, to all sliders adjusted).

\begin{figure}[tbh]
    \centering
    \includegraphics[height=5cm]{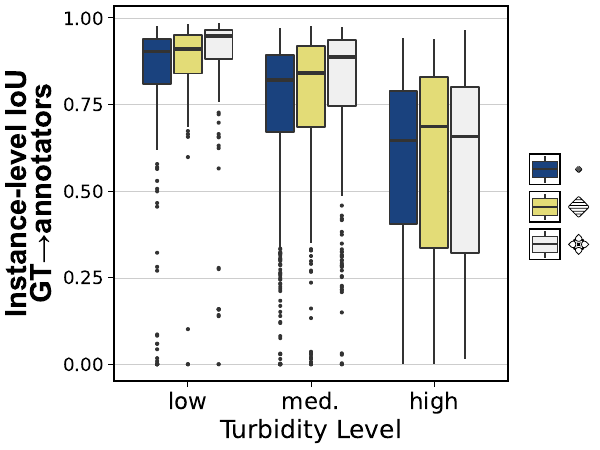}
    \caption{Does \textbf{adjusting individual contours} help? Here we look at the number of adjustments per annotated object for different turbidity levels. For each turbidity level, we create three (roughly) equally sized groups (\twemoji{1f539} small, \twemoji{1f537} medium, \twemoji{1f4a0} large [number of adjustments]), split at the 33rd (100/3) and 66th (200/3) percentile. }
\label{fig:boxplot_event_iou_by_turbidity_level_extra}
\end{figure}

\paragraph{Annotation adjustments} In \cref{fig:boxplot_event_iou_by_turbidity_level_extra} we group annotated objects by number of instance-level adjustments (polygon drawn, polygon moved, polygon deleted, point inserted, point moved, point deleted) per turbidity level.

\begin{figure}[tbh]
    \centering
    \includegraphics[height=4cm]{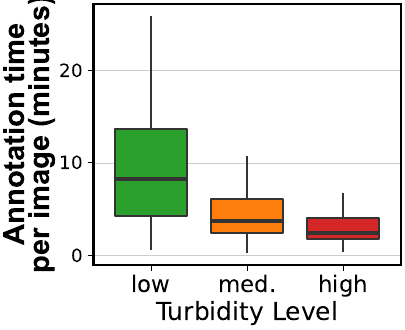}\hfill\includegraphics[height=4cm]{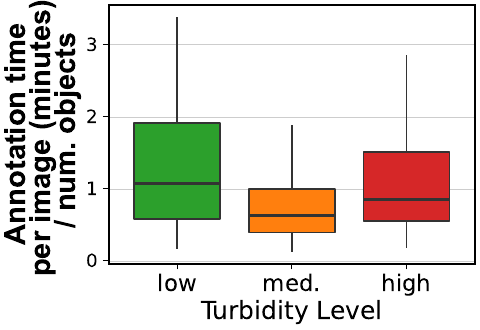}
    \caption{Distribution of annotation time per image (\textbf{left}) and distribution of annotation time per image normalized by the number of annotated objects (\textbf{right}). Outliers are not shown for readability, as they heavily distort the plot scale.}
\label{fig:boxplot_annotation_time}
\end{figure}

\begin{figure}[tbh]
    \centering
    \includegraphics[height=5cm]{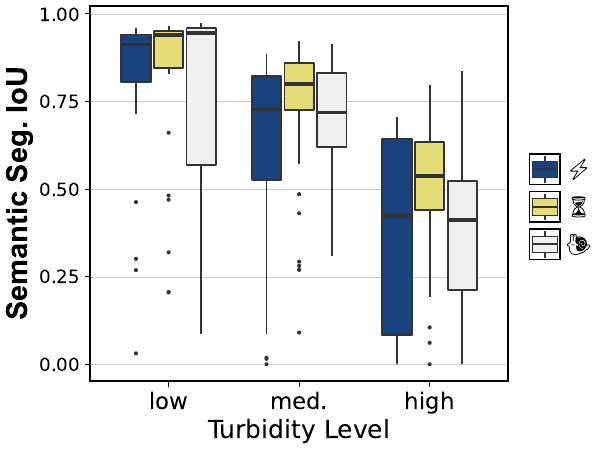}
    \caption{Does \textbf{time per image} help? Here we look at the time in minutes spent on every annotated image, for different turbidity levels. For each turbidity level, we create three  (roughly) equally sized time groups (\twemoji{26a1} fast, \twemoji{23f3} medium-speed, \twemoji{1f40c} slow), split at the 33rd (100/3) and 66th (200/3) percentile.  Note that this plot is the same as Figure~\ref*{fig:boxplot_time_iou_by_turbidity_level} in the main text, just enlarged.}
\label{fig:boxplot_time_iou_by_turbidity_level_extra}
\end{figure}

\paragraph{Time} As shown in \cref{fig:boxplot_annotation_time}, users spent a median of 8.2 minutes annotating clear images, vs. 3.7 minutes under medium turbidity and 2.4 under high turbidity. When normalizing by the number of annotated objects (to get the average time per object), users spend the least time per object in medium turbidity (median of 0.63 minutes per object) and the most time in clear images (1.07 minutes per object). In \cref{fig:boxplot_time_iou_by_turbidity_level_extra} we group annotated images by annotation speed per turbidity level, and compare the 3 speed groups (\twemoji{26a1} fast, \twemoji{23f3} medium-speed, \twemoji{1f40c} slow).

\subsection{All individual annotations}\label{app:indiv_annot}

\ifincludeannotdump

\input{supplementary_cr_rawdata}

\else 

The individual annotations collected in this study can be found in the separate supplementary document.

\fi

\end{document}